\documentclass[10pt,journal,compsoc]{IEEEtran}
\usepackage{amsmath,amsfonts}
\usepackage{algorithmic}
\usepackage{algorithm}
\usepackage{array}
\usepackage[caption=false,font=normalsize,labelfont=sf,textfont=sf]{subfig}
\usepackage{textcomp}
\usepackage{stfloats}
\usepackage{url}
\usepackage{verbatim}
\usepackage{graphicx}
\usepackage{cite}

\usepackage{multirow}
\usepackage[pagebackref=true,breaklinks,colorlinks,citecolor=blue,linkcolor=blue,urlcolor=blue,hypertexnames=false]{hyperref}
\usepackage{url}            
\usepackage{booktabs}       
\usepackage{amsfonts}       
\usepackage{nicefrac}       
\usepackage{microtype}      
\usepackage{graphicx}
\usepackage{amsmath}

\usepackage{array}
\usepackage{colortbl}

\usepackage{makecell}
\usepackage{amssymb}
\usepackage{pifont}

\usepackage[table]{xcolor}   

\usepackage{tabularx}

\newcommand{\xmark}{\ding{55}}%

\begin{document}

\title{AI in Agriculture: A Survey of Deep Learning Techniques for Crops, Fisheries and Livestock}

\author{Umair Nawaz, Muhammad Zaigham Zaheer, Ufaq Khan, Fahad Shahbaz Khan, Hisham Cholakkal, \\Salman Khan, Rao Muhammad Anwer
\IEEEcompsocitemizethanks{\IEEEcompsocthanksitem  U. Nawaz, M. Z. Zaheer, U. Khan, F. S. Khan, H. Cholakkal, S. Khan, and R. M. Anwer are with the MBZ University of AI, Abu Dhabi, UAE. \protect \\
E-mail: umair.nawaz@mbzuai.ac.ae
\IEEEcompsocthanksitem S. Khan is also with the CECS, Australian National
University, Canberra ACT 0200, Australia.
\IEEEcompsocthanksitem  F. S. Khan is also with the Computer Vision Laboratory, Linköping
University, Sweden.
}
}



\IEEEtitleabstractindextext{%
\begin{abstract}
Crops, fisheries, and livestock form the backbone of global food production, essential to feed the ever-growing global population. However, these sectors face considerable challenges, including climate variability, resource limitations, and the need for sustainable management. Addressing these issues requires efficient, accurate, and scalable technological solutions, highlighting the importance of artificial intelligence (AI). This survey presents a systematic and thorough review of more than 200 research works covering conventional machine learning approaches, advanced deep learning techniques (e.g., vision transformers), and recent vision-language foundation models (e.g., CLIP) in the agriculture domain, focusing on diverse tasks such as crop disease detection, livestock health management, and aquatic species monitoring. We further cover major implementation challenges such as data variability and experimental aspects: datasets, performance evaluation metrics, and geographical focus. We finish the survey by discussing potential open research directions emphasizing the need for multimodal data integration, efficient edge-device deployment, and domain-adaptable AI models for diverse farming environments. Rapid growth of evolving developments in this field can be actively tracked on our project page: \url{https://github.com/umair1221/AI-in-Agriculture}
\end{abstract}

}
\maketitle


\section{Introduction}
\IEEEPARstart{A}{griculture} (encompassing activities related to the cultivation of crops, aquaculture, and poultry farming, among other practices aimed at producing food) has experienced several transformative eras, each marked by significant technological advancements and shifts in practices. 
The earliest era, Agriculture 1.0 \cite{aggarwal2022transformations}, relied heavily on human and animal labor. Whether cultivating crops, tending livestock, or gathering fish and other aquatic resources, the tools were primitive, and productivity depended on manual effort. A profound shift occurred with Agriculture 2.0, marked by the widespread use of synthetic fertilizers, pesticides, and mechanical equipment. These innovations dramatically reduced labor demands, boosted yields across crop fields, and gradually influenced early industrial livestock operations and small-scale aquaculture ventures.

As technological innovations continued to advance, agriculture entered the Agriculture 3.0 phase, characterized by ``precision'' methods leveraging digital tools such as GPS and the Internet of Things (IoT). While initially more recognized in crop production, precision agriculture also found applications in livestock (e.g., farmers can track daily milk yields or feed consumption per animal for better feed optimization) and aquaculture (e.g., monitoring water quality, interpreting fish behaviors, and estimating biomass, residual feed, or water quality). The real-time data collection enabled better management of planting schedules, resource usage, and disease prevention across different farming domains. Today, Agriculture 4.0 \cite{gyamfi2024agricultural} builds upon these digital foundations by integrating AI-driven analytics, robotics, and advanced environmental sensing. Such technological convergence has proven beneficial not only for traditional crop farming but also for optimizing livestock operations (e.g., automated feeding systems) and aquaculture (e.g., underwater drones and advanced fish-monitoring sensors). The ultimate aim is to support more efficient, sustainable, and resilient food production ecosystems. Agriculture 4.0 has benefited significantly from the advancements in Artificial Intelligence (AI) and deep learning.

Limited applications of AI in Agriculture 4.0 begun with conventional approaches such as SVMs, ANNs, and Decision Trees. Subsequently, deep learning architectures, such as Convolutional Neural Networks (CNNs), have powered breakthroughs in image classification, object detection, and segmentation across agricultural domains \cite{attri2023review, araujo2023machine, luo2023survey}. For instance, CNN-based models can identify pests in crop fields or classify fish species in ``smart fish farming'' applications \cite{araujo2022fish}. CNN-based visual recognition Models like YOLO \cite{redmon2016you}, ResNet \cite{he2016deep}, Mask-RCNN \cite{he2017mask}, and GoogleNet \cite{szegedy2015going} automate tasks such as fish classification, counting, and segmentation \cite{saleh2024applications}, significantly saving time and labor.

\begin{figure}[hbtp]
    \centering
  \includegraphics[width=\columnwidth]{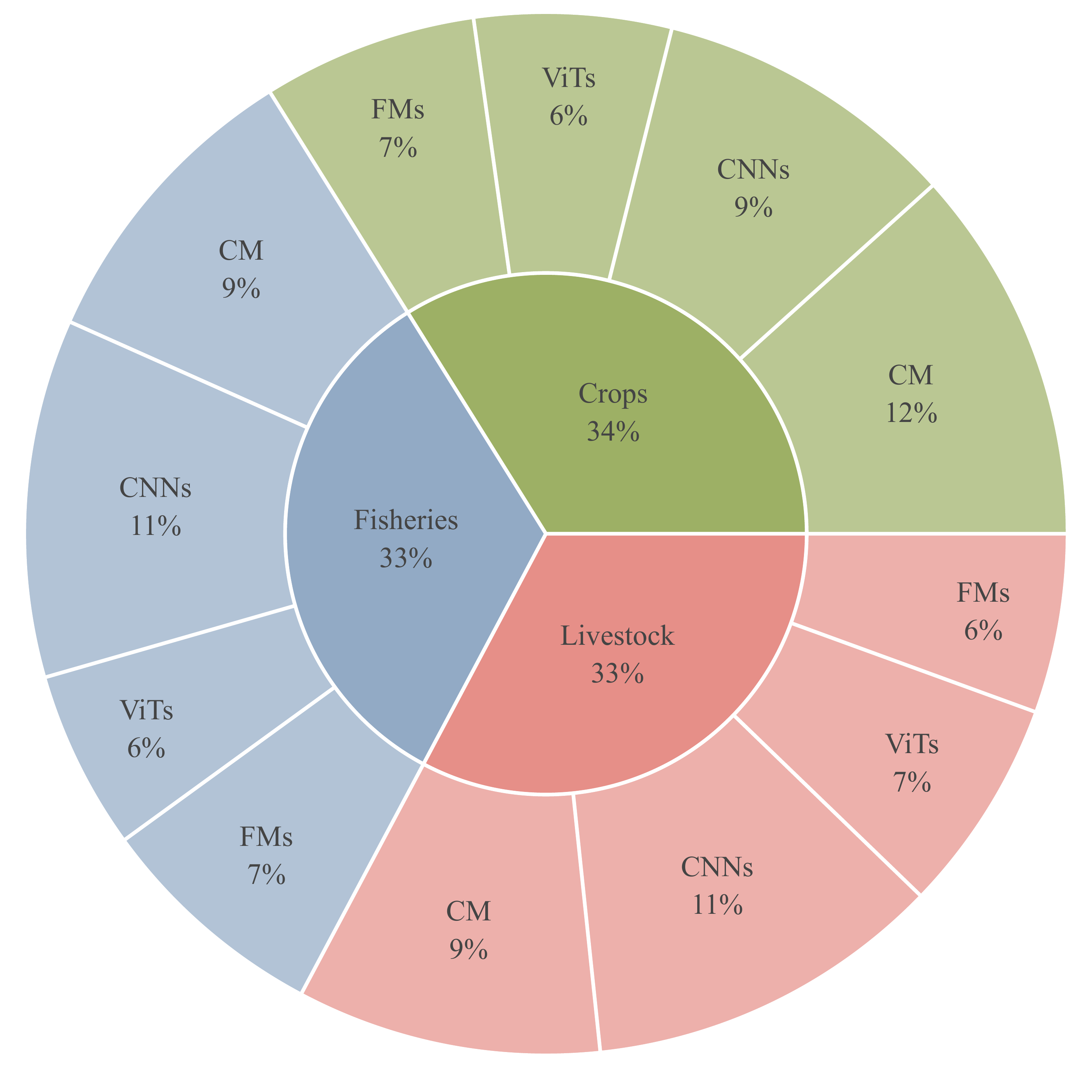}
  \caption{Pie chart representing the distribution of papers discussed in our survey under four methodological categories (CM: Conventional Methods, CNNs: Convolutional Neural Networks, ViTs: Vision Transformers, FMs: Foundational Models). Each domain’s angular span is proportional to its total paper count, and each inner slice’s radial extent encodes the method’s share within that domain in this survey.}
  \label{fig:survey-sections}
\end{figure}

These methods are likely to be valuable as the global population expands, demanding higher-quality yields from crops, livestock, and aquaculture alike. 
Despite these advantages, adopting CNN-based deep learning in agriculture comes with its own challenges. These include environmental variability, extensive data collection requirements, adapting to different applications, and the integration challenges of new AI systems into established practices \cite{luo2023survey}.


Beyond CNNs, deep learning extends itself in shifting toward foundation models, which are large-scale, pre-trained architectures (often based on transformers) \cite{10834497,chen2024internvl,stevens2024bioclip,wang2024sam} that can be adapted to numerous downstream tasks with minimal additional training. While CNNs remain highly effective for targeted image-focused tasks, foundation models hold promise for multi-modal agricultural data, where farms and aquaculture systems generate not just images but also sensor readings, text (e.g., farm logs), weather data, and more. These emerging models aim to unify disparate data sources, bringing new capabilities in predictive analytics, anomaly detection, and decision support.
This ensures that a single model can help grasp multiple tasks and overcome the need for application-specific models. For example, we can have a model that can identify different cattle diseases as well as differentiate between various breeds across multiple livestock animals. 


AI is also taking roots in utilizing macro-level data obtained using remote sensing tools such as satellites and UAVs. Remote sensing using satellites is becoming instrumental in monitoring large-scale crop fields and fish habitats, vast forested areas, aquaponic ponds, and extensive livestock rangelands, particularly addressing tasks such as crop yield assessments, biomass estimation, forest size estimation, herd counting, wildlife observation, and many others.
UAVs also have expanded possibilities in precision livestock farming, as discussed by Yousefi et al. \cite{yousefi2022systematic}. Through models like YOLO \cite{redmon2016you}, Faster-RCNN \cite{ren2015faster}, and Mask-RCNN \cite{he2017mask}, UAVs enable efficient, accurate livestock detection and tracking, particularly important for large herds or remote areas.
Overall, the use of deep learning has significantly improved the analysis of remote sensing data for agriculture \cite{epifani2024survey}.

\begin{figure}[hbtp]    
    \centering
    \includegraphics[width=\columnwidth]{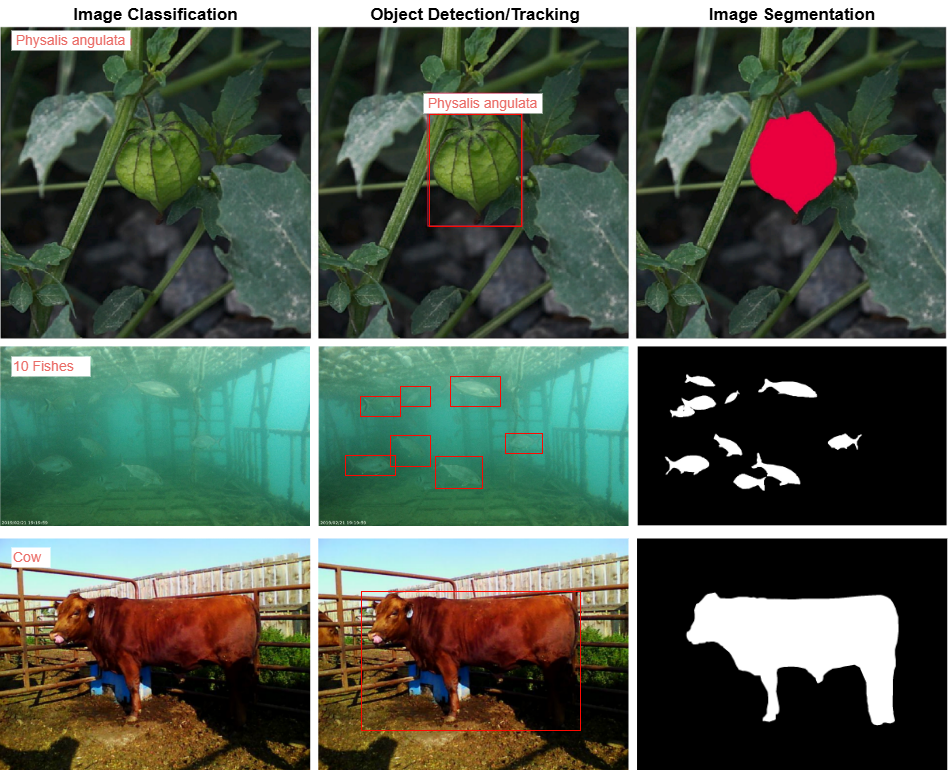}
    \caption{Different downstream tasks in crops, fisheries, and livestock domains. The classification task helps identify the presence of specific classes in an image, the detection/tracking task helps locate specific objects, and the segmentation task helps identify the pixel regions of the specific classes in an image.}
    \label{fig:downstream-tasks-all}
\end{figure}

\noindent\textbf{Scope of this Survey:} With the ever-evolving field of agriculture and its importance in our daily lives, several researchers have conducted extensive surveys to cover the topics of machine learning in agriculture that focus on the trends, challenges, and future aspects. However, these surveys typically focus on specific sectors like crops or fisheries and often discuss them in the context of traditional machine learning or deep learning techniques, as shown in Table~\ref{tab:reviewed-papers}.  
It can be seen that no existing survey covers the broad spectrum of \textit{all} the agriculture sectors, including crops, livestock, and fisheries together, and particularly not from the perspective of recent emergence of \textit{foundational models and vision language models}.

\begin{figure*}[hbtp]
    \centering
    \includegraphics[width=0.94\textwidth]{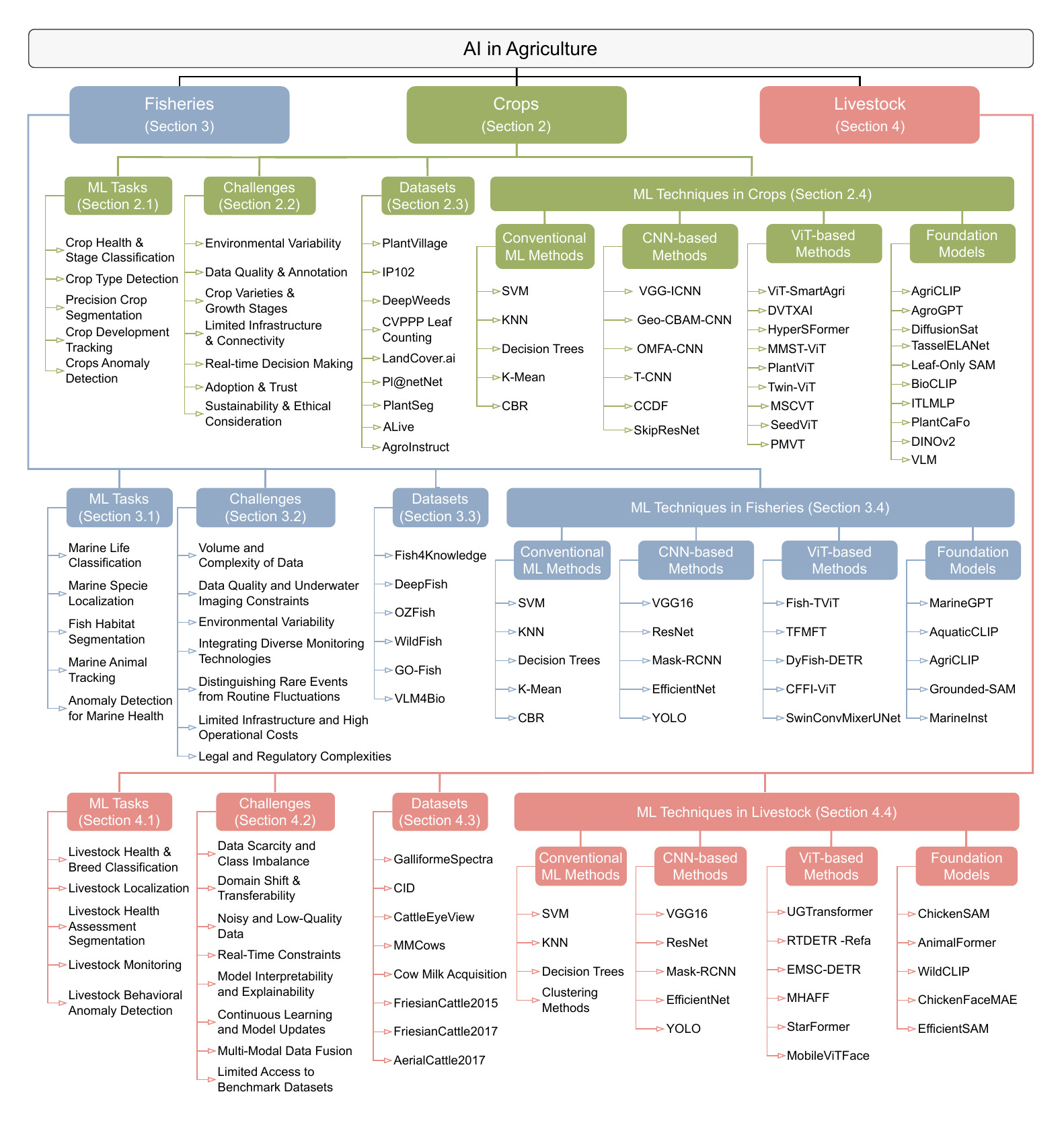}
    \caption{Taxonomy of this survey. AI applications in \textit{agriculture} are divided into three domains: crops, livestock, and fisheries. We highlight various ML tasks for each domain along with the domain-specific challenges. Furthermore, a detailed overview of common datasets in each domain is specified, along with different approaches used, ranging from conventional ML methods to foundation models.}
    \label{fig:taxonomy}
\end{figure*}

In this survey, we present a comprehensive overview of over $200$ papers spanning across conventional machine learning techniques, deep learning methods (CNNs, vision transformers, etc.) as well as the recent emergence of vision-language foundation models in the domain of crops, fisheries and livestock as also shown in terms of paper percentages in Fig.~\ref{fig:survey-sections}. We systematically review and analyze
how these AI techniques are currently being utilized and their potential applications in various tasks like classification, detection, and segmentation, as also shown in Fig.~\ref{fig:downstream-tasks-all} and Table~\ref{tab:research_papers}. In addition, we also discuss the different challenges that are faced in each domain. For \textbf{crops}, in Section \ref{sec:crops}, we review existing AI methods aiding in precision farming, pest management, crop planning, nutrients management, and yield prediction. We review AI methods leveraging data from ground sensors, satellite images, and drones. For \textbf{fisheries} domain, in Section \ref{sec:fish}, our survey covers methods addressing species recognition, sustainable fishing
practices, habitat monitoring, and population dynamics.  We also present review of AI techniques used for fish movements, predicting population trends, and managing aquaculture environments with
minimal human intervention. Moreover, for \textbf{livestock}, in Section \ref{sec:livestock}, we review methods focusing on health monitoring, breed improvement, cattle
disease diagnosis, genetic profiling, and behavior analysis. Figure~\ref{fig:taxonomy} shows the drone view of this survey to give a high-level picture of this survey paper. We also distinguish future research directions and the potential for scalability of AI models to promote sustainability. Lastly, we provide a publicly available \href{https://github.com/umair1221/AI-in-Agriculture}{GitHub} repository to make it effortless for researchers and students to track the latest literature on Agriculture in AI and understand the ongoing as well as future trends.

\begin{table*}[hbtp]
\centering
\caption{Comparative overview of key review papers on AI techniques in agriculture, illustrating that earlier surveys predominantly focused on conventional machine learning and CNN-based methods for specific domains (e.g., crops, livestock, or fisheries), whereas this work expands the scope to include Transformers and foundation models across crop, fisheries, and livestock applications, covering classification, detection, tracking, segmentation, anomaly identification, and generative tasks.}
\label{tab:reviewed-papers}
\resizebox{\textwidth}{!}{%
\begin{tabular}{p{4.5cm} c p{5cm} p{4.5cm} p{4cm}}
\toprule
\textbf{Title} & \textbf{Year} & \textbf{AI Techniques} & \textbf{Domains} & \textbf{Tasks} \\ 
\midrule

Applications of machine learning and deep learning in agriculture: A comprehensive review & 2025 & Conventional ML, CNNs & Crop & Crops Management; Harvest and post-harvest management; Agricultural water management \\ \midrule

Deep Learning for Sustainable Agriculture: A Systematic Review on Applications in Lettuce Cultivation & 2025 & Conventional ML, CNNs, ViTs & Crop & Pest and disease control; Crop monitoring; Field management \\ \midrule

A review of deep learning techniques used in agriculture & 2023 & CNNs & Crop & Yield prediction; Disease detection \\ \midrule
Machine Learning Applications in Agriculture: Current Trends, Challenges, and Future Perspectives & 2023 & Conventional ML & Crop, Livestock & Crop management; Water management; Soil management; Animal management \\ \midrule
A Survey of Computer Vision Technologies in Urban and Controlled-environment Agriculture & 2023 & CNNs & Crop & Growth monitoring; Pest detection \\ \midrule
Recent advances in Transformer technology for agriculture & 2024 & Transformers & Crop, Livestock & Disease diagnosis; Yield estimation; Automatic harvest \\ \midrule
Deep learning in UAV imagery for precision agriculture and wild flora monitoring & 2024 & CNNs & Crop & Monitoring; Classification; Detection \\ \midrule
Data collaborative sensing methods for smart agriculture & 2024 & Conventional ML & Crop & Data acquisition; Integration; Privacy \\ \midrule
Applications of deep learning in fish habitat monitoring & 2024 & CNNs & Fisheries & Classification; Counting; Localization; Segmentation \\ \midrule
A Survey of Deep Learning for Intelligent Feeding in Smart Fish Farming & 2023 & CNNs & Fisheries & Feeding behavior analysis; Growth monitoring \\ \midrule
Deep learning for smart fish farming: applications, opportunities, and challenges & 2021 & CNNs & Fisheries & Identification; Classification; Behavioral analysis; Feeding decision \\ \midrule
A systematic literature review on the use of machine learning in precision livestock farming & 2020 & Conventional ML & Livestock & Grazing management; Animal health \\ \midrule
A systematic literature review on the use of deep learning in precision livestock detection and localization & 2022 & CNNs & Livestock & Detection; Localization \\ \midrule
A systematic literature review on deep learning applications for precision cattle farming & 2021 & CNNs & Livestock & Health monitoring; Identification \\ \midrule
Ours & – & Conventional ML, CNNs, Transformers, Foundation Models & Crop, Fisheries, Livestock & Classification, Detection, Tracking, Segmentation, Anomaly Identification, Generative Tasks \\ 
\bottomrule
\end{tabular}%
}

\end{table*}

\section{AI for Crops}
\label{sec:crops}
In this section, we provide a comprehensive overview of the types of tasks and challenges that exist in the crop domain. Moreover, we also provide extensive discussion on existing datasets and techniques proposed by different researchers.

\begin{table*}[htbp]
\centering
\caption{Summary of different methods in fisheries, livestock, and crops domain covering the dataset and its performance (Perf.) as well. Here, the tasks consists of image classification (IC), image segmentation (IS), object tracking (OT), object detection (OD), and other specialised tasks. Besides, different methods are covered, including conventional methods (CM), convolutional neural networks (CNNs), Vision Transformers (ViTs), vision-language models (VLMs), and foundation models (FMs). A tick ($\checkmark$) indicates the study covers the corresponding task or employs the corresponding method.}
\label{tab:research_papers}
\resizebox{\textwidth}{!}{%
\begin{tabular}{@{}>{\raggedright\arraybackslash}p{1cm}>{\raggedright\arraybackslash}p{1cm}>{\raggedright\arraybackslash}p{3.2cm}>{\raggedright\arraybackslash}p{3.3cm}>{\raggedright\arraybackslash}p{2cm}>{\raggedright\arraybackslash}c>{\raggedright\arraybackslash}c>{\raggedright\arraybackslash}c>{\raggedright\arraybackslash}c>{\raggedright\arraybackslash}c>{\raggedright\arraybackslash}p{1.2cm}>{\raggedright\arraybackslash}p{3.2cm}>{\raggedright\arraybackslash}p{1.2cm}>{\raggedright\arraybackslash}p{2.7cm}>{\raggedright\arraybackslash}p{2.5cm}@{}}
\toprule
\multicolumn{3}{c}{\textbf{Paper}} & \multicolumn{2}{c}{\textbf{Dataset}} & \multicolumn{5}{c}{\textbf{Tasks}} & \multicolumn{5}{c}{\textbf{Methods}} \\
\cmidrule(lr){1-3} \cmidrule(lr){4-5} \cmidrule(lr){6-10} \cmidrule(lr){11-15}
\textbf{Name} & \textbf{GitHub} & \textbf{Venue} & \textbf{Name} & \textbf{Perf. (\%)} & \textbf{\textcolor{red}{IC}} & \textbf{\textcolor{blue}{IS}} & \textbf{\textcolor{brown}{OT}} & \textbf{\textcolor{green}{OD}} & \textbf{\textcolor{orange}{Others}} & \textbf{CM} & \textbf{CNNs} & \textbf{ViTs} & \textbf{VLMs} & \textbf{FMs} \\
\midrule
\midrule

\multicolumn{15}{c}{\textbf{Fisheries}} \\
\midrule
\midrule

\cite{kutlu2017multi} & \xmark & FEB & FinsData & Acc 92 & \checkmark & \xmark & \xmark & \xmark & \xmark & KNN & - & - & - & - \\

\cite{islam2019indigenous} & \xmark & IC4ME2 & BDIndigenous Fish2019 & Acc 89 & \checkmark & \xmark & \xmark & \xmark & \xmark & SVM & - & - & - & - \\

\cite{knausgaard2022temperate} & \xmark & Applied Intelligence & Fish4 Knowledge & mAP 86.96, 99.27 & \checkmark & \xmark & \xmark & \checkmark & \xmark & - & YOLOv3, CNN-SENet & - & - & - \\

\cite{vijayalakshmi2025aquayolo} & \xmark & Scientific Reports & DePondFi, DeepFish, OzFish & mAP@50 90.9, 42.0, 59.1 & \xmark & \xmark & \xmark & \checkmark & \xmark & - & AquaYolo & - & - & - \\

\cite{rani2024novel} & \xmark & Ecological Informatics & Aquatic WeightNet & 99 mAP & \checkmark & \xmark & \xmark & \checkmark & \xmark & Regressor Variants & mYolov8 & - & - & - \\

\cite{gong2023fish} & \xmark & Heliyon & FishNet, Fish-Pak & 94.33 (FishNet), 98.34 (FishPak) & \checkmark & \xmark & \xmark & \xmark & \xmark & - & - & Fish-TViT & - & - \\

\cite{li2024tfmft} & \href{https://github.com/vranlee/TFMFT}{Link} & Comp. and Elec. in Agri. & MFT22 & 30.6 IDF1 & \xmark & \xmark & \checkmark & \checkmark & \xmark & - & - & TFMFT & - & - \\

\cite{yildiz2024fisheye} & \xmark & European Food Research and Technology & Fisheye Freshness & 77.3 & \checkmark & \xmark & \xmark & \xmark & \xmark & KNN, SVM, LR, RF, ANN & VGG19, SqueezeNet & - & - & - \\

\cite{nawaz2024agriclip} & \href{https://github.com/umair1221/AgriCLIP}{Link} & COLING & ALive & Multiple Tasks & \checkmark & \xmark & \xmark & \xmark & \checkmark & - & - & - & CLIP & AgriCLIP, DINO \\

\cite{zheng2024marineinst} & \href{https://github.com/zhengziqiang/MarineInst}{Link} & ECCV & MarineInst20M & Multiple Tasks & \xmark & \checkmark & \xmark & \xmark & \checkmark & - & - & - & MarineGPT, Vicuna, CLIP & SAM, BLIP2 \\

\cite{hasegawa2024robust} & \xmark & Marine Sci. \& Eng. & Fish Image Bank & Acc 94.1 & \xmark & \checkmark & \xmark & \checkmark & \xmark & - & Mask R-CNN & - & - & Grounded SAM \\

\cite{yao2024fmrft} & \xmark & Preprint & Sturgeon Fish Tracking  & Acc IDF 90.3 & \xmark & \xmark & \checkmark & \checkmark & \xmark & - & Mamba & RT-DETR & - & - \\

\cite{alawode2025aquaticclip} & \xmark & arxiv & MAI, SAI, FishNet, FishNet Open Images dataset, Large Scale Fish Dataset, CSC, CC  & Multiple Tasks & \checkmark & \xmark & \checkmark & \checkmark & \checkmark & - & - & - & MarineGPT, CLIP & AquaticCLIP  \\

\midrule
\midrule

\multicolumn{15}{c}{\textbf{Livestock}} \\
\midrule
\midrule

\cite{sowmya2022classification} & \xmark & ICCMDE & Animal Class. & Acc 99 & \checkmark & \xmark & \xmark & \xmark & \xmark & SVM & MobileNet & - & - & - \\

\cite{mahmoud2015cattle} & \xmark & ICIEV & Cattle Muzzle  & Acc 100 & \checkmark & \xmark & \xmark & \xmark & \xmark & KNN, Fuzzy KNN & - & - & - & - \\

\cite{megahed2022comparison} & \xmark & Preventive Vet. Med. & Brucella Disease & AUROC 98.0 & \xmark & \xmark & \xmark & \xmark & \checkmark & Decision Trees, Logistic Regr. & - & - & - & - \\

\cite{ruchay2024barn} & \xmark & Comp. and Elec. in Agri. & On-barn Cattle Face & Acc 97.1 & \checkmark & \xmark & \xmark & \xmark & \xmark & - & VGG16 & - & - & - \\

\cite{wu2023improved} & \href{https://github.com/PuristWu/Identifying-gender}{Link} & Comp. and Elec. in Agri. & Public Chicken Gender & Acc 98.4 & \checkmark & \xmark & \xmark & \xmark & \xmark & - & ResNet50, AlexNet, GoogleNet, VGG16 & - & - & - \\

\cite{bouchekara2022sift} & \xmark & Drones & Cow Images (Kaggle) & Acc 95.5 & \checkmark & \xmark & \xmark & \xmark & \xmark & - & SIFT-CNN & - & - & - \\

\cite{qiao2019cattle} & \xmark & Comp. and Elec. in Agri. & Cattle Images & Acc 92.0 & \xmark & \checkmark & \xmark & \xmark & \xmark & - & Mask R-CNN & - & - & - \\

\cite{de2022automatic} & \xmark & Comp. and Elec. in Agri. & Cattle Rib-Eye Ultrasound & IoU 97.3 & \xmark & \checkmark & \xmark & \xmark & \checkmark & - & UNet++, FCN, U-Net, SegNet, Deeplabv3+ & - & - & - \\

\cite{zhao2024detrs} & \xmark & JRTIP & Cattle Breed
Datasets & IoU 97.3 & \xmark & \checkmark & \xmark & \xmark & \checkmark & - & - & RTDETR‑Refa & - & - \\

\cite{xu2020automatic} & \xmark & ICVCIP & Sheep Counting & Acc 99.0 & \xmark & \xmark & \checkmark & \checkmark & \xmark & - & Yolov3 + Kalman Filter & - & - & - \\

\cite{vayssade2023wizard} & \xmark & Comp. and Elec. in Agri. & Goats Private Data & PGT 86.1 & \xmark & \xmark & \checkmark & \checkmark & \xmark & - & CNN-TF, Yolov7, DeepSort & -  & - & - \\

\cite{qazi2024animalformermultimodalvisionframework} & \xmark & CVPR Workshop & Sheep Activity Dataset & Multiple Tasks & \xmark & \checkmark & \xmark & \checkmark & \checkmark & - & - & ViTPose  & - & Grounding DINO, HQSAM \\

\cite{gabeff2024wildclip} & \href{https://github.com/amathislab/wildclip}{Link} & IJCV & Snapshot Serengeti & Retrieval mAP 60.0 & \checkmark & \xmark & \xmark & \xmark & \checkmark & - & - & -  & CLIP, WildCLIP & - \\

\cite{noe2024vision} & \xmark & ICGEC & Black Cattle Behavior & Acc 94.7 & \xmark & \xmark & \checkmark & \checkmark & \checkmark & - & Yolov9, Deep OC SORT & -  & LLaVA & LLaMA \\

\midrule
\midrule
\multicolumn{15}{c}{\textbf{Crops}} \\
\midrule
\midrule

\cite{zarboubi2025custombottleneck} & \xmark & Comp. and Elec. in Agri. & PlantVillage & Acc 99.12 & \checkmark & \xmark & \xmark & \xmark & \xmark & - & CustomBottleneck-VGGNet, VGG16, VGG19 & - & - & - \\

\cite{lu2025leafconvnext} & \xmark & Comp. and Elec. in Agri. & Plant Diseases Expert, PlantVillage & Acc 99.68 & \checkmark & \xmark & \xmark & \xmark & \xmark & KNN, SVM & LeafConvNeXt & - & - & - \\

\cite{ma2019fully} & \xmark & Plos One & Rice Seedlings \& Weeds & mIoU 0.618 & \xmark & \checkmark & \xmark & \xmark & \xmark & - & SegNet, FCN, U-Net & - & - & - \\

\cite{javanmardi2021computer} & \xmark & Stored Product Research & Custom - Corn Seeds & F1 score 98.1 & \checkmark & \xmark & \xmark & \xmark & \xmark & SVM, KNN, Trees & VGG16 & - & - & - \\

\cite{zhao2021real} & \xmark & Comp. and Elec. in Agri. & Custom - Soybean Seed Defects & Acc. 97.84 & \checkmark & \xmark & \xmark & \xmark & \xmark & - & MobileNetV2-Improved, AlexNet, VGG19, GoogleNet, ResNet50, Shuffle Net, Squeeze Net  & - & - & - \\

\cite{singh2024effective} & \xmark & Exp. Sys. with Applications & PlantVillage \& PlantDoc & Acc. 99.92, 75.72  & \checkmark & \checkmark & \xmark & \xmark & \checkmark & - & LeafyGAN & MobileViT & - & - \\

\cite{parez2023visual} & \xmark & Sensors & PlantVillage, PlantComposite \& Data Repository of Leaf Images & Acc. 100, 99, 98  & \checkmark & \xmark & \xmark & \xmark & \xmark & - & -  & GreenViT & - & - \\

\cite{wang2025insect} & \xmark & Comp. and Elec. in Agri. & Custom - Insect Photos \& VOC2012 & AP 93.8, 69.1 & \xmark & \xmark & \xmark & \checkmark & \xmark & - & Insect-YOLO, YOLOv8m, YOLOv8n  & - & - & - \\

\cite{zhu2025potato} & \href{https://github.com/TomGoo474/Multimodel-AI-in-potato}{Link} & Comp. and Elec. in Agri. & Potato
Leaf Disease, Tomato Leaf Disease, and Eggplant Leaf Disease Datasets & Acc 98.43, 99.20, and 97.30 & \checkmark & \xmark & \xmark & \xmark & \checkmark & - & CT-CNN & Multi-scale ResViT & Multi-scale TextCNN, PotatoGPT & GPT-4 \\

\cite{hu2024lvf} & \href{https://github.com/Peter-blue/LVF}{Link} & Comp. and Elec. in Agri. & Custom, Public Rice, and PlantVillage (Tomato, Maize) & IoU 73.1, -, 83.4, 78.0 & \xmark & \checkmark & \xmark & \xmark & \checkmark & - & RIFN & LCN & Language Vision Framework (LVF) & - \\

\cite{zhang2024visual} & \xmark & Comp. and Elec. in Agri. & Curated - Wheat Disease Semantic Dataset & Acc 98.0 & \checkmark & \xmark & \xmark & \xmark & \checkmark & - & - & - & WDLM-Reasoning, CogVLM2, LLaVA-1.6, InternLM-XComposer2 & - \\

\bottomrule
\end{tabular}
}
\end{table*}

\subsection{Machine Learning Tasks in Crops Domain}

\noindent \textbf{Crop Health and Stage Classification \textcolor{red}{[IC]}}. This task is related to the broad application of machine learning in Image Classification. Image classification typically focuses on identifying broad features or conditions in images. In the context of crop farming, a classic example is disease detection, where images of leaves are classified between healthy or infected \cite{abd2023leaf}. This allows farmers to recognize potential disease attacks at an early stage and avoid any financial or mental stress in terms of yield loss. Similarly, classification can be used to distinguish different growth stages of a plant (e.g., seedling, vegetative, flowering) \cite{yu2023progress,das2022flowerphenonet,song2023recognition}, which is essential for scheduling irrigation and nutrient delivery. Another use case involves classifying images of harvested produce by quality grade which helps in sorting fruits and vegetables quickly and reducing post-harvest labor \cite{soltani2022defect}.

\begin{figure}[hbtp]
    \centering
    \includegraphics[width=\columnwidth]{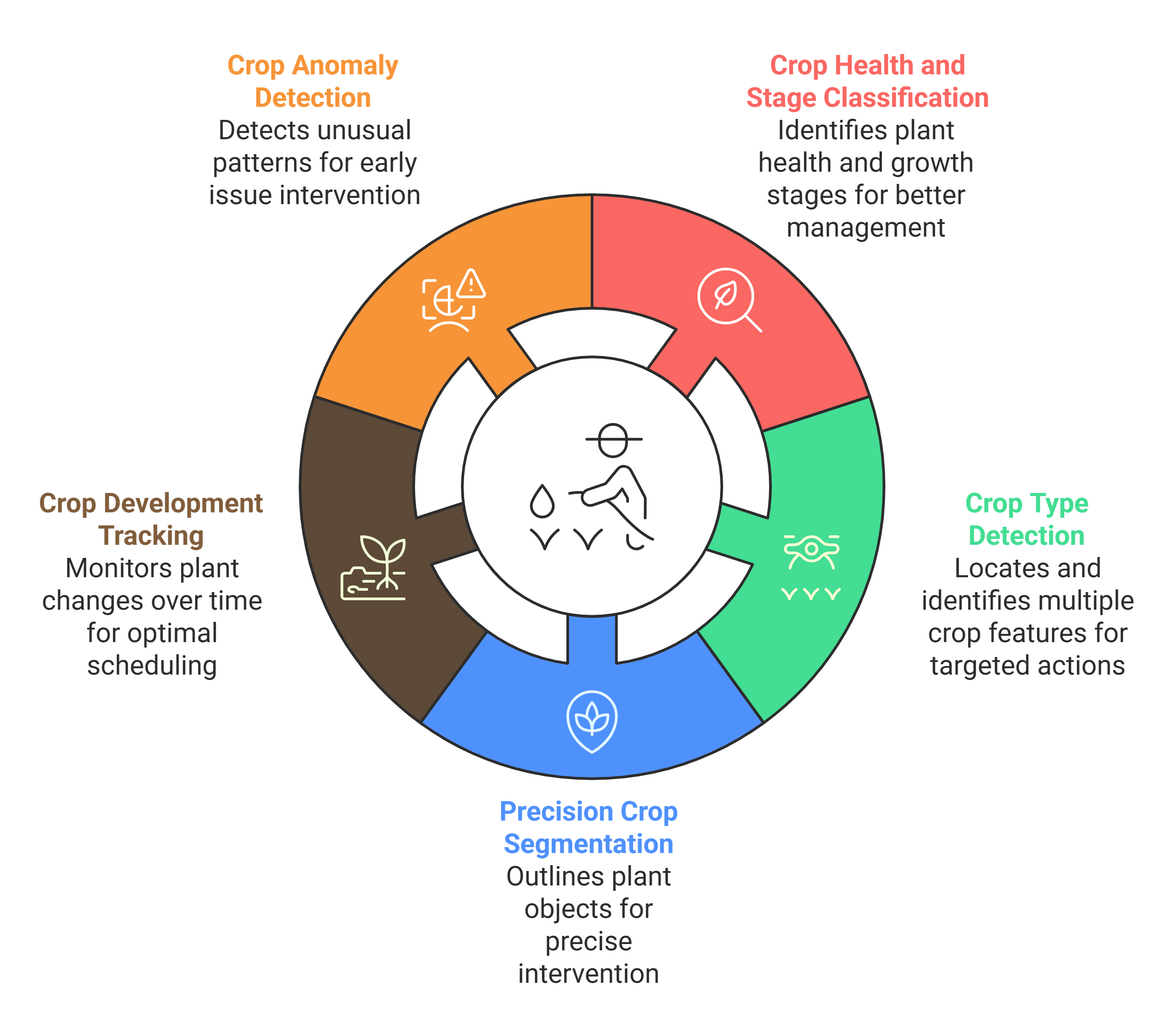}
    \caption{Tasks in the crop domain include anomaly detection for early issue intervention, image classification for assessing plant health, object detection to identify crop features, image segmentation for precise interventions, and object tracking for monitoring changes over time.}
    \label{fig:crops-tasks}
\end{figure}

\noindent \textbf{Crop Type Detection \textcolor{green}{[OD]}}. While classification assigns a label to the image of the crop, object detection identifies and locates multiple items within the frame (Fig.~\ref{fig:crops-tasks}). In crops, one of the most frequent applications is weed detection \cite{wu2021review}. By drawing bounding boxes around weeds in either ground \cite{islam2021early} or drone imagery \cite{rehman2024advanced}, farmers can precisely target unwanted plants which as a result can cut down on herbicide use and overall costs \cite{arguelles2023relational}. Object detection also proves valuable for tasks such as fruit detection, where specifying the exact location of apples, tomatoes, or citrus fruits can help with automated harvesting or yield estimation \cite{ilyas2023automated,darwin2021recognition}. One of the other important tasks in crops is pest detection \cite{chen2021smartphone}, which involves spotting insects or larvae directly on leaves or plant bushes. Moreover, by integrating object detection models with robotics or automated sprayers via UAVs \cite{meshram2022pesticide}, farmers can treat only the affected areas, thereby reducing chemical inputs.

\noindent \textbf{Precision Crop Segmentation \textcolor{blue}{[IS]}}. It takes the idea of detection one step further by outlining the exact shape of plant objects or regions. In the crops domain, segmenting weeds from crops helps machines better distinguish which plants to remove and which to leave untouched \cite{qu2024deep}. This pixel-level precision is particularly useful in dense fields where leaves overlap or where color differences between weeds and crops are minimal \cite{vayssade2023towards}. Segmentation also plays a vital role in monitoring plant growth over time by isolating leaves or stems. Different techniques can be used to measure leaf area or canopy size to assess health and vigor \cite{pagliai2022comparison}. Additionally, advanced segmentation techniques \cite{islam2024nutrient,taha2022using} can detect nutrient deficiencies or water stress in foliage patterns which enables precision interventions before there is a decline in the overall yield.

\noindent \textbf{Crop Development Tracking \textcolor{brown}{[OT]}}.  This task extends detection or segmentation over time, linking the same plants or features across multiple images or video frames. This is often used to monitor changes in plant size, shape, and health throughout the growing season \cite{mahlein2016plant}. For instance, a system might track how quickly a row of seedlings emerges from the soil \cite{yin2021soil}, or how large the fruit on each plant becomes as it ripens. When combined with environmental data like temperature, humidity, and soil moisture, tracking can reveal the influence of specific conditions on plant growth day by day. This information not only aids farm scheduling (e.g., predicting when harvesting should start) but also informs breeding programs seeking traits that succeed under certain environmental pressures.

\noindent \textbf{Crops Anomaly Detection \textcolor{orange}{[AD]}}. Anomaly detection usually focuses on identifying unusual patterns or deviations from the norm. In crop farming, these anomalies might be areas of a field that exhibit unexpected color changes, indicating disease outbreaks or nutrient imbalances \cite{amulothu2024machine,moso2021anomaly}. Sometimes, satellite or drone images can spot regions of a field with sudden plant die-off or significantly reduced vigor compared to surrounding areas. Anomaly detection can also help forecast events like pest infestations or drought stress if the system recognizes early warning signals \cite{amulothu2024machine} (e.g., irregular leaf temperature or soil moisture). Early detection means timely and quick responses which often prevents widespread crop damage or drought outbreak due to disease attack.

\subsection{Challenges in Crop Farming}
\begin{figure}[hbtp]
    \centering
    \includegraphics[width=\columnwidth]{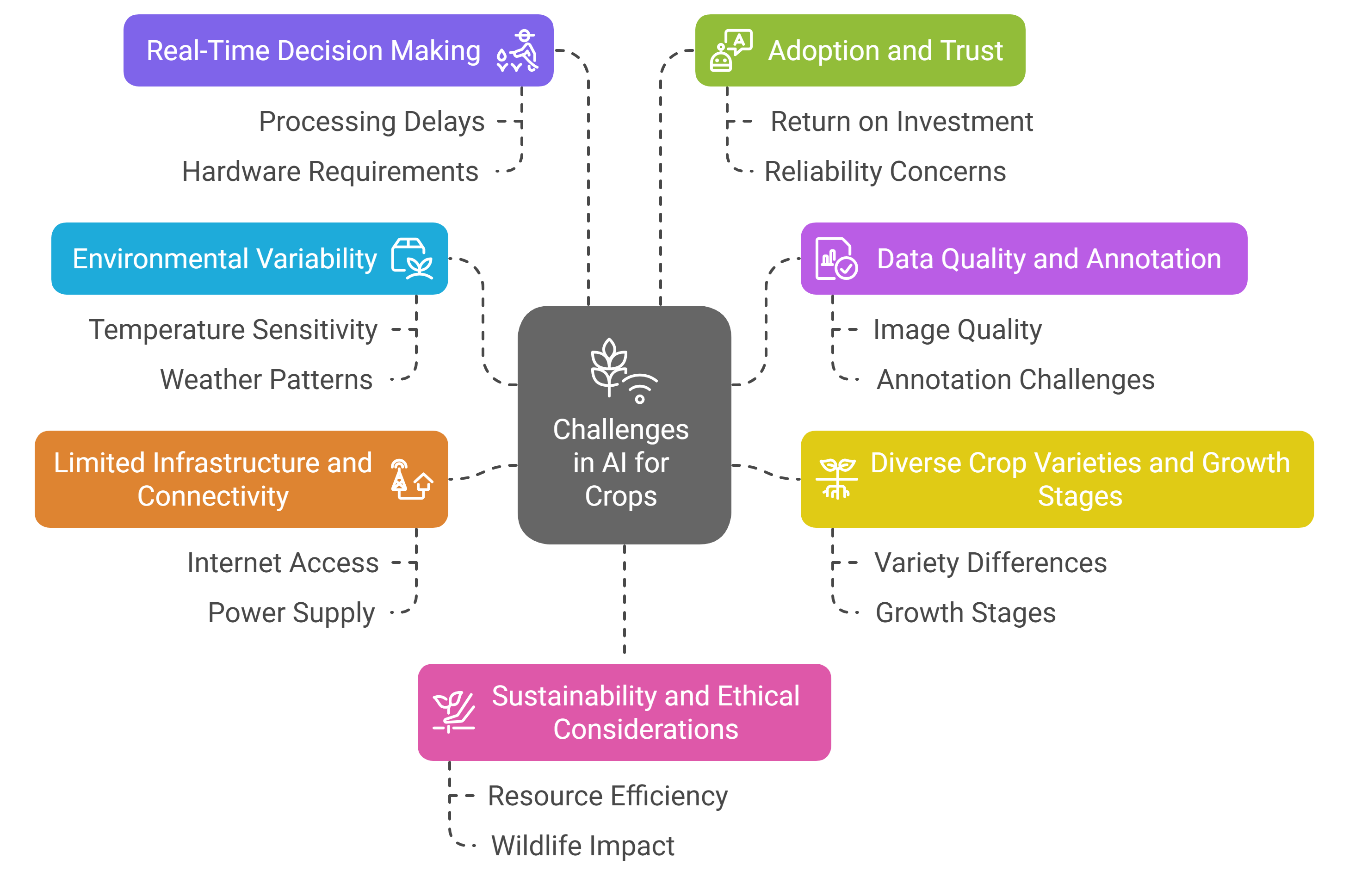}
    \caption{Visualization of the main challenges faced when implementing AI in crops, categorized into real-time decision-making, environmental variability, limited infrastructure, adoption and trust issues, data quality concerns, crop diversity, and sustainability considerations}
    \label{fig:crops-challenges}
\end{figure}

Some of the key challenges in crop farming are shown in Fig.~\ref{fig:crops-challenges} and discussed next.

\noindent \textbf{Environmental Variability.}
Crops are highly sensitive to local conditions such as temperature, rainfall, soil quality, and even microclimates that can differ between two sides of the same field. As a result, AI models trained on one region’s data may not perform well in another, where the climate or soil composition is drastically different \cite{javaid2023understanding}. Moreover, frequent changes in weather patterns from sudden storms to extreme heat can also shift plant growth trajectories, which then require AI systems to adapt or be retrained. This variability makes it challenging to create universal models that remain robust in all environments and across multiple growing seasons.

\noindent \textbf{Data Quality and Annotation.}
Data is one of the key important aspects of designing efficient machine learning solutions, and collecting reliable, high-quality images or sensor data over large fields can be difficult. The image quality may vary significantly from ground-level sources, due to inconsistent angles or obstructions (e.g., tall weeds, equipment in the way). Similarly, drone cameras might capture blurry photos due to wind or low lighting, while satellite imagery can be obscured by clouds. Furthermore, labeling these datasets for tasks like disease identification or weed mapping often requires specialized agricultural knowledge. Without accurate annotations, AI models may result in misclassifying diseases or missing critical plant-stress signals. Therefore, ensuring data consistency and thorough labeling is both time-consuming and expensive, especially for smallholder farms with limited resources.

\noindent \textbf{Diverse Crop Varieties and Growth Stages.}
Unlike typical machine learning applications, which often involve a more uniform and consistent set of classes in each setting, crop farms may feature a wide range of plant species and varieties, each with unique growth patterns and disease profiles. Even within a single crop type, growth stages (seedling, vegetative, flowering, fruiting, etc.) look very different, adding extra complexity for the machine learning model. What appears as a healthy sprout in one crop might resemble a weed in another, which may cause severe confusion. Hence, the designed models must be flexible enough to handle these variations (preferable), or farmers must train separate models for each plant variety and growth phase (costly and not preferable).

\noindent \textbf{Limited Infrastructure and Connectivity.}
Many farming regions lack consistent internet access, stable power, or high-end computing resources. Therefore, deploying computationally intensive models, especially those involving large architectures may consequently be impractical. Although edge computing offers some solutions by allowing on-device processing, these devices themselves are often expensive, require technical expertise, and may still struggle with large datasets. This infrastructure gap becomes more pronounced in remote or developing areas, where digital devices for agriculture are just beginning to emerge.

\begin{figure*}[hbtp]
    \centering
    \includegraphics[width=0.99\textwidth]{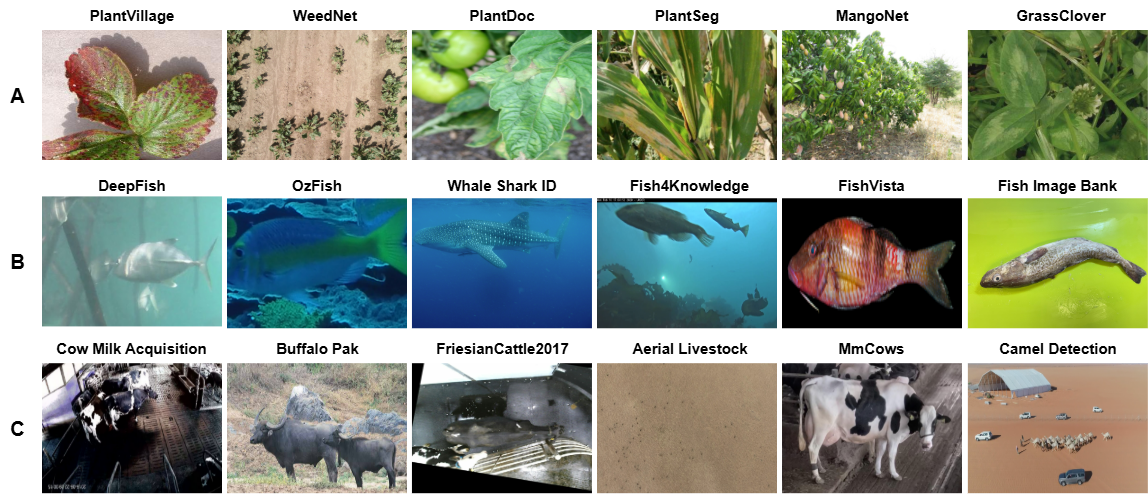}
    \caption{Sample images taken from some of the publicly available datasets used in agriculture. Here, (\textbf{A}) presents crop-focused datasets, including PlantVillage, WeedNet, PlantDoc, PlantSeg, MangoNet, and GrassClover, covering diverse plant health, weed detection, and segmentation tasks. (\textbf{B}) illustrates datasets related to fisheries and marine species identification, such as DeepFish, OzFish, Whale Shark ID, Fish4Knowledge, FishVista, and Fish Image Bank, showcasing underwater image variability and species diversity. (\textbf{C}) displays livestock datasets, including Cow Milk Acquisition, Buffalo Pak, FriesianCattle2017, Aerial Livestock, MmCows, and Camel Detection, highlighting varied scenarios of animal health monitoring, identification, and tracking in different environments.}
    \label{fig:all-datasets}
\end{figure*}
\noindent \textbf{Real-Time Decision Making.}
Timely information can be critical in crop farming, like the detection of a pest outbreak or water stress. Even a few days late can lead to significant yield loss and affects the farmer's annual profit \cite{zhang2022wheat}. However, many AI models need time for data collection, preprocessing, and inference. When run on drones, ground robots, or low-power devices, the processing delay can undermine the benefit of rapid responses. That is why achieving near-real-time or truly real-time analytics often requires specialized hardware, lightweight models, or distributed computing setups, which are costly to afford, and not every farmer can afford them.

\noindent \textbf{Adoption and Trust.}
Farmers may be hesitant to invest in AI-based solutions without clear evidence of return on investment or ease of use \cite{atapattu2024challenges}. They also may have reliability concerns, like if a model incorrectly flags a healthy plant as diseased, the farmer could unnecessarily spray pesticides or remove valuable crops. Conversely, failing to detect an actual disease in time could devastate a harvest. Therefore, building trust among farmers often involves offering transparent, explainable systems and demonstrating tangible benefits, such as reduced pesticide costs, improved yields, or simpler workflows, to convince the farmers of the advancements in this rapidly growing era of technology.

\noindent \textbf{Sustainability and Ethical Considerations.}
Although AI can significantly improve resource efficiency, like minimizing chemical use through precision spraying, there are ethical and environmental questions that need to be addressed. For instance, drones buzzing overhead can disturb local wildlife or infringe on neighboring farm privacy, depending on how data is collected. The balanced innovation with ecological consideration and respect for community interests remains a delicate matter.

\subsection{Existing Datasets in Crop Farming}
This sub-section discusses some of the notable datasets that have been used in research related to AI in crop farming. Table~\ref{tab:datasets-crops} provides a brief overview of these datasets and their characteristics, whereas Figrue~\ref{fig:all-datasets}A shows a few sample images.

\subsubsection{PlantVillage Dataset} 
The PlantVillage dataset \cite{mohanty2016using} is recognized worldwide for its extensive collection of labeled leaf images for different fruits and crops. It includes multiple plant species, such as tomato, potato, apple, strawberry, and maize, with each image labeled as either healthy or showing a particular disease (like blight or mosaic). Because the images often feature individual leaves on plain backgrounds, this dataset is especially useful for building and testing basic image classification or disease-detection models. Many researchers have leveraged this structured dataset to showcase a prototype solution \cite{appavu2025deep} capable of diagnosing plant diseases through smartphone images in field conditions.

\begin{table*}[htbp]
\centering
\caption{Summary of publicly available datasets used in crops domain. Each dataset is categorized based on different characteristics such as the number of classes, annotation type, dataset size, geographical focus, and supported tasks, such as Image Segmentation (IS), Object Detection (OD), and Image Classification (IC). }
\label{tab:datasets-crops}
\renewcommand{\arraystretch}{1.5} 
\resizebox{\textwidth}{!}{%
\begin{tabular}{@{}>{\raggedright\arraybackslash}p{2.5cm}>{\raggedright\arraybackslash}p{1.4cm}>{\raggedright\arraybackslash}p{3.5cm}>{\raggedright\arraybackslash}p{2.5cm}>{\raggedright\arraybackslash}p{2cm}>{\raggedright\arraybackslash}p{2cm}>{\raggedright\arraybackslash}p{1.8cm}>{\raggedright\arraybackslash}p{2.3cm}>{\raggedright\arraybackslash}p{1.5cm}@{}}
\toprule
\textbf{Dataset} & \textbf{Classes} & \textbf{Annotation} & \textbf{Dataset Size} & \textbf{Modality} & \textbf{Resolution} & \textbf{Geographical Focus} & \textbf{Tasks} & \textbf{Link} \\ \midrule

CWFI dataset & 10 & Masks & 60 images & MultiSpectral, Images & High & Germany & \textcolor{blue}{IS} & \href{https://github.com/cwfid/dataset}{Github} \\

Carrot-Weed & 2 & Masks & 39 images & Images & High & - & \textcolor{blue}{IS} & \href{https://github.com/lameski/rgbweeddetection}{Github} \\

Plant seedlings & 12 & Class Labels & 407 images & Images & High & Denmark  & \textcolor{red}{IC} & \href{https://vision.eng.au.dk/plant-seedlings-dataset/}{Link} \\

Grass-Broadleaf & 4 & Class Labels & 15,336 images & Images & Medium & -  & \textcolor{red}{IC} & \href{https://data.mendeley.com/datasets/3fmjm7ncc6/2}{Mendeley} \\

Sugar Beets 2016 & 4 & Multimodal, Masks & 15,336 images & Images & High & Germany  & \textcolor{blue}{IS} & \href{https://www.ipb.uni-bonn.de/datasets_IJRR2017/annotations/}{Dataset} \\

weedNet & 4 & Multispectral, Masks & 465 images & Images & High & -  & \textcolor{blue}{IS} & \href{https://github.com/inkyusa/weedNet}{Github} \\

Leaf counting & 9 & Class Labels & 9372 images & Images & Low & Denmark  & \textcolor{red}{IC} & \href{https://vision.eng.au.dk/leaf-counting-dataset/}{Dataset} \\

OPPD & 47  & Class Labels & 7590 images & Images & Low & Denmark  & \textcolor{green}{OD}, \textcolor{red}{IC} & \href{https://gitlab.au.dk/AUENG-Vision/OPPD/-/tree/master/}{Dataset} \\

Deep Fruits & 7  & Bounding Box & 587 images & Images & High & -  & \textcolor{green}{OD} & \href{https://drive.google.com/drive/folders/1CmsZb1caggLRN7ANfika8WuPiywo4mBb}{GDrive} \\

Orchard Fruit & 3  & Bounding Box, Circular & 3704 images & Images & High & Australia  & \textcolor{green}{OD} & \href{https://data.acfr.usyd.edu.au/ag/treecrops/2016-multifruit/}{Dataset} \\

Date Fruit & 5 & Class Labels &$>$10,000 images & Images & Medium & Saudi Arabia  & \textcolor{red}{IC} & \href{https://doi.org/10.21227/x46j-sk98}{Dataset} \\

MangoNet & 1 & Masks & 49 images & Images & High & India  & \textcolor{blue}{IS} & \href{https://github.com/avadesh02/MangoNet-Semantic-Dataset}{Github} \\

MinneApple & 1 & Masks, Bounding Box &$>$70,000 images & Images & High & United States  & \textcolor{blue}{IS}, \textcolor{green}{OD} & \href{https://doi.org/10.13020/8ecp-3r13}{Dataset} \\

Pl@netNet & 1,081 & Class Labels & 300,000 images & Images & Medium & - & \textcolor{red}{IC}, \textcolor{green}{OD} & \href{https://zenodo.org/records/4726653}{Zenodo} \\

PlantSeg & 115 & Class Labels, Masks & 11,400 images & Images & Medium & Global & \textcolor{red}{IC}, \textcolor{blue}{IS} & \href{https://github.com/tqwei05/PlantSeg}{Github} \\

PlantVillage & 38 & Class Labels & 54,306 images & Images & Medium & - & \textcolor{red}{IC} & \href{https://github.com/spMohanty/PlantVillage-Dataset}{Github} \\

PlantDoc & 30 & Class Labels, Bounding Box & 2,598 images & Images & Medium & - & \textcolor{red}{IC}, \textcolor{green}{OD}  & \href{https://github.com/pratikkayal/PlantDoc-Dataset}{Github} \\

IP102 & 102 & Class Labels, Bounding Box & 75,000 images & Images & Medium & - & \textcolor{red}{IC}, \textcolor{green}{OD}  & \href{https://github.com/xpwu95/IP102}{Github} \\

Sugarcane Billets & 6 & Bounding Box & 156 images & Images & Medium & United States & \textcolor{green}{OD} & \href{https://github.com/The77Lab/SugarcaneBilletsDataset}{Github} \\

DeepSeedling & 1 & Bounding Box & 5,743 images & Images & Medium & United States & \textcolor{green}{OD} & \href{https://figshare.com/s/616956f8633c17ceae9b}{Dataset} \\

GrassClover & 1 & Class Labels, Masks & 8,000 synthetic, 31,600 unlabeled & Images & Med-High & Denmark & \textcolor{red}{IC}, \textcolor{blue}{IS} & \href{https://vision.eng.au.dk/grass-clover-dataset/}{Dataset} \\

\bottomrule
\end{tabular}
}

\end{table*}

\subsubsection{IP102 Dataset}
Another notable dataset in plants is IP102 \cite{wu2019ip102}, which focuses exclusively on insect pests that endanger major crops like wheat, rice, and corn. Each image typically shows a specific pest species captured under real-world conditions including varying lighting, angles, and field environments. While the dataset primarily supports pest detection and classification tasks, some subsets also include different life stages of insects, making it valuable for targeted interventions. By using IP102, agronomists can detect pests early and employ selective pesticide strategies, thereby reducing both crop losses and environmental impact.

\subsubsection{DeepWeeds}
DeepWeeds \cite{olsen2019deepweeds} dataaset showcases multiple weed species that are commonly found across diverse regions. It is designed for weed recognition in agricultural and natural landscapes. It shows images of weeds at different growth stages and lighting conditions, usually within dense vegetation. This variety helps models learn to differentiate crops from unwanted plants (weeds), even in visually complex settings. Using such datasets can result in precise weed management, hence enabling precise herbicide spraying or mechanical removal that saves costs and promotes sustainable farming practices.

\subsubsection{CVPPP Leaf Counting Dataset}
The CVPPP \cite{minervini2016finely} Leaf Counting dataset is focused on plant phenotyping, which provides images where the main focus is on individual plants with distinct leaves. Many researchers use it to develop solutions that segment and count leaves. This helps measure growth rates, leaf area, and overall plant health. Although it primarily features smaller, greenhouse-grown specimens, the methodology can be transferred to field crops when combined with more extensive imagery. By using CVPPP data, tasks like growth monitoring, trait analysis, and automated yield estimation in breeding programs are achievable.

\subsubsection{LandCover.ai}
LandCover.ai \cite{boguszewski2021landcover} is not exclusively crop-oriented but it provides high-resolution aerial and satellite images labeled with various land-cover classes, including certain agricultural fields. This dataset allows the development of classification and segmentation techniques for identifying farmland boundaries, vegetation health, and potential issues like over- or under-cultivated areas. This is how, by integrating land-cover insights with other farm-level data, the users can better allocate resources and monitor larger regions for signs of water stress or crop failure. The mixture of rural and semi-urban landscapes also ensures that models trained on LandCover.ai handle diverse field layouts and land-management practices.

\subsubsection{Pl@ntNet}
The Pl@ntNet-300K \cite{garcin2021pl} dataset is a collection of over 300,000 plant images, covering 1,081 different species. It is derived from the Pl@ntNet project \cite{affouard2017pl} and is unique because it has a lot of variety and imbalanced categories. Some plant types are more common than others, with a large number of images, whereas many look alike, which can make the classification process hard and challenging. This also makes the dataset appropriate for testing new ways of image classification where the goal is not just to find the most likely category but to identify a set of possible categories for each image. This approach helps deal with the uncertainty and complexity of real-world data, where it's often not clear which category an image belongs to.

\subsubsection{PlantSeg}
The PlantSeg dataset \cite{wei2024plantseg} is designed to improve the segmentation and analysis of plant diseases within natural agricultural settings. Unlike most datasets that focus on detection or classification and are often collected under controlled conditions, PlantSeg includes 11,400 detailed segmentation masks across 115 disease categories captured in the wild, capturing the complexity of real-world environments. This extensive dataset is further supported by an additional 8,000 images of healthy plants that offer a unique way for developing more robust disease segmentation models.

\subsubsection{ALive}
The agriculture sector benefits significantly from advancements in AI-driven image-text pairing, exemplified by the creation of the ALive \cite{nawaz2024agriclip} dataset. This dataset, integral to the AgriCLIP framework, encompasses approximately 600,000 curated image-text pairs specifically designed for crops, fisheries, and livestock applications, ranging from disease detection to fishery species recognition. This dataset employs a customized prompt generation strategy that utilizes metadata and class-specific details, which provides rich descriptive image-guided captions that significantly enhance the granularity and contextuality of the data. The dataset facilitates more effective zero-shot learning by incorporating fine-grained visual and textual cues essential for tasks such as weeds or nutrient deficiency detection, thereby addressing some of the key challenges in the crops domain.

\subsubsection{AgroInstruct}
The AgroInstruct \cite{awais2024agrogpt} dataset is introduced to address the gap in domain-specific multimodal datasets for agriculture. This dataset comprises 70,000 expert-tuned entries created to enhance language models' understanding of agricultural concepts without requiring joint image-text data. It Utilizes vision-only agricultural datasets by synthesizing instruction-tuning data through a novel pipeline that leverages class-specific information and large language models to generate context-rich image descriptions and corresponding expert annotations. This method enables the effective training of models like AgroGPT \cite{awais2024agrogpt}, which are tailored to address complex agricultural queries by providing detailed insights into the agricultural domain.


\subsection{Machine Learning Techniques in Crops Farming}

\subsubsection{Conventional Approaches}

\noindent \textbf{Support Vector Machines.} SVMs are ideal for scenarios where data is limited or where on-device computing power is scarce. Many recent studies demonstrate how thoughtful feature selection and parameter tuning can enable SVMs to perform on par with or even surpass several other neural approaches.
Camps-Valls et al. \cite{camps2003support} explored the application of SVMs to classify different crop types from hyperspectral images and compared their performance to that of common neural network approaches (multilayer perceptrons, radial basis functions, and co-active neural fuzzy inference systems). The experiments used six hyperspectral scenes acquired with a 128-band HyMap sensor and contained six classes (corn, sugar beet, barley, wheat, alfalfa, and soil). The paper evaluated classification accuracy across four input configurations, i.e., using all 128 bands (without pre-processing) as well as representative feature subsets of 6, 3, and 2 bands \cite{gomez2003semi,gomez2003cart}, thereby simulating cases where pre-processing (feature selection) is feasible. The validation was conducted via cross-validation on 150 samples per class and then tested on over 327,000 pixels per scene. Here, SVMs consistently outperformed their counterpart neural approaches (e.g., multilayer perceptions, Radial Basis Functions, and Co-Active Neural Fuzzy Inference System), and remain computationally viable without a prior feature-selection stage.

Addressing the need for precise yield forecasts to aid farmers’ planning, Priyadharshini et al. \cite{priyadharshini2022enhanced} presented a tailored linear SVM classifier that selects optimal support vectors based on auxiliary agronomic inputs. They compiled a dataset of 650 observations across twelve variables, such as soil nitrogen content, annual groundwater level, rainfall, and temperature, sourced from official government records. After cleaning (outlier removal) and scaling these inputs, they carried out yield prediction as a binary classification problem having two classes, i.e., crop yield fit and not fit. The linear SVM model then finds the hyperplane of maximum margin in this feature space, optimizing the distance between the two classes. 

Moving forward to the early detection of diseases, Kumar et al. \cite{kumar2017precision} proposed a system that fuses environmental sensing with image-based infection analysis to identify diseases in sugarcane crops. In this system, the fixed sensors continuously record field conditions (temperature, humidity, and soil moisture), while a network-connected camera captures leaf images at regular intervals. The image-processing pipeline first uses K-nearest neighbors clustering to segment regions of interest, then applies an SVM classifier to distinguish healthy foliage from early signs of infection (e.g., leaf scald, red stripe, or mosaic patterns). Furthermore, the sensory alerts and classification results are also transmitted wirelessly to a control unit, which synthesizes recommendations for timely interventions.

\noindent \textbf{k-Nearest Neighbours.} kNN is considered a strong and intuitive, instance-based, learning method where a query point is labeled according to the majority class among its k closest neighbors in feature space. It makes no parametric assumptions, and that is why it is simple to implement and adapt naturally to multi-class problems. One drawback, however, is that it may suffer from high memory and computation costs at prediction time because of the higher number of comparisons with the nearest neighbors. Sundari et al. \cite{rekha2021crop} present a crop‐recommendation system that leverages the simplicity and interpretability of kNN to suggest optimal crops based on soil and climate features. They conducted laboratory analysis to obtain input data such as temperature, rainfall, pH, and historical production, followed by preprocessing (handling missing values, encoding categorical crop labels, etc.), scaling, and dividing into 80/20 training/testing splits. The KNN classifier is then evaluated across various k values, achieving its highest training‐data accuracy of 98.10\% at $k=5$. The authors reported that this performance holds on unseen test samples, demonstrating KNN’s effectiveness even without careful feature selection or complex modeling. Kumar et al. \cite{kumar2023precision} developed a crop-recommendation system also based on KNN but evaluated both accuracy and F1-score to ensure balanced precision-recall. They used a publicly available Kaggle dataset that contains soil NPK (nitrogen, phosphorous, potassium) levels, temperature, humidity, pH, and rainfall. They encoded categorical features and partitioned the data into the same splits of 80/20. With $k=10$, their KNN model achieved 96\% accuracy and an F1-score of 0.96 on the test set, outperforming decision trees, random forests, SVMs, and neural networks on the same tasks. 

Moving on to leaf diseases, Vaishnnave et al. \cite{vaishnnave2019detection} propose an automated pipeline for detecting and classifying groundnut-leaf diseases, specifically early leaf spot, late leaf spot, rust, and bud necrosis, using kNN in place of SVM for improved performance. Their approach consists of following steps: (1) image acquisition via camera or smartphone, (2) preprocessing to denoise and enhance contrast, (3) segmentation into binary masks and conversion to HSV space, (4) color-cooccurrence–based feature extraction across H, S, and V channels, and (5) KNN classification using multiple values of k. They collected 250 images, out of which 45 denoised samples trained the classifier and 105 unseen images were used to test the model. Li and Ercisli \cite{li2023data} shift the focus from model tweaks to data quality by introducing the KNN-Distance Entropy (KNN-DE) metric to assess and screen the most informative crop-pest images for few-shot learning. They used a balanced dataset of six pest classes (1,000 images each) and extracted 512-dimensional features via ResNet-18 and computed an entropy score over the K closest class prototypes for each sample. In meta-task experiments with budgets of 10–100 samples per class, training exclusively on the top-KNN-DE high-informative images yielded up to 17\% higher accuracy than training on the lowest-KNN-DE samples, and reduced necessary data (and thus training time) by a factor of 2.5. They further demonstrated KNN-DE’s utility in identifying negative augmented samples that degrade performance, which underscores the method’s value for data-efficient, sustainable agricultural AI in crops.

\noindent \textbf{Decision Trees.} Decision Trees partition the feature space by making a sequence of hierarchical, interpretable decisions where each node tests a feature’s value, and the branches represent a possible outcome. These trees balance predictive power with transparency where they capture non-linear relationships and interactions while providing a clear `if-then' logic that is easy to visualize and explain. Chopda et al. \cite{chopda2018cotton} apply a classic Decision Tree Classifier to real-time cotton disease detection by combining environmental sensing and mobile‐app feedback. Their system captures ambient temperature and soil moisture via DHT11 and KG003 sensors connected to an Arduino Uno, which transmits the readings over Wi-Fi (ESP8266) to a Thingspeak server, which then uses a decision tree trained on historical sensor data to predict the onset of key cotton diseases (Anthracnose, Greymildew, Wilt). The pipeline includes sensor data acquisition, preprocessing, database comparison, classification via decision tree rules, and delivery of alerts through an Android application. The hardware architecture and sensor specifications ensure low-cost, in-field deployment, which demonstrates how simple supervised models can enable smart farming on resource-constrained devices.

Cintra et al. \cite{cintra2011use} introduced fuzzy decision trees (FUZZYDT) to highlight any coffee rust outbreaks by leveraging eight years of monthly infection data from experimental plots in Brazil. They constructed six binary classification datasets by varying attribute subsets (meteorological and planting‐density features) and infection‐rate thresholds ($\geq5$ pp and $\geq10$ pp). The authors used triangular fuzzy sets to fuzzify numeric attributes and applied standard entropy‐based splitting but retained multiple rules firing with degrees of compatibility. In comparison to J48 decision trees, the fuzzy models achieved lower cross-validated error rates, simpler rule bases, and enhanced interpretability, allowing agronomists to trigger preventive or curative actions based on linguistic thresholds of temperature and humidity. Lin et al. \cite{lin2022validation} developed a post-classification workflow that refines the USDA’s Cropland Data Layer (CDL) using a spatial-temporal decision tree algorithm. First, a spatial filter is used to flag isolated and boundary pixels by comparing each candidate to its eight neighbors. Then, nine years of historical CDL maps determined a dominating land cover class per pixel (using thresholds of $\geq7/9$ years for constant features, $\geq5/9$ for rotating crops). The pixels having current classification conflicts with both neighborhood majority and historical dominance are marked for correction. The decision tree then reassigns classes based on combined spatial context and temporal frequency. An iterative processing augmented by road-mask post-processing using TIGER data yields refined CDL maps (2017–2020) that show markedly reduced misclassification on high-resolution reference imagery and maintain strong county-level acreage agreement with NASS statistics. Kalichkin et al. \cite{kalichkin2021application} demonstrated how decision trees could forecast spring wheat yield in the Western Siberian forest-steppe using minimal public data. They compiled 2001–2019 observational records by incorporating qualitative factors (management intensity, crop rotation order, soil-tillage system) and quantitative agro-meteorological inputs (sums of active temperatures and precipitation during key growth periods). By using both CART and conditional-inference tree algorithms, they trained on 80\% of the data and tested on 20\% where the best CART model achieves a training MAE of 3.455 q/ha ($R^{2}$ = 0.895) and a test MAE of 4.446 q/ha ($R^{2}$ = 0.811), producing a set of clear, rule-based solution that link agronomic practices and weather to expected yield outcomes.

\noindent \textbf{K-Means.} K-Means is an unsupervised clustering algorithm that groups data into K clusters by iteratively assigning points to the nearest cluster centroid and then updating centroids to the mean of assigned points. It excels in identifying k number of groups in data. Aldino et al. \cite{aldino2021implementation} apply K-Means clustering to group sub-districts in South Lampung Regency by their corn-planting feasibility by using two years of harvest area and production data from the Central Bureau of Statistics. They first preprocess the data to extract key features like harvested area (ha) and production (tons) for 2018–2019 and then set k = 2 to distinguish high and low productivity clusters. To begin with, the initial centroids are chosen based on extreme values for area and yield, and euclidean distances are computed for each sub-district to assign it to the nearest cluster. The algorithm is executed in an iterative manner by recalculating centroids as the mean of assigned points until stabilization, which then results in a clear division of regions that require targeted agronomic support. Sethy et al. \cite{sethy2017detection} proposed a 3-means clustering segmentation method to detect and segment diseased regions on rice leaves. By starting with RGB leaf images captured in the field, they converted these to $L, a, b$ color space for better discrimination of color features. Afterward, the K-Means algorithm with $k = 3$ was applied to partition pixels into healthy tissue, disease symptoms, and background, which they merged to separate the defective leaf areas and calculate disease severity by measuring the proportion of diseased pixels. Their approach demonstrates fast computation and effective segmentation of discolorations and lesions and offers a low-cost tool for early rice disease monitoring.

Ota et al. \cite{ota2022weed} addressed the challenge of robust crop versus weed detection in heavily infested fields by combining crop-row detection from depth data with K-Means classification in RGB-D images. In the training phase, they extracted plant blobs that were connected to the detected crop row. It was identified as the fitting ground-surface wave patterns in the depth map to ensure a higher proportion of crop examples. They compute features like mean RGB, depth, and blob shape metrics for these blobs and run K-Means ($k = 2$) to learn centroids for crop and weed classes. During detection, blobs within the crop-row region are classified by nearest-centroid distance, while off-row blobs are labeled as weeds. This hybrid method maintains high accuracy even when weeds dominate the scene. Moreover, Yuan et al. \cite{yuan2024research} integrated K-Means cluster analysis with a linear programming framework to optimize multi-crop planting strategies under both deterministic and uncertain conditions. They first model resource constraints, including land area, cropping seasons, crop rotation rules, and economic returns in a mixed-integer linear program that decides crop allocations per plot. Then, they introduced robust optimization constructs by defining uncertainty sets around key variables to handle parameter uncertainty for price, yield, and cost fluctuations. Finally, they refined the strategy by clustering crops with similar substitutability and complementarity profiles via K-Means, thereby adjusting the objective function to balance economic efficiency and risk. This combined approach yields planting plans that maximize expected farmer income while safeguarding against worst-case scenarios.

\noindent \textbf{Case Based Reasoning (CBR).} CBR solves new problems by adapting solutions from the most similar past cases stored in a case library. Rather than building a global model, CBR retrieves, reuses, and revises previous examples to construct solutions which makes it particularly well-suited for domains where expert examples are plentiful and problem contexts repeat. Lawanna and Suwannayod \cite{lawanna2025agricultural} introduced a conceptual framework named CBRIoT that combines CBR with Internet-of-Things (IoT) data streams to support agricultural decision-making in real-time. Their system uses continuous sensor feeds like soil moisture probes, weather stations, and drone imagery into a central knowledge base of past cases i.e., environmental contexts paired with successful interventions. When this system is presented with a new scenario, it retrieves the most similar historical cases, adapts their recommended actions, and delivers tailored guidance (e.g., irrigation schedules and pest controls). This system has been evaluated on multiple datasets like Agriculture-Vision, Extended Agriculture-Vision, and Smart Agriculture, where CBRIoT outperformed traditional Vegetation Index Models, machine-learning classifiers, geostatistical interpolators, and rule-based systems in decision-making accuracy, responsiveness, and scalability. 

Moosavi et al. \cite{moosavi2025development} developed a Fuzzy Case-Based Reasoning (FCBR) decision-support system aimed at optimizing water management in smart agriculture. Their architecture follows the classic CBR \textit{5R} cycle which includes representation, retrieval, reuse, revision, and retention, but it augments it with fuzzy logic to handle uncertainty in sensor data like soil moisture, plant water stress, and microclimate variables. Here, the new real-time measurements are fuzzified into linguistic categories (e.g., low, medium, high moisture), and similar past cases are retrieved based on a composite fuzzy-similarity metric. The retrieved irrigation strategies are then adapted via fuzzy rules to the current conditions and tested, and the outcome is retained to continuously enrich the case base. The usage of this method in fields demonstrated significant reductions in water usage, lower irrigation costs, and improved crop productivity compared to non-adaptive approaches.

Zhai et al. \cite{zhai2020applying} applied CBR enhanced with a learning-based adaptation strategy to precision irrigation scheduling in grape vineyards. Rather than relying on fully specified mathematical models, their system represents each irrigation event as a feature vector of meteorological inputs and past irrigation amounts, retrieves the closest matching past events, and then revises the proposed irrigation volume by learning how similar past cases were adjusted. Specifically, they frame revision as a \textit{situation–action} pair, i.e., differences between the new and retrieved cases are quantified, and adaptation knowledge, which is extracted from how prior solutions were modified, is applied to refine the irrigation plan. In experiments, their approach predicted reference evapotranspiration and irrigation volumes with deviations of only ~5.4\% and ~7.9\% respectively, illustrating robust performance even with limited data

Windi et al. \cite{windi2023application} present a CBR-based expert system for early diagnosis of rice plant diseases including leaf blight, tungro, rice blast, dwarf virus, etc. Their framework captured each scenario as a case comprising symptom attributes like leaf lesion patterns, plant vigor, and the corresponding treatment. After receiving a new symptom description, the system retrieves the most similar stored cases using a weighted-attribute similarity function, reuses the associated treatments as provisional solutions, revises them through user feedback or further symptom checks, and then retains successful new cases to expand its knowledge. By emulating expert reasoning and continuously updating its case base, the system enables farmers to identify and treat rice diseases more accurately and promptly

\subsubsection{CNNs}
Convolutional Neural Networks (CNNs) have transformed how we analyze crop images in agriculture. This section discusses how different CNN methods help identify crop types, detect diseases, and estimate yields. 

\noindent \textbf{VGG-ICNN.}
Thakur et al. \cite{thakur2023vgg} introduces a novel lightweight CNN model named VGG-ICNN which is designed for identifying crop diseases using plant-leaf images. It incorporates the feature extraction capabilities of VGG16 and multi-scale processing from GoogleNet Inception v7 \cite{szegedy2015going}, making it robust yet efficient with around 6 million parameters. VGG-ICNN is evaluated on five diverse datasets including PlantVillage\cite{mohanty2016using}, Maize \cite{chen2020using}, Apple \cite{thapa2020plant}, Rice \cite{chen2020using}, and Embrapa \cite{barbedo2018annotated}, VGG-ICNN demonstrated superior accuracy and consistency across different crop species and conditions, achieving up to 99.16\% accuracy on the PlantVillage dataset. This model outperforms several deep learning models including EfficientNet \cite{tan2019efficientnet}, MobileNet \cite{howard2017mobilenets}, and GhostNet \cite{han2020ghostnet}.

\noindent \textbf{Geo-CBAM-CNN.}
The Geo-CBAM-CNN \cite{wang2021new} model is an advanced attention-based CNN that incorporates geographical information within the Convolutional Block Attention Module (CBAM) to specifically address the geographic heterogeneity of crop phenology. The integration of spatial and spectral attention mechanisms allows the model to focus more effectively on relevant features which improves the classification accuracy significantly. This model is validated on four main crops, including corn, cotton, soybean, and winter wheat, over six U.S. counties (Randolph, Mississippi, Haskell, Traill, Sherman, Decatur), where the model achieved an overall accuracy of 97.82\% and high scores in other metrics such as the Kappa coefficient and Macro-average F1 score. This model represents a significant advancement in precision agriculture, providing a powerful tool for crop classification using time-series satellite imagery, optimized to handle the spatial variability in crop spectral signatures.

\noindent \textbf{OMFA-CNN.}
OMFA-CNN \cite{kumar2024advanced} model is developed for high performance in detecting multiple potato diseases, which utilizes a deep learning framework that includes three convolutional layers, three max-pooling layers, and two fully interconnected layers. This configuration allows for a detailed analysis of potato leaf images to facilitate early and accurate disease detection. The model is evaluated on the PlantVillage \cite{mohanty2016using} dataset with seven common potato diseases including early blight, late blight, blackleg, potato virus y, potato cyst nematode, potato wart diseases, and fusarium dry rot.

\noindent \textbf{T-CNN.}
The Trilinear Convolutional Neural Network (T-CNN) \cite{wang2021t} model enhances the feature extraction capabilities of traditional CNNs through the integration of bilinear pooling. This approach allows the model to capture complex interactions between features at different scales, which is important for differentiating between subtle variations in disease symptoms on plant leaves. The trilinear component refers to the processing of image data through three distinct pathways within the CNN, where each optimizes different aspects of the feature extraction process such as color, texture, and shape before merging them for final classification. Furthermore, the bilinear pooling effectively combines features from two CNN streams, multiplying them in a pairwise manner and then applying pooling operations. This enhances the model's ability to focus on critical features while ignoring irrelevant background noise. The model was tested on real-world and controlled environment datasets, showing improved identification accuracies of 84.11\% and 99.7\% for crops and diseases in natural settings, respectively.

\noindent \textbf{CCDF.}
Khan et al. \cite{khan2018ccdf} developed a correlation coefficient and deep features (CCDF) based framework that begins with enhancing the contrast of the input images to better highlight diseased areas. This is followed by a segmentation technique based on the correlation coefficient, which meticulously differentiates diseased regions from healthy backgrounds. For the recognition phase, the system employs two deep pre-trained models, VGG16 \cite{simonyan2014very} and Caffe AlexNet \cite{krizhevsky2012imagenet}, renowned for their deep feature extraction capabilities. These models extract a comprehensive set of features, which are then fused and refined through a genetic algorithm to select the most impactful features for disease classification. This method ensures a high level of accuracy by employing a multi-class SVM for the final classification step, showcasing the system’s ability to integrate advanced image processing and deep learning techniques for practical agricultural applications. This framework achieves a high classification accuracy of 98.60\% on the PlantVillage and CASC-IFW datasets.

\noindent \textbf{SkipResNet.}
SkipResNet \cite{hu2024skipresnet} is a modified version of the ResNet \cite{he2016deep} architecture which is designed specifically to enhance crop and weed recognition accuracy. This model introduces multiple input paths and skip connections at various layers of the network which aims to preserve the integrity of input features throughout the depth of the network. These modifications address the common problem of feature dilution in deep neural networks by allowing the original input data to skip certain layers and merge directly with deeper layers' outputs. Such a configuration is often helpful in maintaining essential details that are important for accurate classification. The network employs several path selection algorithms to optimize the handling of these multiple streams, which ensures that the most effective features are used for learning. Hence, by incorporating path selection algorithms that optimize the learning process, SkipResNet achieves superior performance on plant seedling \cite{giselsson2017public} and weed-corn \cite{jiang2020cnn} datasets with an accuracy of 95.07\% and 99.24\%, respectively.

\subsubsection{Vision Transformers (ViTs)}

\noindent \textbf{ViT-SmartAgri.}
ViT-SmartAgri \cite{barman2024vit} applies the Vision Transformer (ViT) \cite{dosovitskiy2020image} to the task of detecting plant diseases directly from smartphone-captured images by focusing particularly on tomatoes. The model is evaluated on a comprehensive dataset of 10,010 images of tomato leaves which covers 10 distinct disease classes along with healthy class. This dataset was employed to train both the ViT and the Inception V3 \cite{szegedy2016rethinking} models to assess and compare their effectiveness in recognizing and classifying various tomato diseases. The ViT model reliance on self-attention mechanisms was able to outperform Inception V3 by achieving a higher testing accuracy of 90.99\%. This effectiveness underscores the model's potential for deployment in a smartphone application which enables on-the-spot disease diagnosis in outdoor field conditions.

\noindent \textbf{DVTXAI.}
The DVTXAI \cite{kamal2025dvtxai} model combines a deep vision transformer architecture with explainable AI (XAI) capabilities to enhance the detection and diagnosis of plant diseases. This model is specifically designed for high accuracy and interpretability where the model utilizes the Plant Village \cite{mohanty2016using} dataset. As the dataset contains a variety of plant species and disease types, the model is evaluated by focusing particularly on diseases affecting potatoes and tomatoes leaves. What sets DVTXAI apart is its integration of XAI techniques, particularly SHapley Additive exPlanations (SHAP) values, which provide insights into the model's decision-making processes by highlighting feature contributions. The model is compared with other architectures like ViT-SmartAgri, ViT, MaxViT and others.

\noindent \textbf{HyperSFormer.}
HyperSFormer \cite{xie2023hypersformer} represents a significant advancement in hyperspectral image processing for agriculture by using a transformer-based architecture which is optimized for semantic segmentation tasks. This model reconfigures the standard SegFormer by incorporating an enhanced Swin Transformer \cite{liu2021swin} as its encoder and maintaining the original decoder. It leverages the unique properties of hyperspectral images, which capture a wide spectrum of light not visible to the human eye, making them particularly useful for detailed crop analysis. The model was evaluated on three public hyperspectral datasets, demonstrating its robustness across different environmental conditions and its superior performance in dealing with imbalanced data and complex negative samples. The introduction of the Hyper Patch Embedding (HPE) module enables the model to extract detailed spectral and spatial details, hence improving the segmentation accuracy significantly. Additionally, the Adaptive Min Log Sampling (AMLS) strategy and the novel loss function integrating dice loss and focal loss further helps to optimize the training process by addressing challenges related to imbalance dataset and enhancing model sensitivity to rare but critical features.

\noindent \textbf{MMST-ViT.}
The MMST-ViT model \cite{lin2023mmst} integrates a Multi-Modal, Spatial, and Temporal Transformer to address the complexities of crop yield prediction under varying climatic conditions. This model utilizes a unique combination of visual remote sensing data and numerical meteorological data to predict crop yields at the county level across the United States. The Multi-Modal Transformer utilizes both data types to model short-term meteorological effects on crop growth, while the Spatial Transformer captures high-resolution spatial dependencies among counties for precise agricultural tracking. The Temporal Transformer is designed to understand long-term climate effects on crop patterns. Moreover, a novel aspect of MMST-ViT is its multi-modal contrastive learning technique, which pre-trains the model without extensive human supervision and enhances its ability to generalize across different datasets and conditions. The model has been extensively tested on over 200 counties and has shown superior performance over traditional models with metrics such as Root Mean Square Error (RMSE), R-squared, and Pearson Correlation Coefficient.

\noindent \textbf{PlantViT.}
PlantViT \cite{thakur2021vision} is a hybrid model combining a CNN and a Vision Transformer to detect plant diseases from images. This model processes images through three convolutional layers to generate feature embeddings, which are then passed through a multi-head attention module. PlantViT has been evaluated on two large-scale datasets named PlantVillage \cite{mohanty2016using} and Embrapa \cite{barbedo2018annotated}, achieving accuracies of 98.61\% and 87.87\%, respectively. The model's architecture allows it to effectively learn from image data by enhancing its ability to focus on relevant features for disease detection. The model has a very lightweight design with only 0.4 million trainable parameters that outperform the other techniques, and thus making it suitable for deployment in resource-constrained environments.

\noindent \textbf{Twin-ViT.}
Padshetty \cite{padshetty2025novel} introduces a Twin Vision Transformer (Twin-ViT) that utilizes deformable attention to enhance the classification of crop diseases. The model incorporates a Depthwise Separable Visual Geometry Group (DS-VGG16) technique for initial feature extraction, which reduces computational complexity and enhances feature extraction efficiency. The Twin-ViT model integrates a transformer encoder with an Edge Aware Enhancement Module (EAEM) and a classifier head that includes a Squeeze and Excitation block to refine feature representation and improve classification accuracy. This model is particularly effective in handling the variability of disease representations across different crops. The use of deformable attention allows the model to adaptively focus on the most relevant parts of the image and making the process compute efficient.

\noindent \textbf{MSCVT.}
The MSCVT model \cite{zhu2023crop} combines multiscale convolution techniques with a vision transformer architecture to enhance crop disease identification process. This hybrid model incorporates a multiscale self-attention module that utilizes both convolutional and self-attention mechanisms. By doing so, it effectively captures both local and global image features. The model is structured around a modified ResNet architecture with inserted self-attention modules across multiple stages, thus allowing for detailed feature processing at different scales. The key to its efficiency is the use of inverted residual blocks, which reduces parameter count while maintaining model depth and complexity. This model was tested on two prominent datasets named the PlantVillage \cite{mohanty2016using} dataset and the privately collected Apple Leaf Pathology dataset, therefore, achieving recognition accuracies of 99.86\% and 97.50\%, respectively. These results significantly outperform traditional CNN models like VGG16 \cite{simonyan2014very}, MobileNetV1 \cite{howard2017mobilenets}, ResNet18 \cite{he2016deep}, and others.

\noindent \textbf{SeedViT.}
SeedViT \cite{chen2022vision} adapts the vision transformer architecture for the task of classifying maize seed quality. Unlike the standard ViT, which requires large datasets to train effectively, SeedViT introduces modifications that allow it to work well with smaller datasets, comprising only 2,500 images. This model utilizes self-attention mechanisms to bypass the need for convolutional layers focusing instead on global dependencies within the image data. SeedViT has been compared against traditional deep learning approaches like CNNs and machine learning algorithms such as SVM and KNN, where it demonstrated superior performance with accuracies reaching up to 97.6\%. The novel visualization of attention maps underscores SeedViT's potential to improve maize seed classification by reducing reliance on manual inspection and enhancing objective quality assessments.

\noindent \textbf{PMVT.}
The plant-based MobileViT (PMVT) \cite{li2023pmvt} focuses on the operational constraints of mobile devices and balancing the computational efficiency. The model integrates a mobile vision transformer model with convolutional block attention modules (CBAM) to enhance focus on critical features while managing resource use efficiently. By employing a larger 7x7 convolution kernel within the inverted residual blocks, PMVT better captures long-range dependencies among image features. The model is extensively tested across multiple datasets including wheat, coffee, and rice where it achieved best accuracies by outperforming established lightweight models like MobileNetV3 and heavyweight models. This model not only demonstrates its efficacy in lab settings but also through a practical application with a plant disease diagnosis app.

\subsubsection{Foundation Models for Crop Farming}

\noindent \textbf{AgriCLIP.}
AgriCLIP \cite{nawaz2024agriclip} introduces a vision-language model tailored broadly for agriculture applications, including crop farming. The model addresses the domain shift issue in the standard CLIP model trained originally on general computer vision images. This model utilizes a new large-scale dataset named ALive, comprising approximately 600,000 image-text pairs across diverse agricultural categories. The innovative aspect of AgriCLIP lies in its dual training approach, combining image-text contrastive learning and self-supervised learning to enhance both global semantic and local fine-grained feature extraction. This method significantly improves zero-shot classification performance in agriculture-specific tasks by incorporating domain-specialized training that refines the model’s ability to distinguish subtle differences crucial for agricultural applications like plant disease classification, nutrients deficieny identification, fruits leaf classification, and many others.

\noindent \textbf{AgroGPT.}
AgroGPT \cite{awais2024agrogpt} addresses the gap in domain-specific conversational models for agriculture by introducing expert-level instruction-tuning data from vision-only agricultural datasets. This model is trained on a new 70,000-sample dataset called AgroInstruct, which excels in complex crops and plant dialogues and provides in-depth insights into different crop images. It uses large language models to generate text based on class-specific information and agricultural data which enhance its ability to provide precise and contextually relevant responses. AgroGPT demonstrates significant improvements in identifying and discussing detailed crop farming concepts compared to general-purpose models which make it a best option for applications that require expert agricultural knowledge.

\noindent \textbf{DiffusionSat.}
DiffusionSat \cite{khanna2023diffusionsat} pioneers a generative foundation model designed explicitly for satellite imagery using diffusion models that is adapted to handle the unique characteristics of remote sensing data that is unique in comparison to the ground-level data. This model supports high-resolution image generation, superresolution, and temporal prediction tasks by incorporating metadata like geolocation directly into the generation process. It is trained on a diverse set of publicly available satellite datasets. The model leverages metadata-rich inputs to produce high-quality satellite images and perform complex generative tasks such as in-painting and temporal generation. Its advanced conditioning methods allow it to effectively simulate future scenarios or reconstruct past events from satellite data.

\noindent \textbf{TasselELANet.}
TasselELANet \cite{ye2024vision} is designed specifically for agricultural vision tasks, focusing on plant detection and counting using a highly efficient convolutional network architecture. It utilizes an encoder with a 16-fold downsampling layer and a decoder with only 2 feature layers, significantly deviating from traditional deep network designs. At the core of TasselELANet is the Efficient Layer Aggregation Network (ELAN), which optimizes gradient propagation paths through cross-stage fusion strategy. This setup enhances both the learning capabilities and inference speed, effectively handling the challenges of agricultural image recognition such as variable plant sizes, occlusions, and complex growth environments. The model was tested on three challenging plant datasets named maize tassels, wheat ears, and rice panels where it achieved high accuracy and determination coefficients.

\noindent \textbf{Leaf-Only SAM.}
Leaf Only SAM \cite{williams2024leaf} employs the Segment Anything Model (SAM) for leaf segmentation in potato plants. This approach uses a zero-shot approach
and uses a series of post-processing steps to segment leaf area from the background region. The performance of Leaf Only SAM was compared to a fine-tuned Mask R-CNN \cite{he2017mask} model on a novel potato leaf dataset, where SAM achieved average recall and precision scores but did not outperform the Mask R-CNN. The method demonstrates SAM's potential as a zero-shot classifier with the addition of post-processing steps which offers a viable solution for applications where annotated datasets are scarce.

\noindent \textbf{BioCLIP.}
BioCLIP \cite{stevens2024bioclip} is a vision foundation model explicitly designed to understand the rich, hierarchical structure of biological taxonomy. It is pre-trained on the TREEOFLIFE-10M \cite{stevens2024bioclip} dataset, which is a massive and ML-ready dataset of over 10 million images covering more than 450,000 species of plants, animals, and fungi (taxa), where each image is labeled with a flattened string that concatenates taxonomic ranks from kingdom down to species. By pairing images with these comprehensive taxonomic names and applying a CLIP-style \cite{radford2021learning} contrastive learning objective, BioCLIP learns to align visual features with their biological context, enabling the model to generalize naturally to unseen species in the natural world using higher-level groupings like genus or family.

In practice, BioCLIP supports zero and few-shot classification across ten diverse fine-grained biology benchmarks, including a newly curated Rare Species dataset drawn from IUCN red-list \cite{IUCNRedL61} taxa, by matching image embeddings to appropriate taxonomic names or common names provided at test time. This taxonomy-aware training strategy yields substantial gains over generalist CLIP \cite{radford2021learning} models, with BioCLIP achieving an average absolute improvement of 17 percent in zero-shot tasks and 16 percent in few-shot tasks. The intrinsic analyses further confirm that BioCLIP’s learned embeddings form clearer, more fine-grained clusters that mirror taxonomic ranks, demonstrating its ability to internalize and exploit the tree-of-life hierarchy for robust biological image understanding.

\noindent \textbf{ITLMLP.}
Image-Text-Label Multi-Modal Language Pretraining (ITLMLP) \cite{cao2023cucumber} model introduces a novel approach to cucumber disease recognition by using a multi-modal language model that integrates image, text, and label information. This model addresses the challenges of small sample sizes typical in agricultural applications by utilizing multi-modal contrastive learning, image self-supervised learning, and label information. The method was tested on a diverse dataset of cucumber diseases where it achieved comparative recognition accuracy of 94.84\%. The ITLMLP framework effectively measures the sample distance in a unified image-text-label space which is why it enhances both the performance and generalizability of the model across different datasets and agricultural environments.

\noindent \textbf{PlantCaFo.}
The PlantCaFo \cite{jiang2025plantcafo} is a few-shot learning approach utilizing foundation models like CLIP \cite{radford2021learning} and DINO \cite{caron2021emerging} for plant disease recognition, which highlights the efficiency of using pre-trained models in new domains with limited data availability. This model incorporates a lightweight dilated contextual adapter (DCon-Adapter) and a weight decomposition matrix (WDM) to adapt and fine-tune foundational model knowledge for specific few-shot learning tasks in agriculture. The model was extensively evaluated on the PlantVillage \cite{mohanty2016using} dataset and an out-of-distribution dataset, showing impressive performance improvements over traditional methods like Tip-Adapter. The model achieves an accuracy of 93.53\% in a "38-way 16-shot" scenario, and 6.80\% accuracy improvement over the baseline on the Cassava dataset.

\noindent \textbf{DINOv2.}
Gustineli et al. \cite{gustineli2024multi} present a novel approach to multi-label plant species classification using a self-supervised DINOv2 \cite{oquab2023dinov2} model for the PlantCLEF 2024 challenge. The approach uses transfer learning to extract generalized feature embeddings from a substantial dataset comprising 1.4 million images spanning 7,800 species. These embeddings are utilized to train classifiers that can predict multiple plant species within a single image. The methodology involves preprocessing the images into a manageable format, extracting embeddings with both base and fine-tuned DINOv2 models, and training linear classifiers on these embeddings. The classifiers are then used to make predictions on a test set using a grid-based image prediction method for multi-label classification.

\noindent \textbf{VLM.}
Dong et al. \cite{dong2024visual} present an approach to improve anomaly detection in plant disease recognition using a multi-modal model that integrates visual information with vision-language models like CLIP \cite{radford2021learning}. This method specifically addresses the challenge of identifying unknown disease classes that are not present in the training dataset, which traditional text-concept-guided models often struggle with due to their reliance on textual prompts. The model's design incorporates visual features directly into the anomaly detection process, which enhances its ability to distinguish between the variations that signify diseases outside the known classes. By calculating uncertainty scores through a feature-matching technique, the model evaluates test sample features against known training samples. This visual guidance helps overcome the limitations of text-only models in the fine-grained task of plant disease detection, where visual symptoms are as essential for accurate classification. The authors also include the adaptation of fine-tuning paradigms such as Context Optimization (CoOp) \cite{zhou2022learning} and Conditional Context Optimization (CoCoOp) \cite{zhou2022conditional}, along with Visual Prompt Tuning (VPT) \cite{jia2022visual} and Vision-Language Prompt Tuning (VLPT) \cite{shen2024multitask}. These adaptations are critical for integrating visual prompts more deeply within the model. The experimental results on PlantVillage \cite{mohanty2016using} dataset show the model's enhanced performance with significant improvements noted in classifying between known and unknown classes.

\section{AI for Fisheries}
\label{sec:fish}
The fisheries domain is an essential part of Agriculture because it is a significant source of food and income for millions of people around the globe. It also plays a vital role in ensuring global food security and preserving biodiversity \cite{alsaleh2023role} by maintaining the balance of species and encouraging the conservation of habitats like coral reefs and mangroves.
The complexity of aquatic ecosystems is gradually growing, and therefore, the use of AI in fisheries management has become essential. This section covers the challenges and advancements made in the fisheries domain under the umbrella of AI methodologies.

\subsection{Machine Learning Tasks in Fisheries}
\begin{figure}[hbtp]
    \centering
    \includegraphics[width=\columnwidth]{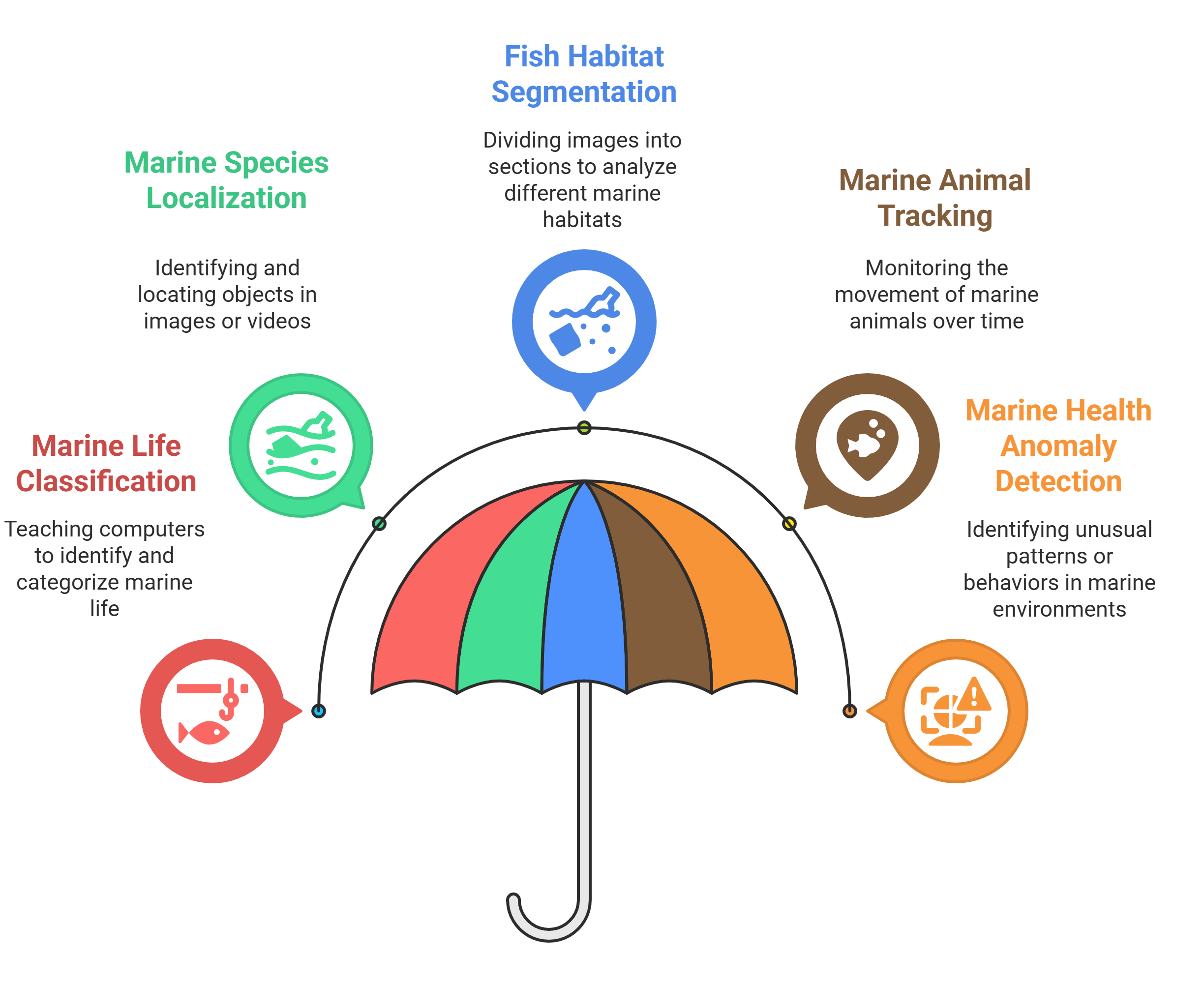}
    \caption{Illustration of various AI tasks in marine life analysis, including Object Detection for identifying and locating marine creatures, Image Classification for categorizing species, Image Segmentation for detailed analysis, Object Tracking to monitor movement over time, and Anomaly Detection for spotting unusual behaviors or patterns.}
    \label{fig:fisheries-tasks}
\end{figure}

Fisheries rely on various data‐driven tasks to maintain sustainable practices and protect marine ecosystems. At a broad level, the goal of these tasks includes species identification, biodiversity tracking, illegal catch prevention, and seafood authenticity certification \cite{tejaswini2024automatic,Kuhn2024kl} as presented in Fig.~\ref{fig:fisheries-tasks}. Additionally, it includes fish population monitoring, where analysts examine fish aggregations to count their numbers and overall well-being \cite{li2021automatic} along with the ecosystem assessment, which involves evaluating the health of broader marine environments to carry out responsible management measures \cite{yang2021deep, gladju2022applications}. Several specific AI-oriented tasks, discussed next, offer more precise analytical capabilities in Fisheries:

\noindent \textbf{Marine Life Classification. \textcolor{red}{(IC)}} This task involves the identification of different types of fish or other marine life captured in underwater images. For example, it includes the classification of different species, such as salmon, trout, and tuna, based on their visual characteristics. In general, it is critical for recognizing and categorizing underwater objects, particularly fish species or any signs of environmental disturbances, as they convey relevant details about the population tracking and habitat assessment \cite{elmezain2025advancing}. 
For this purpose, many recent studies have employed deep learning (DL) models trained on diverse image datasets to capture subtle shape and texture features \cite{zhao2024review2}. For example, Jareño et al. \cite{jareno2024automatic} demonstrated that transfer learning with models such as ResNet152V2 \cite{he2016identity} and EfficientNetV2L \cite{tan2021efficientnetv2} can improve accuracy in fish species identification.  Peddina et al. \cite{peddina2024optimized} introduced the Chimp-based Google Deep Network (CbGDNet), achieving 99.16\% accuracy in fish classification, which is on par (if not better) with the human knowledge of fish species. Moreover, Hybrid models that combine CNNs with conventional machine learning (ML) algorithms like SVMs and decision trees also perform well in handling the complexities of underwater images \cite{li2022recent}. These hybrid models benefit from the robust feature extraction of deep learning and the effective data separation capabilities of traditional algorithms, which are helpful in handling the non-linear and complex nature of underwater images.

\noindent \textbf{Marine Specie Localization. \textcolor{green}{(OD)}} It is a step beyond classification and is used to identify and pinpoint the location of fish or other objects within underwater images or videos. It helps recognize different marine species by providing a bounding box around each one to indicate their exact position in the frame. This is fundamental for tracking fish movements, spotting invasive species, and studying underwater habitats \cite{lopez2021automatic}. However, overlapping fish, coral occlusions, murky water, and constantly changing light often complicate such detections \cite{fu2023rethinking,chen2024underwater,cui2025fish}.
For this purpose, many researchers proposed different solutions ranging from traditional ML to advanced DL approaches. Specifically, DL methods, which are particularly specialized CNNs, help overcome these issues by learning from diverse datasets to identify marine life in varied conditions \cite{liu2024deepseanet,li2023deep,siri2024enhanced,elmezain2025advancing}. Li et al. \cite{li2022fish} proposed a Rotated Object Detection (ROD) technique that uses rotating bounding boxes to adapt to fish orientations and detect fish faces in underwater conditions. By adopting the rotating bounding boxes approach, more accurate and less redundant detections were observed, as the simple bounding box detection may not be able to cater to the varying fish movements. Additionally, the FE-Det framework by Luo et al. \cite{luo2024fe}, and the work by Shin et al. \cite{rani2024novel} address challenges like image distortions and edge artifacts (a common issue in fisheye lens cameras used in marine studies), ensuring reliable detection across the entire image.

\noindent \textbf{Fish Habitat Segmentation. \textcolor{blue}{(IS)}} This task involves dividing an underwater image into different sections to better identify and analyze individual elements, such as separating fish from coral or differentiating between various types of seaweed. 
This technique is important for habitat studies, fish size estimation, fish type identification, stock assessment for sustainable yield, fish disease detection, and assessing the health of marine habitats \cite{petrellis2021measurement,li2022advanced,khabusi2024enhanced}. 
However, such segmentation can be affected by conditions like murky water and shifting light at different depths, which blur the boundaries and make segmentation challenging \cite{zhang2024rusnet}. To reduce these complications, many studies used context-awareness, which analyzes the overall spatial relationships in an image rather than focusing on isolated elements. Utilizing this approach proves particularly useful in crowded marine scenes where either the objects are overlapped with each other, or they are closely packed \cite{zuo2025improving}. Preprocessing is another critical step for refined segmentation. As shown by Awalludin et al. \cite{awalludin2020review}, techniques like noise reduction and contrast enhancement, using thresholding and edge detection, prepare images by clearly distinguishing marine animals from their background, which is especially important in environments with low visibility.

Moreover, deep learning techniques have significantly improved the segmentation outcome. Garcia et al. \cite{garcia2020automatic} demonstrated the use of Mask R-CNN for instance segmentation, effectively handling overlapping objects and varied textures. This method allows for precise segmentation of individual fish, which aids in the accurate measurement of fish sizes and behaviors and is also a key factor in both regulatory compliance and conservation efforts. Additionally, the texture analysis has also been incorporated to further enhance segmentation accuracy by examining surface details. For example, the unique texture of a fish's scales can help differentiate species and even identify signs of diseases \cite{soto2019diagnosis}.

\noindent \textbf{Marine Animal Tracking. \textcolor{brown}{(OT)}} It is used to monitor the movement of marine animals over time within a video. It can be useful for behavioral studies, tracking the migration patterns of tagged fish, or monitoring fish activity in aquaculture tanks like aquariums. Such tracking of fish or other underwater animals also assists in making broader ecological or management decisions. Some factors like rapid fish motion, complex seabed topography, and variable underwater visibility make this task extremely challenging \cite{gupta2021dftnet}. To mitigate these challenges, variants of the Kalman filter \cite{khodarahmi2023review} and DeepSORT \cite{wojke2017simple}, which were originally designed for simple tracking conditions, have been adapted for the underwater context. By combining spatial information (where the fish are) with temporal data (how they move over time), these methods accurately predict fish trajectories even under dynamic conditions \cite{liu2024fishtrack}. For instance, Lopez-Marcano et al. \cite{lopez2021automatic} demonstrated that automated computer vision techniques can significantly reduce manual labor while reliably tracking fish in coastal environments. Meanwhile, sonar datasets like the Caltech Fish Counting Dataset \cite{liu2024fishmot} provide low-visibility and high-density scenarios for the development of real-time fish counting and biomass estimation. In addition, integrating data from multiple sensors such as visual, infrared, and sonar helped overcome common underwater issues and improved tracking accuracy \cite{heshmat2025underwater}.

\noindent \textbf{Anomaly Detection for Marine Health. \textcolor{orange}{(Others)}} Similar to other sectors of Agriculture, in Fisheries it is also crucial to identify unusual patterns or behaviors in underwater imagery, such as detecting diseased fish, spotting invasive species, or identifying environmental stressors \cite{wang2020anomalous}. This helps in maintaining the health of marine ecosystems and managing fish populations effectively. It also focuses on uncovering unusual or suspicious patterns, often related to illegal fishing activities, unreported activities, unauthorized vessel movements, or sudden ecological disruptions. However, detecting anomalies in marine environments is challenging due to natural variability. Some of the factors, like seasonal shifts, water quality changes, and irregular human activity, can all affect marine life in ways that are hard to predict. Therefore, successful identification of normal variation and the actual threat is essential as the failure can either cause unnecessary alarm or allow serious incidents to go unnoticed \cite{lopes2021prediction}.

For this purpose, unsupervised techniques like autoencoders and isolation forests are used to learn what normal behavior looks like from historical data with no need for specific examples of anomalies and flag significant deviations as potential anomalies \cite{bedja2022smart}. But in case the data on specific irregularities is available, semi-supervised or supervised learning methods can be used for more precise detection. However, these techniques must be continually updated to adapt to new conditions or changes in marine behaviors. For example, Rodrigues et al. \cite{rodriguez2024identification} used Automatic Identification System (AIS) data to identify silence anomalies, that is, the period when vessels turn off their AIS signals, which may suggest an attempt to hide illegal activities. Their statistical framework helps differentiate between regular operational changes and deliberate manipulations, particularly near coasts and exclusive economic zones (EEZs), where illegal fishing is more likely to occur.

\subsection{Challenges in Aquatic Farming}

\begin{figure}[hbtp]
    \centering
    \includegraphics[width=\columnwidth]{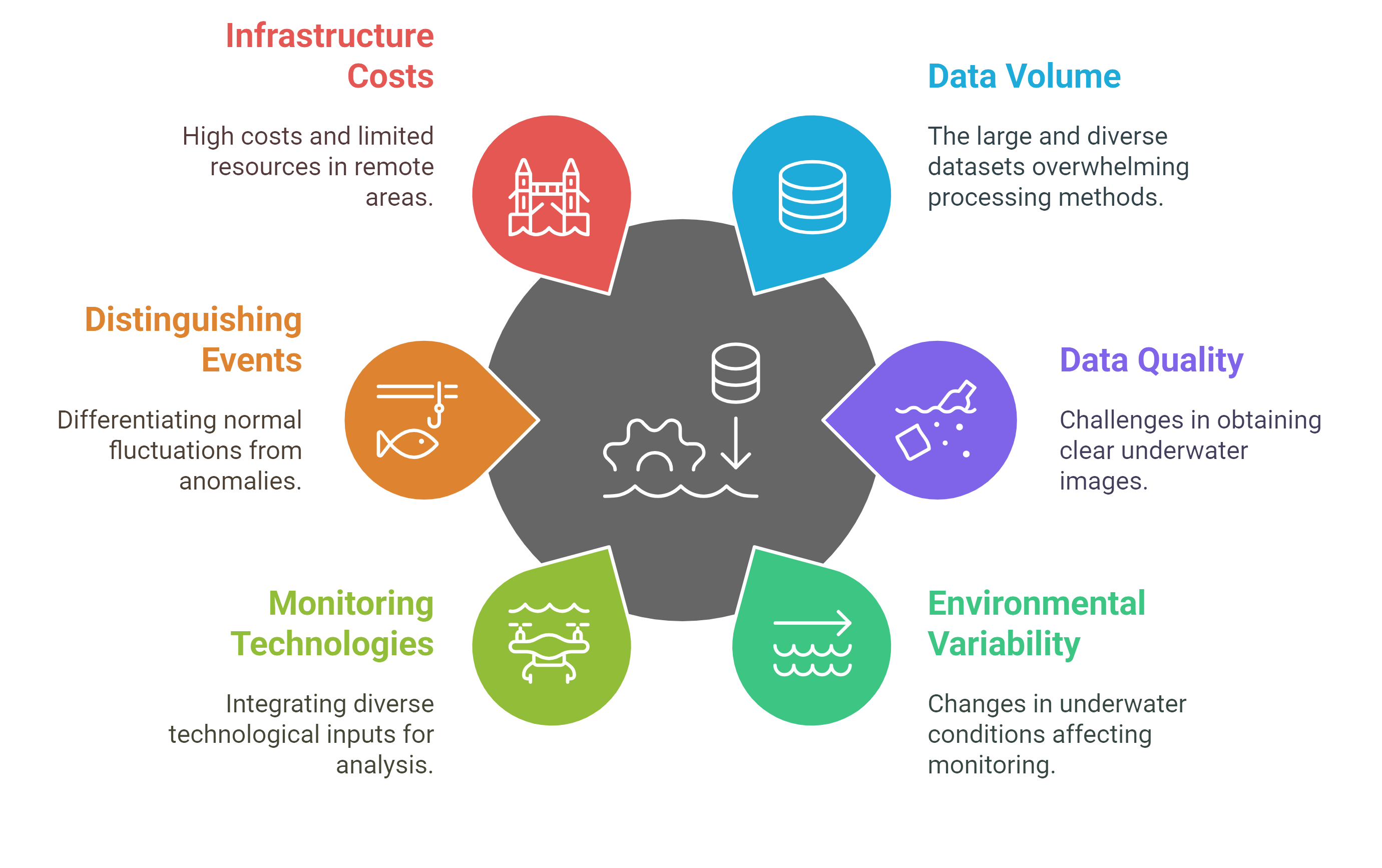}
    \caption{Illustration of major challenges such as high infrastructure costs, large data volumes, poor data quality, environmental changes, advanced monitoring needs, and identifying anomalies in fisheries management}
    \label{fig:challenges-fisheries-all}
\end{figure}

The fisheries domain often deals with complex and ever-changing marine environments, which present a range of technical and practical challenges as depicted in Fig.~\ref{fig:challenges-fisheries-all}. From vast data streams to underwater imaging issues, each challenge requires specialized solutions and adaptive approaches.

\noindent \textbf{Volume and Complexity of Data.} Recent trends in fisheries farming enable the generation of diverse types of data in various formats, including genetic markers, acoustic signals, and extensive underwater imagery or video clips \cite{jahanbakht2021internet}. Handling such large, diverse datasets can quickly overwhelm conventional data-processing methods \cite{yang2021deep}. Moreover, data is often captured using different sensors or systems that use distinct sampling rates or file formats, making it difficult to align and merge them into a cohesive dataset \cite{gladju2022applications}. These limitations underscore the need for advanced data-fusion techniques and robust storage infrastructures.

\noindent \textbf{Data Quality and Underwater Imaging Constraints.} Collecting clear images in underwater environments is inherently challenging as low visibility, moving light sources, and water turbidity can distort images or introduce noise, thereby reducing their reliability for tasks like species identification or fish population assessments. Although specialized convolutional neural networks (CNNs) and preprocessing methods (e.g., noise reduction, contrast adjustment) can overcome these issues, truly murky conditions continue to pose serious obstacles as they are not easily addressable \cite{rani2024novel}.

\noindent \textbf{Environmental Variability.} Underwater environments may shift dramatically due to currents, temperature differences, and seasonal changes. These fluctuations affect fish behavior, habitat conditions, and image characteristics, creating challenges for monitoring and analysis. AI models trained on one set of environmental conditions may fail or produce less accurate results in a different season, area, or depth. Consequently, there is a need for research into techniques that allow models to adapt or recalibrate, ensuring more robust performance across different habitats and time periods.

\noindent \textbf{Integrating Diverse Monitoring Technologies.} Many fisheries research programs combine data from underwater video feeds, sonar-based fish counters, and Automatic Identification System (AIS) logs that track vessel activities \cite{li2021automatic, lopez2021automatic, rodriguez2024identification}. While each source provides valuable insights, bringing them together raises significant technical challenges, such as aligning timestamps and reconciling differences in resolution or scale. However, when these data sources are implemented successfully, multi-sensor fusion can reveal patterns or anomalies that would otherwise go unnoticed by relying on a single data source. For example, cross-validating fish distribution (camera/sonar) against vessel behaviors (AIS) to better detect illegal fishing or environmental disturbances.

\noindent \textbf{Distinguishing Rare Events from Routine Fluctuations.} Marine environments naturally exhibit considerable variety, where fish may change their migration routes or feeding habits without any noticeable external factors. This makes it difficult to separate normal behavioral changes from genuinely abnormal or illegal activities \cite{lopes2021prediction}. Models must learn the difference between ordinary fluctuations and genuine anomalies that require further exploration. To deal with this problem, AI systems often rely on adaptive or semi-supervised learning methods, which can refine their internal definitions of `normal' based on newly observed data. This approach lowers the risk of false alarms while improving the chances of catching real irregularities, such as unregistered fishing or sudden ecological disruptions.

\noindent \textbf{Limited Infrastructure and High Operational Costs.} Many fishermen operate in remote regions or under resource constraints, which makes it challenging to deploy advanced monitoring equipment or maintain robust data-processing systems. High-capacity sensors, underwater cameras, and specialized AI hardware can be expensive to acquire and operate. As a result, smaller communities or less-funded projects may struggle to adopt the latest technology, limiting the overall impact of AI-based fisheries management.

\noindent \textbf{Legal and Regulatory Complexities.} Fisheries are subject to a range of national and international regulations designed to protect marine ecosystems and ensure fair resource allocation. Introducing AI-based tools into these frameworks can raise questions about data ownership, privacy, and accountability. For instance, if an AI model flags potential illegal activity, many disputes may arise over its reliability or legal standing. Navigating these regulatory uncertainties can slow technology adoption, particularly when multiple jurisdictions are involved.

\begin{table*}[htbp]
\centering
\caption{Summary of publicly available datasets used in fisheries domain. Each dataset is categorized based on different characteristics such as the number of classes, annotation type, dataset size, geographical focus, and supported tasks—such as Image Segmentation (IS), Object Detection (OD), and Image Classification (IC). }
\label{tab:datasets-fisheries}
\renewcommand{\arraystretch}{1.5} 
\resizebox{\textwidth}{!}{%
\begin{tabular}{@{}>{\raggedright\arraybackslash}p{2.5cm}>{\raggedright\arraybackslash}p{1.4cm}>{\raggedright\arraybackslash}p{3.5cm}>{\raggedright\arraybackslash}p{2.5cm}>{\raggedright\arraybackslash}p{2cm}>{\raggedright\arraybackslash}p{2cm}>{\raggedright\arraybackslash}p{1.8cm}>{\raggedright\arraybackslash}p{2.3cm}>{\raggedright\arraybackslash}p{1.5cm}@{}}
\toprule
\textbf{Dataset} & \textbf{Classes} & \textbf{Annotation} & \textbf{Dataset Size} & \textbf{Modality} & \textbf{Resolution} & \textbf{Geographical Focus} & \textbf{Tasks} & \textbf{Link} \\ \midrule

Deepfish & 2 & Class Labels, Masks & 40k labels, 3.2k point-level annotations, 620 masks & Video/Images & High & Australia & \textcolor{red}{IC}, \textcolor{blue}{IS}, \textcolor{green}{OD} & \href{https://github.com/alzayats/DeepFish}{Github} \\

Croatian Fish Dataset & 12 & Class Labels & 794 images & Images & Medium & Croatia & \textcolor{red}{IC} & \href{http://www.inf-cv.uni-jena.de/fine_grained_recognition.html/datasets}{Dataset} \\

Fish in seagrass habitats & 2 & Class Labels, Bounding Box, Masks & 4280 images & Images & High & Australia & \textcolor{red}{IC}, \textcolor{blue}{IS}, \textcolor{green}{OD} & \href{https://github.com/globalwetlands/luderick-seagrass}{Github} \\

Fish4Knowledge & 2 & Bounding Box & 27,230 images & Images & Low & Taiwan & \textcolor{red}{OD} & \href{https://github.com/Callmewuxin/fish4konwledge}{Github} \\

Fish-Pak & 6 & Class Labels & 915 & Images & High & Pakistan & \textcolor{red}{IC} & \href{https://data.mendeley.com/datasets/n3ydw29sbz/3}{Mendeley} \\

Fishes in the Wild & 2 & Bounding Boxes & 3167 images & Video/Images & High & South California & \textcolor{red}{IC}, \textcolor{green}{OD} & \href{https://www.st.nmfs.noaa.gov/aiasi/DataSets.html}{NOAA} \\

OzFish & 507 & Bounding Boxes & 80k labels, 45k annotations & Images & - & Australia & \textcolor{red}{IC} & \href{https://github.com/open-AIMS/ozfish}{Github} \\

QUT Fish Dataset & 468 & Class Labels & 3,960 images & Images & Low & Australia & \textcolor{red}{IC} & \href{https://www.dropbox.com/s/e2xya1pzr2tm9xr/QUT_fish_data.zip?dl=0}{Dropbox} \\

Whale Shark ID & 543 & Bounding Boxes & 7800 & Images & High & Australia & \textcolor{green}{OD} & \href{https://lila.science/datasets/whale-shark-id}{ILIA} \\

Large Scale Fish & 9 & Class Labels & 9000 images & Images & - & Turkey & \textcolor{red}{IC} & \href{https://www.kaggle.com/crowww/a-large-scale-fish-dataset}{Kaggle} \\

NCFM & 8 & Class Labels & ~16k images & Images & Medium & Australia & \textcolor{red}{IC} & \href{https://www.kaggle.com/c/the-nature-conservancy-fisheries-monitoring/data}{Kaggle} \\

Mugil liza sonar & 1 & Counting & 500 images & Video/Images & Medium & Brazil & \textcolor{red}{IC} & \href{https://doi.org/10.5281/zenodo.8384812}{Zenodo} \\

MSRB Dataset & NA & Synthesized Images & ~6k images & Images & Medium & - & \textcolor{orange}{Other} & \href{https://github.com/ychtanaka/marine-snow}{Github} \\

WildFish & 1000 & Class Labels & ~54k images & Images & Medium & Global & \textcolor{red}{IC} & \href{https://github.com/PeiqinZhuang/WildFish}{Github} \\

SUIM & 8 & Segmentation Masks & ~1.5k images & Images & Low - High & - & \textcolor{blue}{IS} & \href{https://github.com/xahidbuffon/SUIM}{Github} \\

VLM4Bio & - & 469k QA pairs & 30,000 images & Images & Medium & Global & \textcolor{orange}{Other} & \href{https://www.imageclef.org/}{ImageCLEF} \\

FishVista & 1900 & Class Labels, Masks & 60,000 images & Images & Medium & Multi & \textcolor{orange}{Other} & \href{https://github.com/Imageomics/Fish-Vista}{Github} \\

DZPeru - Multiple Fish Datasets & Multiple & Class Labels, Bounding Boxes, Masks & - & Video/Images & Low - High & Global & \textcolor{red}{IC}, \textcolor{blue}{IS}, \textcolor{green}{OD}, \textcolor{orange}{Other} & \href{https://github.com/DZPeru/fish-datasets}{Github} \\

Fish Image Bank & 24 & Segmentation Masks & 405 images & Images & High & Japan & \textcolor{blue}{IS} & \href{https://haselab.fuis.u-fukui.ac.jp/research/fish/fib.html}{Dataset} \\

\bottomrule
\end{tabular}
}

\end{table*}

\subsection{Existing Datasets in Aquatic Farming}
Several different fisheries datasets are summarized in Table~\ref{tab:datasets-fisheries} and Fig.~\ref{fig:all-datasets}B, whereas a detailed overview of some key datasets is given below:

\subsubsection{Fish4Knowledge} 
The Fish4Knowledge dataset \cite{fisher2016fish4knowledge} consists of 27,230 underwater images captured off the coast of Taiwan, featuring 23 different fish species. A notable aspect of this dataset is its imbalanced nature, i.e., the top 15 species account for 97\% of the images, with the most common species representing about 44\% of the total. The image resolutions range from approximately 30$\times$30 pixels to 250$\times$250 pixels, with most images taken from a lateral perspective. This view is beneficial for studying the dorsoventral elongate body plan, which is a common feature among the selected species, and it gives a distinct shape to the fish when viewed from the side. The images also have generally light backgrounds that improve the visibility and silhouette of the fish, which is an advantage for tasks like species identification and behavioral analysis.

\subsubsection{DeepFish} 
The DeepFish dataset \cite{saleh2020realistic} was collected from 20 coastal habitats in tropical Australia. The data collection team positioned cameras on metal frames and lowered them from a vessel to the seabed. They recorded uninterrupted natural behavior of fish video footage in full HD resolution (1920$\times$1080 pixels). An acoustic depth sounder and GPS further added accurate depth and location information to the dataset with minimal disturbance to the marine environment. By having 39,766 video frames capturing natural fish behaviors, DeepFish offers a reliable alternative to traditional and more intrusive data collection methods, such as diver-led censuses, which often disrupt marine life and can lead to biased data. DeepFish, therefore, supports more accurate ecological research and sustainable fisheries management.

\subsubsection{OZFish} 
OZFish \cite{AIMSUWACurtin2019} was developed collaboratively by the Australian Institute of Marine Science (AIMS), the University of Western Australia (UWA), and Curtin University. It is a public fish data resource for different applications of ML/DL in marine research. This dataset supports the automated annotation of Baited Remote Underwater Video Stations (BRUVS) \cite{langlois2020field} and features a dual annotation approach: one set of bounding boxes outlines all fish in an image, while a ``point-and-labelled'' set marks a specific point on each fish along with its species name. It contains around 80k fish cropped images for 70 different families, 507 fish species, and 200 different genera. The dual structure annotations are also combined with manual verification and the inclusion of MaxN (the highest count of fish detected in a single video frame in one-hour time period) values, enhancing the accuracy of models in recognizing and classifying diverse marine life \cite{whitmarsh2017big}.

\subsubsection{WildFish}

The WildFish dataset \cite{zhuang2018wildfish} is one of the most extensive collections available for wild fish recognition, containing 54,459 images across 1,000 fish categories. It surpasses many other datasets, such as the popular CUB-200-2011 dataset \cite{wah2011caltech}, in terms of the number of categories, images per category, and diversity within genera and species. This variety makes WildFish an excellent resource for training models for fine-grained fish recognition across different aquatic environments.

The dataset also presents challenges due to high intra-class variations and minimal differences between some species. For instance, the appearance of Acanthurus Coeruleus can change significantly from juvenile to adult stages, while species like Cleidopus Gloriamaris and Monocentris Japonica differ only by minor features such as a red stripe. Nonetheless, an experiment has been conducted where a ResNet50 model was trained on WildFish and then tested on the QUT dataset \cite{anantharajah2014local}. It demonstrated robustness and strong generalization, thereby solidifying its role in advancing ML applications in the fisheries domain.

\subsubsection{GO-Fish}
The GO-FISH dataset \cite{oremus2024geolocated} maps out where different fish species are known to spawn, focusing on commercially important fishes. It draws information from well-known sources like FishBase\cite{froese2010fishbase} and the SCRFA\cite{russell2014status}. This resource is particularly valuable for researchers and conservationists interested in fish reproduction patterns. However, the dataset mainly covers well-studied regions and does not fully represent spawning activities in remote or less explored areas.

The data in GO-Fish is derived from various field research notes, resulting in a range of precision from broad regional labels (e.g., North Atlantic) to more specific site information. It also contains some imprecise entries as the collection methods are not standardized and include qualitative descriptions. Additionally, the dataset does not track changes over time, limiting its ability to reflect shifts in spawning locations due to environmental changes. 
For more accurate analyses, it is recommended to use the data at a detailed resolution (0.5° × 0.5° or finer) and to combine it with long-term environmental data.

\subsubsection{VLM4Bio}
VLM4Bio \cite{maruf2024vlm4bio} is a specialized dataset with about 469K question-answer pairs designed to evaluate the pre-trained VLMs in the context of biological trait discovery from images, with a particular focus on applications in fisheries. This dataset comprises an extensive collection of around 30,000 images which covers a diverse range of fish species. Each image in the dataset is annotated with detailed descriptions that captures different phenotypic traits such as size, shape, color patterns, and other distinctive features that are important for species identification and trait analysis. The primary objective of creating the VLM4Bio dataset is to facilitate the development and testing of VLMs that can effectively interpret and analyze complex biological data. The dataset is structured to support a range of tasks, from basic species classification to more complex trait discovery tasks which require understanding subtle visual features within the images. This rich annotation also allows zero-shot and few-shot learning, where models are tasked to make predictions about new or rare species without extensive prior training.

\subsection{Machine Learning Techniques in Fisheries}

\subsubsection{Conventional Approaches}

\noindent \textbf{Support Vector Machines (SVMs).} SVMs are widely used in fisheries research for classification and regression tasks \cite{naveen2024advancements}. SVMs work by finding the decision boundary that best separates different classes in the data, such as various fish species or potential fishing zones while maximizing the margin between them \cite{sonmez2024enhancing}. This approach is particularly effective when analyzing complex environmental data, including sea surface temperature and chlorophyll levels. For example, Ya'acob et al. \cite{ya2024review} demonstrated that SVMs could classify fishing zones with an accuracy of 97.6\%, outperforming methods like naive Bayes, which achieved 94.2\%. Furthermore, the use of kernel functions enables SVMs to handle non-linear relationships, making them robust for modeling the multifaceted nature of marine data \cite{ogunlana2015fish}.

\noindent \textbf{K-Nearest Neighbors (KNN).} The KNN algorithm is well-known for its simplicity and effectiveness in fisheries research, particularly for species identification and environmental monitoring \cite{reddy2024deep}. KNN classifies new data points by finding the most similar examples from the training set, which is useful for recognizing fish species from underwater images. Islam et al. \cite{islam2021design} showed that integrating KNN with Internet of Things (IoT) devices allows for real-time monitoring of water quality parameters such as pH, temperature, and turbidity, which are critical factors in maintaining healthy aquatic habitats. Winiarti et al. \cite{winiarti2020consumable} present a kNN‑based system that uses HSV color features from fish meat and GLCM texture features from scales to classify fish freshness for mackerel and tilapia meat, and their scales. They demonstrated a simple, low‑cost alternative to sensor‑based methods for automated assessment of fish suitability for consumption. This method improves feature selection and boosts classification accuracy, even in complex underwater environments where challenges such as water opacity and motion can complicate data collection. Such enhancements support more precise and informed decision-making in fisheries conservation and management.

\noindent \textbf{Decision Trees. }Decision Trees (DT) are used in fisheries management to analyze and monitor marine ecosystem health. For example, a study in the North Sea has employed DTs to categorize the ecosystem into states such as \textit{improving}, \textit{deteriorating}, or \textit{stable} by scoring various ecological and environmental indicators \cite{lockerbie2018applying}. This structured approach simplifies complex data and enables comparisons across different marine systems. Moreover, DTs are valuable in predictive tasks, for instance, forecasting of the fish survival rates under changing environmental conditions. By splitting data into branches based on key variables, DTs provide a clear and visual pathway that shows how different factors influence the outcomes \cite{uddin2024data}.

\noindent \textbf{K-Means Clustering.} K-means clustering is a technique for analyzing the spatial and temporal patterns of fishing activities. A study by Sunarmo et al. \cite{affandi2020clustering} applied K-Means to Vessel Monitoring System (VMS) data, using the Elbow method \cite{cui2020introduction} to determine the optimal number of clusters. This process revealed clear patterns in different activities of fishing, which then guides different strategies to prevent overfishing and thus improve the sustainability of resources  \cite{scientific2024clustering}. Additionally, K-Means clustering can also help identify subpopulations within a broader fish stock by differentiating groups based on different stages of growth, survival rates, or environmental influences. Such key information is critical for adjusting fishing quotas and protecting vulnerable groups, ultimately supporting more data-driven fisheries management \cite{xiao2024population}.

\noindent \textbf{Case-Based Reasoning (CBR).} CBR supports decision-making in fisheries management by leveraging past experiences to address current challenges. Thomas et al. \cite{thomas2023case} demonstrated how CBR can model the location choice behavior of recreational fishers in Connecticut, offering valuable insights into different factors that influence these decisions under varying environmental conditions. This method allows managers to adapt successful strategies from previous historical events to new contexts. For example, if a particular approach effectively controls an invasive species in one area, CBR can help tailor that strategy for similar challenges elsewhere, even when dealing with different species \cite{salsabila2024expert}.

\subsubsection{CNNs}

\noindent \textbf{VGG16. }VGG16 is a popular CNN architecture with 16 trainable layers, including 13 convolutional layers that use 3x3 filters with a stride of 1, interspersed with $2\times2$ max pooling layers. It ends with three fully connected layers, the first two having 4096 units each, and the last with 1000 units for classification. It has been used for fish identification and has delivered high accuracy and robust performance \cite{liu2024research}. For instance, Wibisono et al. \cite{wibisono2024classification} employed VGG16 with transfer learning to develop a marine fish classification system, reporting a training accuracy of 100\% and a validation accuracy of 99.3\%. This model is effective at capturing subtle differences between fish species, which is also a critical capability for biodiversity monitoring, regulatory enforcement, and population assessments. 
In a similar study, Hindarto et al. \cite{hindarto2023comparative} found that while VGG16 outperforms lighter models like MobileNet in terms of accuracy, it does require more computational resources, which may not be feasible for real-time deployment in marine environments. 

\noindent \textbf{ResNet.} Residual Networks (ResNet) \cite{he2016deep} are widely used for classifying marine fish species \cite{dash2024fish,mathur2021fishresnet} due to their deep architecture, ability to capture intricate image features, and overcoming the vanishing gradient problem of the learning process.
Gao et al. \cite{gao2024research} enhanced ResNet50 with a Dual Multi-Scale Attention Network (DMSANet), enabling the model to focus on fine details and achieving an accuracy of 98.75\%. The deep layers in ResNet are found responsible for helping in learning complex patterns in fish scales, colors, and shapes, thereby leading to highly accurate classifications \cite{nisa5097940improved}. Some other studies, such as Malathi et al. \cite{malathi2023optimzied}, have refined ResNet50’s feature extraction capabilities by introducing novel hybrid capuchin-based coevolving particle swarm optimization (HC2PSO) algorithm to boost both precision and efficiency in marine species recognition.

\noindent \textbf{Mask R-CNN.} Mask R-CNN \cite{he2017mask} excels in detecting and segmenting real-world objects and, therefore, has also been adapted to identify individual fishes in underwater images, making it valuable for both commercial applications and research. Yi et al. \cite{yi2024coordinate} developed a Coordinate-Aware Mask R-CNN that integrates CoordConv and Group Normalization to tackle challenges like class imbalance and varied underwater conditions. In another application, Mask R-CNN was adapted to measure fish morphological features such as body length and width, which are critical for smart mariculture \cite{chang2021fish}. This automated approach reduces the subjectivity and labor associated with manual measurements, thereby enhancing operational efficiency and contributing to more informed conservation strategies \cite{jain2024advancing}.

\noindent \textbf{EfficientNet.} EfficientNet \cite{tan2019efficientnet} is a powerful CNN architecture that has been adapted for fish detection and classification tasks. Its optimized design allows deployment on devices with limited processing power, such as onboard systems in marine vessels or remote monitoring stations, thereby also supporting continuous data collection in diverse marine settings \cite{kannan2024accurate}. For example, Gong et al. \cite{gong2022fine} enhanced EfficientNet with attention mechanisms to improve fine-grained classification. This adaptation enabled the model to distinguish between species with only subtle differences demonstrating its robustness in challenging conditions. Another work by Syifa et al. \cite{syifa2023yolov7} combined EfficientNet with YOLOv7 to both detect and classify fish in an aquaculture environment. This hybrid approach achieved high accuracy and mean average precision (mAP) scores, highlighting the benefits of integrating EfficientNet with other DL models.

\noindent \textbf{You Only Look Once (YOLO).} YOLO \cite{redmon2016you} is a real-time object detection model known for its speed and accuracy. By predicting multiple bounding boxes and their class probabilities in a single evaluation, YOLO is particularly well-suited for dynamic marine environments where rapid tracking of fish movement and detection of environmental changes are essential \cite{narang2024edge}. In fisheries, YOLO processes video streams in real-time, enabling researchers and marine experts to monitor fish behavior as they occur in natural habitats. This capability is crucial for situations such as tracking spawning events or identifying invasive species. Additionally, YOLO's efficiency allows it to run on lower-end hardware that is very common in remote underwater vehicles or automated monitoring systems \cite{feng2024ceh}. Despite its rapid processing, YOLO maintains high accuracy, ensuring reliable species identification for biodiversity assessments and regulatory compliance, which are key components for sustainable fisheries management and ecological preservation.

\subsubsection{Vision Transformers (ViTs)}
ViTs \cite{dosovitskiy2021imageworth16x16words} have emerged as a powerful alternative to traditional CNNs, particularly for processing complex visual data in fisheries research. By treating image patches as sequences, ViTs capture intricate spatial relationships that are vital for analyzing the dynamic and diverse marine environment. Below is an overview of various specialized ViT models applied in the fisheries domain.

\noindent \textbf{Fish-TVIT.} The Fish-TViT model \cite{gong2023fish} addresses fish species classification by utilizing a vision transformer (ViT) combined with transfer learning. The paper uses several preprocessing steps, such as enhancing the raw fish images through cropping and cleaning, followed by applying other data augmentation techniques (such as random flipping) to increase the diversity of the training dataset. Subsequently, the model employs a pre-trained ViT to expedite the training process, which leverages vast and diverse generalized features learned from the ImageNet dataset \cite{deng2009imagenet} to extract critical details from the images effectively. Unlike traditional CNNs that process entire images at once, the transformer structure processes the images by dividing them into patches, embedding these patches spatially, and then feeding them through multiple transformer blocks. This approach enables the model to capture subtle variations in color, texture, and shape features that are important for distinguishing between visually similar fish species \cite{gao2024research}.

For classification, the extracted features are input into a Multi-Layer Perceptron (MLP) that finalizes the species identification. The Fish-TViT model incorporates a label smoothing loss function to overcome overfitting and enhance the generalization capabilities of the classifier. Additionally, it utilizes Gradient-weighted Class Activation Mapping (Grad-CAM) \cite{selvaraju2017grad} to visualize and understand which parts of the images are most informative for making classification decisions. It has been tested on both low-resolution marine and high-resolution freshwater fish datasets. This combination of various techniques enables the Fish-TViT to achieve high classification accuracies and outperform traditional CNN models like ResNet50 \cite{he2016deep}, VGG19 \cite{simonyan2014very}, DenseNet121 \cite{huang2017densely}, and ConvMixer \cite{trockman2022patches}. Ultimately, Fish-TVIT provides a more subtle understanding of marine ecosystems by classifying both low and high-resolution marine and freshwater fish images, respectively.
Such fine-grained analysis helps support maintaining sustainable fish populations and also protect endangered species \cite{elmezain2025advancing}.

\noindent \textbf{TFMFT.} TFMFT \cite{li2024tfmft} stands for transformer-based multiple fish tracking model. It utilizes spatio-temporal data in a transformer-based architecture with four components named CNN block (used for initial feature extraction), encoder block, decoder block, and ID matching module to track multiple subjects at the same time across video frames. By leveraging transformer mechanisms, the model processes extensive video data to identify key behavioral trends and shifts in habitat use as done in \cite{yao2024fmrft}. This model adeptly handles complex scenarios that are often encountered in fish tracking, such as occlusions and dynamic background changes, making it highly suitable for real-world applications. 
The tasks of predicting the positions of fish within a frame and tracking their movements across sequences addressed by TFMFT are invaluable for understanding fishing quotas, fish migration patterns, spawning events, overall population dynamics \cite{fu2024simulation}, and habitat protection measures to ensure sustainable use of marine resources \cite{mandal2024ai}. 
A notable feature of the TFMFT model is the Multiple Association (MA) method, which enhances tracking accuracy by reducing the likelihood of misidentifying fish when they cross paths or momentarily hide behind each other. The model also uses intersection-over-union (IoU) matching with an ID matching module, which helps ensure that each fish is tracked consistently in challenging conditions.

The effectiveness of the TFMFT model was tested on the MFT22 dataset \cite{li2024tfmft}, which includes a collection specifically created to include a variety of fish species and environmental settings. The model demonstrated high accuracy and robustness using a high Identification F-Score (IDF1), thereby outperforming other tracking techniques such as OC-Sort, CMFTNet, TrackFormer, and FairMOT in handling occlusions and maintaining reliable tracking under dynamic conditions.

\noindent \textbf{DyFish-DETR.} DyFish-DETR (Dynamic Fish Detection Transformer) \cite{wang2024dyfish} is a specialized model designed to adapt the original DETR architecture \cite{carion2020end} for the challenging conditions of underwater environments. It consists of two main components: (1) DyFishNet, a feature extraction network designed to capture detailed textural information (including variations in scales and skin textures that are essential for distinguishing between similar species) from fish images, and (2) a slim hybrid encoder, which enhances the model's efficiency by reducing computational load and improving overall speed, ensuring applicability in resource-limited settings.
The overall approach enables processing and interpreting complex scenes from underwater videos where fish may be moving rapidly or obscured by aquatic flora and fauna. 
This capability is essential for studying behavioral patterns, inter-species interactions, and environmental responses, while the real-time processing supports immediate conservation actions and informed management decisions \cite{kong2024conceptual, singamaneni2024optimizing}.

\noindent \textbf{CFFI-VIT.} The CFFI-ViT \cite{liu2024cffi} model is a specialized version of the basic ViT, which is tailored for the precise classification of fish feeding intensity in aquaculture environments. It is specifically focused on the rainbow trout fish and addresses the specific needs of real-time monitoring in aquaculture. The core modification in the CFFI-ViT model involves a significant reduction in the number of transformer encoder blocks, decreasing from 12 to just 4. This reduction dramatically lowers the model's computational needs, making it more suitable for deployment on mobile and edge devices with limited processing capabilities. Additionally, a residual module is added to the head of the transformer. This module improves the model’s ability to extract essential features from images, which helps in classifying between strong, moderate, and weak feeding intensities.

The efficiency of the CFFI-ViT is highlighted by its reduced computational load, which is 65.54\% lower than the standard ViT, yet it achieves a 5.4\% improvement in classification accuracy on the validation set. When it is compared to other CNN models like ResNet34 \cite{he2016deep}, MobileNetv2 \cite{howard2017mobilenets}, VGG16 \cite{simonyan2014very}, and GoogLeNet \cite{szegedy2015going}, the CFFI-Vit not only shows higher accuracy but also demonstrates the potential for real-time application without sacrificing performance. These enhancements make the CFFI-Vit an excellent model for aquaculture practitioners in understanding monitoring growth, migration, responses to environmental stressors, and conservation strategies, enabling sustainable fisheries management \cite{lai2024artificial, veiga2024fine}.

\noindent \textbf{SwinConvMixerUNet. }The SwinConvMixerUNet \cite{pavithra2024efficient} model uniquely combines two different techniques named swin transformer \cite{liu2021swin} and ConvMixer module. The Swin transformer model excels at handling the spatial dynamics of images. Therefore, it adapts very well to varying textures and backgrounds, such as murky waters or reef environments. This capability is vital as underwater images often suffer from poor visibility and distortion caused by water movements. The ConvMixer module, on the other hand, enhances the model's ability to process image details by effectively mixing information across different channels. This helps in capturing diverse features of underwater objects, whether they're small fish or large seabed structures, across multiple scales. 

The model works by first using swin transformers in its encoder to deeply analyze and extract features from the images. Then, the ConvMixer modules in the decoder reconstructs a detailed segmentation map from these features. The architecture is specially designed to be efficient with high-resolution images by employing techniques like patch partitioning to ensure that the model can handle large amounts of data without overwhelming computational resources. Moreover, the model incorporates specific image processing enhancements to overcome common underwater imaging issues like light absorption and scattering, improving the overall clarity and usability of the output. This identifies and classifies various marine elements, which is particularly beneficial for tasks such as habitat mapping and species identification in intricate underwater landscapes \cite{zhou2024benthic}. Additionally, the model’s ability to process dynamic changes like coral bleaching or invasive species encroachment enables continuous monitoring and real-time data analysis.

\subsubsection{Foundation Models for Aquaculture}



\noindent \textbf{MarineGPT.}
MarineGPT \cite{zheng2023marinegpt} is a two-stage, domain-specialist vision–language foundation model built to unlock the secrets of the ocean. In the first stage, MarineGPT undergoes marine-specific continuous pre-training on Marine-5M which is a dataset of over five million image–text pairs curated from documentaries, marine websites (e.g., FishDB \cite{yang2020fishdb}, Reeflex \cite{Mainpage41}), and field surveys where each image is paired with rich captions expanded through 129 hierarchical attributes (size, shape, diet, habitat, etc.) to inject detailed marine knowledge. This stage optimizes not only the linear projection layers but also the Q-Former atop a BLIP-2 \cite{li2023blip} ViT encoder, ensuring that visual embeddings capture the fine-grained distinctions among hundreds of reef fish, corals, and megafauna. In the second stage, the model is fine-tuned on 1.12 million high-quality, instruction-following image–text pairs generated via 50 marine-expert templates and ChatGPT/GPT-4 \cite{achiam2023gpt}, thereby teaching MarineGPT to interpret user intent and to produce scientifically grounded, multi-sentence responses aligned with those instructions.

When compared to generalist multi-modal assistants like MiniGPT-4 \cite{zhu2023minigpt} and GPT-4V, MarineGPT excels at fine-grained recognition and domain-specific knowledge delivery. In terms of qualitative performance, MarineGPT generates long, detailed descriptions (including both common and scientific names), ecological roles, conservation recommendations, and multi-turn dialogues about species such as dog-faced pufferfish, weedy seadragons, or bottlenose dolphins. MarineGPT can distinguish visually similar species (e.g., three different cichlids or rays) and answer follow-on queries about habitat, diet, or threats with accurate, reference-style information. 


\noindent \textbf{AquaticCLIP.}
AquaticCLIP \cite{alawode2025aquaticclip} is a contrastive vision–language foundation model specifically tailored for underwater scene analysis. It is pre-trained on a dataset of two million image–text pairs, curated from diverse aquatic sources such as YouTube documentaries, National Geographic, marine biology textbooks, and social media, with captions both manually refined and generated via MarineGPT \cite{zheng2023marinegpt} at image and instance levels. To adapt CLIP’s \cite{radford2021learning} dual‐encoder framework to the aquatic domain, AquaticCLIP introduces two lightweight, learnable modules: a prompt-guided vision encoder that aggregates patch embeddings through cross‐attention with learned prompts, focusing on the most semantically relevant regions, and a vision-guided text encoder that refines textual representations by attending to these visual prompts. A semantic cleaning module further distills generated captions to retain only the keywords most aligned with each image, and the entire model is trained end-to-end with a bidirectional cross-modal contrastive loss.

When evaluated in zero-shot classification setting across seven underwater benchmarks including marine animal images (MAI) \cite{MarineAn38}, sea animal images (SAI) \cite{SeaAnima84}, fine-grained fish (FishNet \cite{khan2023fishnet}, FishNet Open Images dataset \cite{kay2021fishnet}, Large Scale Fish Dataset \cite{ulucan2020large}), coral species (CSC) \cite{Coralspe98}, and coral classification (CC) \cite{CoralsCl57}, AquaticCLIP achieves up to 96.8\% accuracy and 96.4\% F1 on CSC, and consistently outperforms prior VLMs (e.g. Frozen CLIP \cite{radford2021learning}, CoOp \cite{zhou2022learning}, MarineGPT \cite{MarineAn38}) by 5–8 F1 points. In few-shot and linear‐probe evaluations, it further boosts mean accuracy from 82\% to 92\% on fine-grained tasks. These gains across classification, detection, segmentation, and counting tasks demonstrate that AquaticCLIP’s combination of large-scale, domain-rich pretraining and its prompt and vision-guided encoders yields robust, generalizable representations for a wide spectrum of aquatic computer-vision applications.



\noindent \textbf{AgriCLIP.} The AgriCLIP \cite{nawaz2024agriclip} framework introduces an approach for improving image-text model performance for zero-shot classification tasks in agricultural tasks, including fisheries. The paper describes the construction of a specialized image-text dataset called ALive, tailored for the crops, livestock, and fisheries sectors. This dataset encompasses approximately 600,000 image-text pairs, curated to include a wide variety of fish species across different environments, from indoor tanks to vast underwater settings.

AgriCLIP leverages this comprehensive dataset to fine-tune a vision and language model, CLIP \cite{radford2021learning}, by combining contrastive and self-supervised learning methods. This approach allows the model to understand and categorize complex visual and textual data related to fish by training a self-supervised DINO \cite{caron2021emerging} model and then aligning its vision encoder with the CLIP vision and text encoders. This alignment enhances the ability of the model to perform zero-shot classification tasks effectively. The model's architecture, which integrates these learning strategies, has shown significant improvements in identifying and classifying fish species, achieving notable gains of 24.5\% in accuracy for the fisheries domain over the generic CLIP model.


\noindent \textbf{Grounded-SAM.} The grounded-segment-anything (Grounded-SAM) \cite{ren2024grounded} is a foundation model adapted for identification, segmentation, and size measurement of individual fishes in \cite{hasegawa2024robust}. 
The framework employs ResNet50 \cite{he2016deep} to identify the type and quantity of the fish. In addition, it utilizes Mask Keypoint R-CNN \cite{he2017mask} to detect key anatomical points in fish, which aids in the calculation of fish length and other vital metrics.
To support and enhance the accuracy of this framework, the authors utilize the Fish Image Bank (FIB) dataset \cite{hasegawa2024robust}, which is a curated collection of different fish images, masks, and kypoints.

\noindent \textbf{MarineInst.} MarineInst \cite{zheng2024marineinst} is a specialized foundation model that aims to enhance marine image analysis. It is specifically designed to handle the unique challenges associated with underwater imaging, such as poor visibility, color distortion, and dynamic lighting, which are common in marine environments. MarineInst operates by segmenting individual objects within an underwater image and then generating descriptive captions for each identified instance. This dual functionality not only allows for precise object detection vital in cluttered underwater scenes but also adds a layer of semantic understanding by describing what each object is. For instance, in a fisheries context, MarineInst can differentiate and describe various species of fish, noting features such as size, color, and behavior within their natural habitat. Here, the instance segmentation module is built on top of SAM \cite{kirillov2023segment} model along with binary instance filtering in the mask decoder. It also uses the frozen VLMs along with the Vicuna model \cite{vicuna2023} for instance captioning. 

The model was trained on the MarineInst20M dataset, which consists of 20 million instance masks and is the largest of its kind. It also contains 2.65M mask samples, where 1.89M are from human-annotated instances and 0.76M non-instance masks constructed using SAM. The remaining 17.3M instances are the model-generated masks after the binary instance filtering. 

\section{AI for Livestock}
\label{sec:livestock}
Livestock farming is another cornerstone of global agriculture, providing meat, dairy, eggs, and other essential animal products. As consumer demand intensifies and sustainability becomes a paramount concern, modern livestock systems have begun incorporating data-driven techniques to optimize productivity, enhance animal welfare, and minimize ecological impact \cite{rai2023advancing}. AI has been utilized for many different innovations, offering new ways to monitor, analyze, and improve day-to-day farming practices \cite{subeesh2021automation}. However, implementing AI in livestock management involves addressing various complexities, including environmental differences among farms, varying animal behaviors across species, and constraints related to data collection or regulatory frameworks.


\subsection{Machine Learning Tasks in Livestock Farming}

\begin{figure*}[hbtp]
    \centering
    \includegraphics[width=\textwidth]{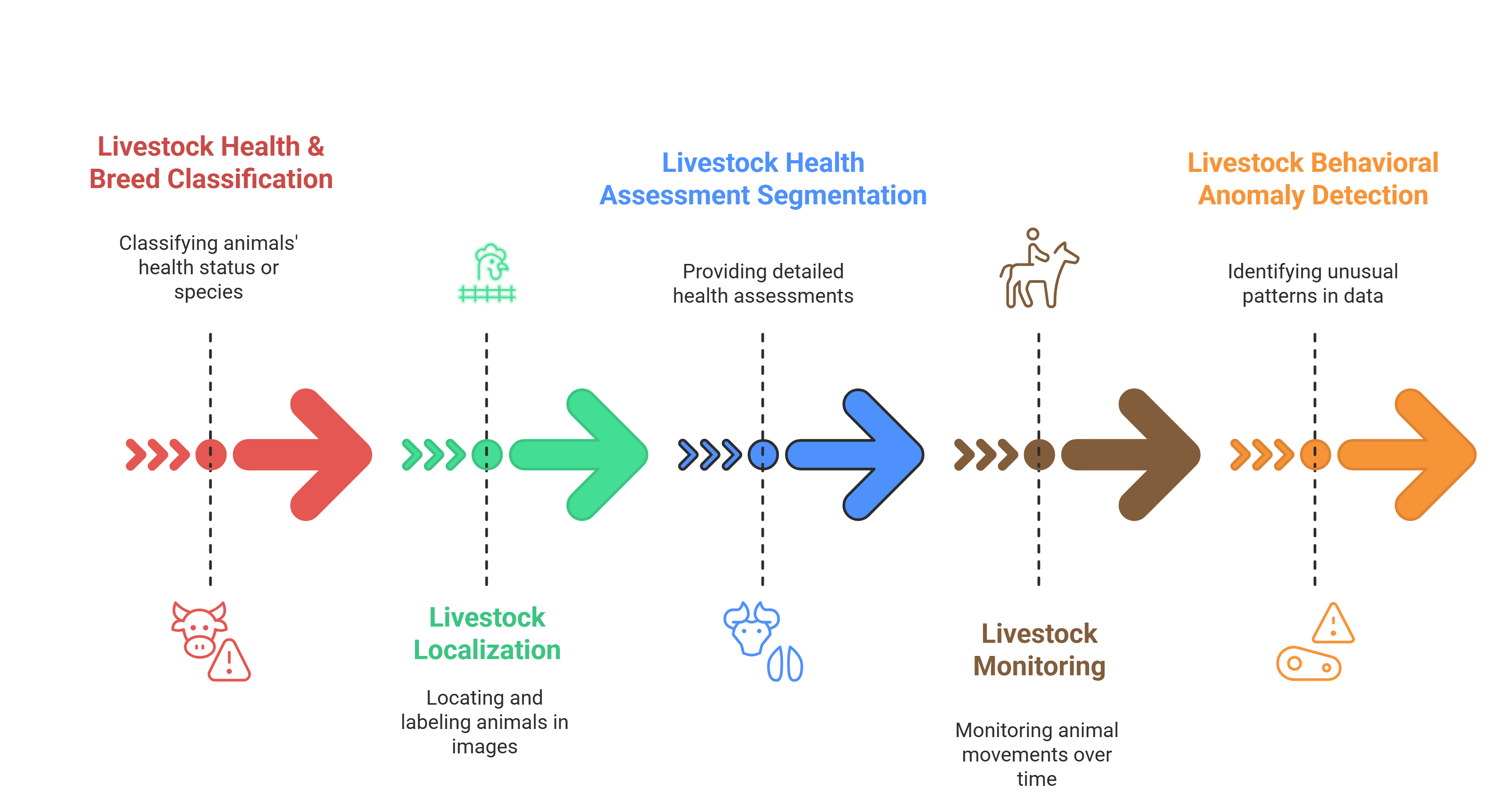}
    \caption{Different tasks used in managing livestock health and behavior, including image classification to identify species or health status, object detection for locating animals, image segmentation for detailed health assessments, object tracking to monitor movements, and anomaly detection for identifying unusual patterns.}
    \label{fig:livestock-tasks}
\end{figure*}

This sub-section provides a brief introduction to different machine learning tasks in livestock farming, as also shown in Fig.~\ref{fig:livestock-tasks}.

\noindent \textbf{Livestock Health \& Breed Classification. \textcolor{red}{(IC)}} In livestock farming, one of the most common uses of classification is health monitoring and disease classification. For example, a camera can capture images of cows in a barn, and a DL model classifies each image as healthy or potentially sick based on the visual features of the cow's appearance. These features might include droopy ears, irregular posture, or even subtle differences in skin texture that indicate stress. By continuously classifying new images, livestock farmers can spot health issues early and intervene before they spread throughout the herd \cite{neethirajan2021digital}. Moreover, classification task can also help monitor general well-being or productivity. For instance, an AI framework might identify whether a chicken is laying or resting, helping to track egg-production cycles. However, the limitation of such a task is that it doesn’t locate specific animals or body parts within each image. This means that while agronomists know something might be amiss, they may not immediately know which specific cow or chicken needs attention, which is a challenge addressed by more advanced tasks like object detection and segmentation \cite{li2021practices}.


\noindent \textbf{Livestock Localization. \textcolor{green}{(OD)}} It is a step forward by locating each animal subject in an image, typically by showing bounding boxes around them. This is particularly important on large farms where hundreds or thousands of animals share the same space. Here, if the system detects each animal and labels them accordingly, e.g., Cow 101 or Chicken 45, farmers can count how many animals are in a pen, and identify any individual that is going through distress. This process is key for day-to-day operations such as making sure each animal is being fed or ensuring none are missing from a herd. Moreover, object detection supports more detailed insights into herd or flock behavior. For instance, if a camera feed in a pasture consistently shows groups of sheep huddled in certain areas, the managers can use that information to decide where to place shade or water sources. Similarly, if we combine it with automated alerts, object detection can also signal that too many animals are crowding around a feeding station, which can indicate potential feed shortages or unbalanced resource distribution \cite{ganga2024object}. In short, while detection algorithms require more computational power than classification, they offer a fine-grained understanding of what’s happening in real time.


\noindent \textbf{Livestock Health Assessment Segmentation. \textcolor{blue}{(IS)}} While object detection locates subjects with bounding boxes, image segmentation produces more precise boundaries at the pixel level. In livestock settings, this can be extremely valuable for detailed health assessments. For example, segmenting a cow’s legs or udder might reveal small swellings or injuries that would be missed by a simple bounding box. This depth of detail allows veterinarians or farm staff to measure body parts over time, noting changes in size, shape, or color that might indicate an emerging health issue. Segmentation also helps in complex environments where animals overlap or share tight living spaces \cite{xu2020automated}. By separating each animal from the background or neighboring animals, the system can measure behavioral cues, such as how closely animals group together or how much personal space each one maintains. This kind of information informs welfare assessments, allowing farmers to adjust pen layouts, feeding areas, or herd sizes. Although segmentation methods require more robust datasets and can be computationally heavier, they provide the higher level of detail needed for precision livestock management, where every subtle change matters.

\noindent \textbf{Livestock Monitoring. \textcolor{brown}{(OT)}} It takes detection or segmentation over time, linking the same animals across consecutive frames in a video stream \cite{pereira2022sleap}. In practical terms, this is how farms determine each animal’s movement patterns, feeding frequency, and social interactions over hours, days, or weeks \cite{compte2024animal,huang2022deep}. Tracking can illuminate, for example, that a particular cow has sharply reduced its daily walking distance which can be an early sign of lameness \cite{zheng2023cows} or that certain chickens stay near feeding areas for too long \cite{yang2024deep}. In addition, tracking is essential for understanding group behavior in dynamic environments as well. If a system consistently sees that animals from different pens mix too often, farmers might need to re-examine fencing or gating strategies to prevent cross-contamination of diseases or fights between incompatible groups. Beyond behavior monitoring, tracking can be combined with wearable sensors like RFID tags to verify the identity of each animal, offering a double check on video-based methods. This approach helps maintain accurate records of each animal’s history, from health checks to feeding schedules and ultimately supporting both better welfare and farm efficiency.


\noindent \textbf{Livestock Behavioral Anomaly Detection. \textcolor{orange}{(Others)}} Livestock environments can be unpredictable, with machinery failures, climate shifts, or sudden disease outbreaks all posing risks. Anomaly detection systems learn what normal patterns look like whether in terms of sensor data (e.g., temperature, humidity, feed levels) or video feeds (movement patterns and group dynamics) and then flag significant deviations \cite{kakar2024timetector,ismail2019efficient}. For instance, a sudden drop in feed consumption could indicate a hidden abnormality in the feeding mechanism, or it might reveal an emerging disease if multiple animals stop eating at once. The advanced anomaly detection approaches can also incorporate external information, such as weather forecasts or disease reports in nearby regions, to anticipate possible threats. In practical terms, this might mean sounding an alert if the barn’s temperature is rising faster than expected on a hot day, prompting immediate ventilation checks \cite{park2021anomaly}. By catching these early events, farmers reduce downtime, avoid large-scale disease spread, and maintain a stable environment for the animals. Anomaly detection thus acts as a safeguard, helping farm managers stay proactive rather than reacting only when problems become severe.


\subsection{Challenges in Livestock}
\begin{figure}[hbtp]
    \centering
    \includegraphics[width=\columnwidth]{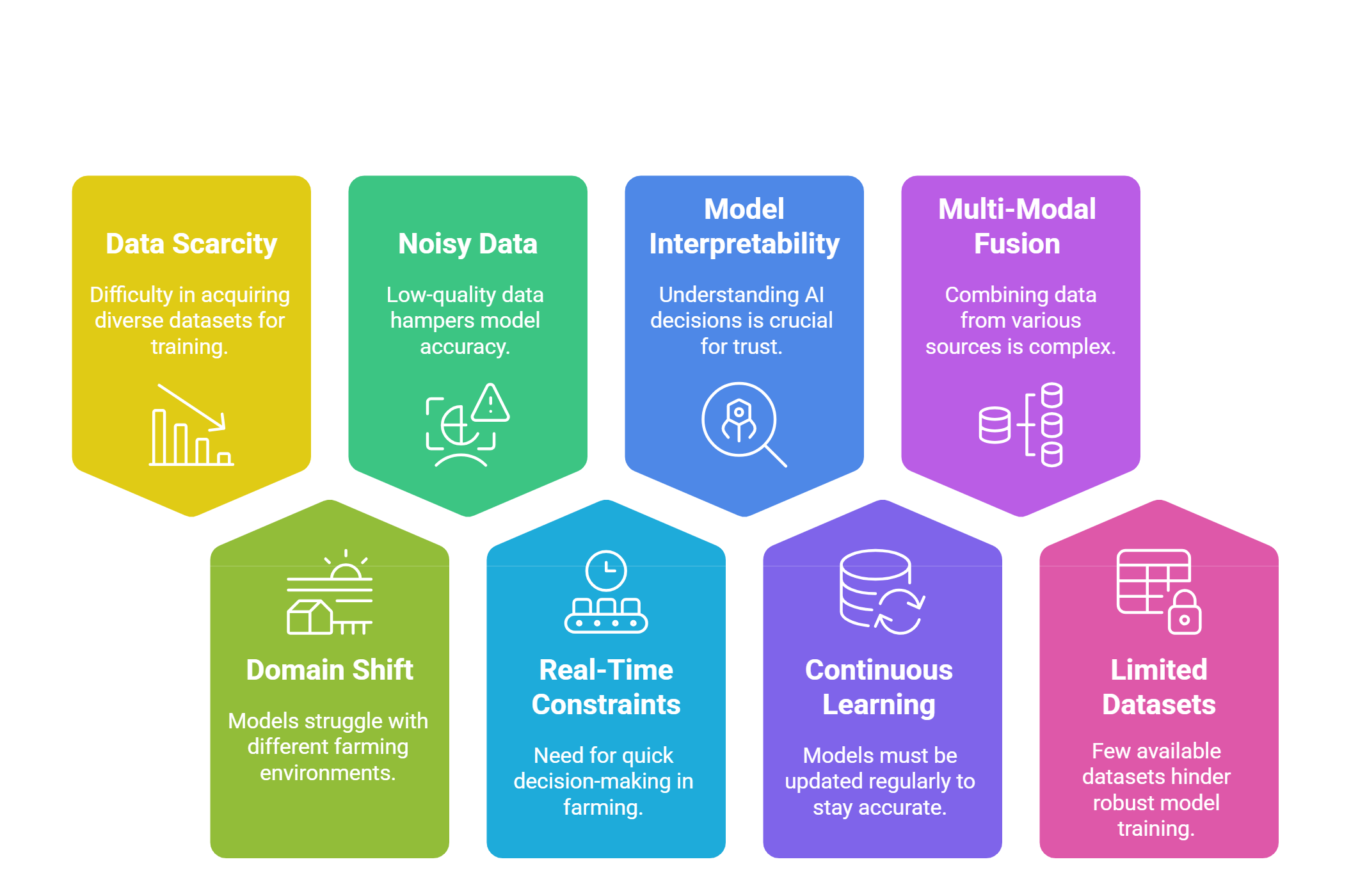}
    \caption{Challenges in the livestock domain include data scarcity, noisy data, model interpretability, multi-modal fusion, domain shift, real-time constraints, continuous learning, and limited datasets, which impact the effectiveness of AI in livestock farming}
    \label{fig:challenges-livestock}
\end{figure}

There are a variety of challenges associated with livestock farming using deep learning from both technical and non-technical perspectives (Fig.~\ref{fig:challenges-livestock}). Some of them are discussed below briefly.

\noindent \textbf{Data Scarcity and Class Imbalance. }One core challenge in deep learning is the need for large, diverse training datasets. In livestock, acquiring such datasets can be difficult, especially for rarer conditions like specific cattle diseases or unusual behaviors of livestock animals. As a result, models may end up with class imbalance, where healthy examples far outnumber the disease examples leading to biased performance that overlooks critical cases. Therefore, collecting more balanced data is required which often takes a lot of time, specialized veterinary guidance, and careful curation to ensure that every significant category (e.g., illnesses, behaviors, breeds) is properly represented.

\noindent \textbf{Domain Shift and Transferability. }Deep learning models trained in one environment (indoor farming) may struggle for a different context data like outdoor farming under variable weather. This domain shift happens due to differences in lighting conditions, background features, or even the appearance of animals in different breeds. To address this issue, different transfer learning techniques \cite{jiang2022transferability} are introduced which partially address this by retraining a portion of a model on new data, but compiling enough representative data from multiple environments still remains a challenge. Furthermore, the research must also consider potential concept drift, where animal behavior or farm conditions change over time, making previously learned representations less reliable.

\noindent \textbf{Noisy and Low-Quality Data.} Noisy data is available for almost every domain in its different forms. For livestock farming, it involves challenging imaging conditions like dust, dim lighting, motion blur from moving animals, age growth, and occlusions caused by crowding or farm equipments. These factors lead to noisy or low-quality data, which can hamper the model’s accuracy in tasks such as image classification or object detection. To overcome this noisy impact, preprocessing techniques like image enhancement, denoising, or data augmentation need to be introduced. However, using these techniques can also add complexity and may only partially compensate for truly poor image conditions which is often the case. Moreover, sensors like accelerometers or RFID tags can also produce inaccurate readings if devices are not well-maintained or if animals tamper with them \cite{costa2021review}.

\noindent \textbf{Real-Time Constraints. }Many farms require real-time or near-real-time decision processes which helps automate the farming process. For instance, a veterinarian might need to know the real-time health status of cattles for identifying if the catlles are able to eat properly and does not suffer from any health issue. For this purpose, DL architectures like CNNs or ViTs can be computationally heavy which make quick inference difficult without powerful GPUs or specialized hardware. On the other hand, using remote servers for remote farms often struggles with limited connectivity or older infrastructure, where implementing fast, on-premises AI solution can be costly and technically challenging. For this purpose, techniques like model compression, quantization, and edge computing solutions are often explored, but each approach adds layers of complexity to an existing deployment process.

\noindent \textbf{Model Interpretability and Explainability.} Model interpretability means how easily we can understand why an ML model makes certain decisions, whereas explainability is about using methods to clearly show what influences these decisions. Now, as DL models grow more complex, interpretability becomes a significant concern. Farmers and veterinarians may be skeptical of these black box predictions without a clear explanation for why an animal is flagged as sick or stressed. In regulated industries especially those involving food production, an explainable AI approach helps build trust in the customer perception and that is how it meets the higher compliance needs. To overcome this issue, techniques like Grad-CAM and attention visualization are explored, which highlight the regions in an image that influenced the predictions and provide more clear insights. However, achieving such an explainable approach requires additional development and often specialized technical knowledge, which is often lacking in this domain.

\noindent \textbf{Continuous Learning and Model Updates.} In livestock farming, things change often which includes the addition of new animals, dietary changes, and season shifts which occur time-to-time. For example, a model trained for cow baby classification will not be able to work better for the elder cows as the age growth introduces significant amount of visual differences in physical appearance. Therefore, if the models that monitor these changes are not updated on a regular basis, they can start making errors or skip important anomalies that could lead to negative outcomes. To prevent this, these models need to be updated continuously with newly updated details. This requires good organization of data, frequent updates, and methods to ensure that new information doesn't erase what the model has already learned.

\noindent \textbf{Multi-Modal Data Fusion.} Effective monitoring of livestock animals often involves using data from different sources like video cameras, thermal cameras, movement trackers, and climate sensors, along with past health records. DL systems that handle such different modalities need to combine these varied types of data, such as merging video with sensor readings in real time. That is why designing such networks that fuse different data types is more complicated than single-modality models which often involve specialized modules and ensembling methods \cite{xiao2022survey}. However, if the fusion process is properly integrated, this multi-modal approach provides deeper insights, like linking animal behavior changes to temperature changes. However, it also comes with its own challenges like keeping the data in synchronization, aligning different data types correctly (text vs images), and managing the high computing needs.

\noindent \textbf{Limited Access to Benchmark Datasets.} Compared to fields like general object recognition (e.g., ImageNet \cite{deng2009imagenet}), publicly available livestock datasets are far fewer and small in size. Most existing datasets lack sufficient variety in breed, geography, or environment to train robust models. This shortage of standardized benchmarks makes it harder to compare different solutions. To overcome this issue, there is a need for more collaborative data-sharing efforts among research institutions, commercial farms, and government entities. However, data privacy and competition concerns can work as a barrier, so establishing trusted platforms and clear legal frameworks is required.

\begin{table*}[hbtp]
\centering
\caption{Summary of publicly available datasets used in livestock domain. Each dataset is categorized based on different characteristics such as the number of classes, annotation type, dataset size, geographical focus, and supported tasks—such as Image Segmentation (IS), Object Detection (OD), and Image Classification (IC). }
\label{tab:datasets-livestock}
\renewcommand{\arraystretch}{1.5} 
\resizebox{\textwidth}{!}{%
\begin{tabular}{@{}>{\raggedright\arraybackslash}p{2.5cm}>{\raggedright\arraybackslash}p{1.4cm}>{\raggedright\arraybackslash}p{3.5cm}>{\raggedright\arraybackslash}p{2.5cm}>{\raggedright\arraybackslash}p{2cm}>{\raggedright\arraybackslash}p{2cm}>{\raggedright\arraybackslash}p{1.8cm}>{\raggedright\arraybackslash}p{2.3cm}>{\raggedright\arraybackslash}p{1.5cm}@{}}
\toprule
\textbf{Dataset} & \textbf{Classes} & \textbf{Annotation} & \textbf{Dataset Size} & \textbf{Modality} & \textbf{Resolution} & \textbf{Geographical Focus} & \textbf{Tasks} & \textbf{Link} \\ \midrule

GalliformeSpectra & 10 & Class Labels & 5050 images & Images & Medium & Germany, UK, USA, Egypt, Italy, Romania, Bangladesh & \textcolor{red}{IC} & \href{https://data.mendeley.com/datasets/nk3zbvd5h8/1}{Mendeley} \\

FriesianCattle2017 & 1 & Bounding Box & 940 images & Images & Medium & - & \textcolor{green}{OD} & \href{https://data.bris.ac.uk/data/dataset/2yizcfbkuv4352pzc32n54371r}{Dataset} \\

AerialCattle2017 & 1 & Bounding Box & 34 Videos, 46,430 Images & Video/Images & High & - & \textcolor{green}{OD} & \href{https://data.bris.ac.uk/data/dataset/2yizcfbkuv4352pzc32n54371r}{Dataset} \\

Aerial Livestock & 3 & Bounding Box & 89 images & Images & High & - & \textcolor{green}{OD} & \href{https://doi.org/10.1007/s41095-019-0132-5}{Paper} \\

HolsteinCattle & 383 & Class Labels, Masks & 3694 images & Images & High & Netherlands & \textcolor{blue}{IS}, \textcolor{green}{OD} & \href{https://doi.org/10.34894/7M108F}{Dataset} \\

CowBehavior & 1 & Bounding Box &$>$1.7M images & Videos/Images & Medium & - & \textcolor{green}{OD} & \href{https://doi.org/10.5281/zenodo.3981400}{Zenodo} \\

OpenCows2020 & 1 & Bounding Box & 3703 Images & images & High & - & \textcolor{green}{OD} & \href{https://doi.org/10.5523/bris.10m32xl88x2b61zlkkgz3fml17}{Dataset} \\

CowDatabase2 & - & Depth Maps, Point Clouds & 119 images & Images & High & Russia  & \textcolor{orange}{Other} & \href{https://github.com/ruchaya/CowDatabase2}{Github} \\

NWAFU CattleDataset & 16 & Pose Landmarks & 2432 images & Images & Medium & -  & \textcolor{orange}{Other} & \href{https://github.com/fqcwd/CMBN/tree/main}{Github} \\

CID & 8 & Class Labels, Satistical Data & 17,899 images & Images & High & Bangladesh  & \textcolor{red}{IC}, \textcolor{orange}{Other} & \href{https://github.com/bhuiyanmobasshir94/CID}{Github} \\

CattleEyeView & 1 & Bounding Box & 30,703 images & Video/Images & High & -  & \textcolor{green}{OD}, \textcolor{brown}{OT}, \textcolor{orange}{Other} & \href{https://github.com/bhuiyanmobasshir94/CID}{Github} \\

Chicken Gender & 2 & Class Labels, Bounding Box & 1800 images & Images & High & China  & \textcolor{red}{IC}, \textcolor{green}{OD} & \href{http://dx.doi.org/10.3390/e22070719}{Paper} \\

Poultry Disease & 4 & Class Labels & 1255 images & Images & High & Tanzania  & \textcolor{red}{IC} & \href{https://doi.org/10.5281/zenodo.5801834}{Zenodo} \\

Broiler Dataset  & 6 & Class Labels & 10,000 images & Images & High & Egypt  & \textcolor{red}{IC} & \href{https://www.mdpi.com/2077-0472/13/8/1527}{Paper} \\

Camel Detection  & 1 & Bounding Box & 417 images & Images & Medium & -  & \textcolor{green}{OD} & \href{https://universe.roboflow.com/khalid-moftah-32hjl/camel-detiction}{RoboFlow} \\

Camel Recognition  & 1 & Bounding Box & 546 images & Images & Medium & -  & \textcolor{green}{OD} & \href{https://universe.roboflow.com/rovrest0gmailcom/camel-detection}{RoboFlow} \\

Horses  & 17 & Class Labels & 10,500 images & Video/Images & Medium & -  & \textcolor{green}{OD} & \href{https://ieeexplore.ieee.org/document/10624972}{Paper} \\

Cow Milk Acquisition  & 10 & Class Labels &$>$1.7M images & Video/Images & Medium & Finland  & \textcolor{red}{IC} & \href{https://doi.org/10.23986/afsci.111665}{Paper} \\

MMCows  & 16 ID \& 7 Behaviors & Multi-Modal & 4.8M images & Video/Images & Medium & Finland  & \textcolor{red}{IC} & \href{https://github.com/neis-lab/mmcows}{Github} \\

Sheep Activity  & 6 & Class Labels & 149,327 images & Video/Images & High & Pakistan  & \textcolor{red}{IC} & \href{https://data.mendeley.com/datasets/w65pvb84dg/1}{Mendeley} \\

Buffalo Pak  & 3 & Class Labels & 325 images & Images & Medium & Pakistan  & \textcolor{red}{IC} & \href{https://data.mendeley.com/datasets/vdgnxsm692/2}{Mendeley} \\

LEsheepWeight  & 3 & Statistical & 6,373 images & Images & Medium & Pakistan  & \textcolor{orange}{Other} & \href{https://doi.org/10.1016/j.compag.2023.107667}{Paper} \\

\bottomrule
\end{tabular}
}

\end{table*}

\subsection{Datasets in Livestock}
In this sub-section, we cover different datasets of livestock ranging from cows to camels as shown in Fig.~\ref{fig:all-datasets}C for different applications.

\subsubsection{GalliformeSpectra}
The GalliformeSpectra \cite{himel2024galliformespectra} dataset is a compiled collection of ten distinct hen breeds which is collected to enhance research in poultry science, genetics, and agricultural studies. It consists of 1010 original images which capture the unique physical attributes and feather patterns of each breed, such as Bielefeld, Blackorpington, Brahma, and others. The images are sourced from various global locations to offer a diverse and rich resource of genetic traits across different hen breeds.

\subsubsection{CID}
The cow images dataset (CID) \cite{shagor2022cid} for regression and classification focuses on enhancing livestock management and procurement within the cattle trading industry. This dataset consists of 17,899 images of cows, annotated with detailed vitals such as gender, color, breed, feed, age, teeth, height, weight, price, and size. Each entry in the dataset includes up to 4 high-resolution images and 30 video processed images. The CID data is aimed to support both classification and regression tasks using vitals data. The dataset addresses challenges such as the cumbersome traditional methods of estimating cattle vitals, which are often costly and inefficient. Also, it contains 8 different cattle breeds with imbalanced data distribution.

\subsubsection{CattleEyeView}
The CattleEyeView \cite{ong2023cattleeyeview} dataset is tailored specifically to enhance precision livestock management through advanced computer vision techniques. It features top-down video footage of cattle, which effectively minimizes occlusions and ensures uniformity in the imagery captured. This perspective is beneficial for tasks such as counting and tracking the movement of cattle within a farm setting. The dataset comprises 14 video sequences that total over 30,703 frames, with comprehensive annotations including bounding boxes for both the body and head, tracking IDs for individual cattle, keypoints for pose estimation, and segmentation masks. These detailed annotations make CattleEyeView a valuable resource which can used to improve livestock management practices such as health assessment, behavior monitoring, and growth tracking, thereby contributing to more efficient and sustainable farming operations.

\subsubsection{MMCows}
MmCows \cite{vu2024mmcows} dataset offers a comprehensive and multimodal approach to precision livestock farming (PLF), particularly focusing on dairy cattle. This dataset was thoroughly gathered over two weeks time and features a variety of synchronized, high-quality data concerning behavioral, physiological, and environmental factors of dairy cattle. It includes data from wearable and implantable sensors fitted on ten milking Holstein cows, capturing metrics like ultra-wideband (UWB) movements, inertial activities, and body temperatures. Additionally, the dataset comprises 4.8 million frames from high-resolution cameras providing multiple isometric views, along with environmental conditions captured through temperature and humidity sensors. A unique aspect of this dataset is its extensive annotation, which covers one full day’s worth of image data, totaling 20,000 frames with detailed annotations including 213,000 bounding boxes. These annotations offer precise information about the cows' locations and behaviors, serving as ground truth for validating the efficacy of various PLF technologies. The MMCOWS dataset supports sustainable dairy farming practices by facilitating detailed analysis of cattle behavior and environmental interactions.

\subsubsection{Cow Milk Acquisition}
Olli Koskela et al. \cite{koskela2022deep} introduced an extensive collection of video data intended for the deep learning-based monitoring of cow behaviors near an Automatic Milking Station (AMS). The data was collected over a continuous two-month period, capturing nearly 19 hours of video, which translates to more than 1.7 million still images. These images were classified into ten different categories corresponding to the most significant behaviors observed in the vicinity of the milking station, such as feeding, resting, and various social interactions among the cows. The aim of this dataset is to provide a substantial basis for training CNNs to recognize and analyze specific cow behaviors, which are critical indicators of animal welfare. This dataset enhances the understanding of cow behaviors in a controlled environment, which is oftentimes the case.

\subsubsection{Cows2021}
The Cows2021 \cite{gao2021towards} dataset introduces a substantial collection of identity-annotated Holstein-Friesian cattle images and videos aimed at advancements in automated animal identification through self-supervised ML techniques. This dataset was curated at the University of Bristol and it comprises 10,402 RGB images and 301 videos, capturing the distinctive black and white coat patterns of the cattle from a top-down perspective within a farm environment. The dataset is particularly designed for developing identification models that can recognize individual animals based on their unique physical markings, which are visible in the high-resolution images provided from the top-view approach. Each image in the Cows2021 dataset is annotated with oriented bounding boxes that highlight the torso of the cattle, omitting parts like the head, neck, legs, and tail to focus on the unique body markings for identification purposes. The videos extend the dataset's utility by offering dynamic views of the animals, enabling the study of movement and behavior over time.

\subsubsection{FriesianCattle2015}
The FriesianCattle2015 \cite{andrew2016automatic} dataset consists of dorsal RGB-D images captured from real-world farm settings. The data capture was performed using the depth-sensing camera to segment cattle from their environment accurately. The system utilizes ASIFT (Affine-SIFT) descriptors to handle the affine transformations typical in dynamic, real-world scenarios where cattle may move or be partially obscured.  The identification process involves putting an RGB-D camera 4m above the ground and depth-segmenting the cattle, then selectively extracting coat pattern features that are predicted to be distinct across individual cows. The resolution of 512 $\times$ 512 pixels was used for images, whereas 1920 $\times$ 1080 was used for the videos at 30 fps. The data contains two splits named training and testing where each split contains 83 and 294 images for 10 and 40 cows, respectively. The extracted features from this data are processed using an SVM with radial basis function (RBF) kernels, which classifies the features based on their uniqueness to each species. 

\subsubsection{FriesianCattle2017}
The FriesianCattle2017 \cite{andrew2017visual} dataset comprises 940 RGB images captured from a fixed camera setup above a barn walkway. The images are used to identify individual cattle based on unique dorsal coat patterns. The dataset features images of 89 different Holstein Friesian cattle, captured post-milking within a controlled environment. The images vary significantly in detail, providing a comprehensive basis for training DL models.

\subsubsection{AerialCattle2017}
AerialCattle2017 \cite{andrew2017visual} dataset is collected using a DJI Inspire MkI UAV which includes videos of cattle from an overhead perspective, facilitating the study of cattle movement and behavior in a less constrained outdoor setting. It consists of 34 video clips capturing small herds grazing, with each clip lasting about 20 seconds and recorded at a high resolution to ensure detailed visibility of the animals' coat patterns. 

\subsection{Machine Learning Techniques in Livestock}

\subsubsection{Conventional Methods}
Conventional methods have formed the backbone of AI applications in livestock, particularly when data was scarce and computational resources were limited. These algorithms typically rely on handcrafted features, for instance, color histograms or shape descriptors in images, or simple statistical metrics in time-series sensor data. While DL now dominates many advanced techniques, these classic methods still serve crucial roles in proof-of-concept studies, pilot implementations, and scenarios requiring interpretability or low computational overhead.

\noindent \textbf{Support Vector Machines (SVMs).} SVMs have been popular for binary classification tasks in livestock such as differentiating healthy animals from those showing early disease symptoms. In the study by Sowmya et al. \cite{sowmya2022classification}, SVM is used in conjunction with the MobileNet architecture to classify various animal images, including hen, cats, sheeps, cow, and horses. This approach has proven highly effective, achieving an impressive classification performance of 99\%. The use of SVM here enhances the ability to classify images accurately by handling features extracted from the MobileNet architecture, which is known for its efficiency in mobile applications. This method is useful for applications like monitoring wildlife near forest areas, potentially reducing animal-vehicle collisions. 

Wan and Bao \cite{wan2010animal} discuss an SVM-based expert system designed to diagnose animal diseases quickly and accurately. This is an important capability for livestock management in remote areas with poor diagnostic facilities. The system leverages SVM's pattern recognition capabilities to classify diseases based on symptoms. This application of SVM helps overcome the limitations of manual disease classification, which can be labor-intensive and error-prone. Furthermore, Xiao et al. \cite{xiao2022cow} explore a refined approach to identify individual cows in a free-stall barn setting using a combination of the Mask R-CNN and SVM model. They effectively segmented images of cows captured from a top-view camera setup, which minimizes occlusions and variations caused by different angles or distances. This segmentation helped in extracting distinct shape features from the cows' backs. These extracted features were then classified using an SVM, which is fine-tuned to distinguish between individual cows with high precision. The system demonstrated an impressive identification accuracy rate of 98.67\%. Moreover, Thermal images of dairy sheeps were used to detect signs of mastitis, with SVMs achieving high accuracy after features like udder temperature patterns were carefully extracted. 

\noindent \textbf{KNN.} Mahmoud et al. \cite{mahmoud2015cattle} introduce a system for cattle classification that leverages the Fuzzy K-Nearest Neighbor classifier (FKNN) combined with the Expectation Maximization (EM) image segmentation algorithm. The system enhances cattle muzzle pattern recognition, which is essential for effective cattle management and tracking. The EM algorithm optimizes the extraction of texture and color features from muzzle images, which are then classified by the FKNN. This method outperforms traditional K-Nearest Neighbor (KNN) classifiers, achieving a classification accuracy of 100\% compared to KNN’s 88\%, thus offering a more reliable and precise solution for managing cattle health records, tracking vaccinations, and ensuring traceability. Another study by Tian et al. \cite{tian2021real} presents a real-time behavioral recognition system for dairy cows that utilizes data from geomagnetic and acceleration sensors attached via collars. This multi-sensor approach enables the detailed tracking and classification of cow behaviors such as feeding, ruminating, running, and resting. By integrating KNN with Random Forest (RF) algorithms into a fusion model, the system achieves an impressive average recognition accuracy of 98.51\%. This model significantly outperforms other methodologies like Gradient Boosting Decision Tree (GBDT) and SVM, providing continuous and accurate monitoring that benefits dairy management by promoting animal welfare and optimizing production. Furthermore, Adanigbo et al. \cite{adanigbofuzzy} outlines a system utilizing the FKNN classifier enhanced by Principal Component Analysis (PCA) for detecting lameness in cattle which is a prevalent issue impacting dairy farm economics. By reducing data dimensionality through PCA and employing FKNN for classification, the system overcomes the limitations of manual and semi-automatic lameness detection methods, delivering robust performance metrics such as sensitivity, specificity, and area under the receiver operating characteristic curve (ROC). This automated lameness detection system facilitates early diagnosis and treatment.

\noindent \textbf{Decision Trees.} Decision tree algorithms are proving highly effective in livestock management, as demonstrated in several studies focused on enhancing animal welfare and disease prevention. A study by Khanh et al. \cite{khanh2016classification} demonstrates the innovative use of a three-degree-of-freedom (3-DOF) accelerometer to monitor and classify the behavior of cows. It shows that by attaching accelerometers to the cows, we can collect data on their movements and activities which can then be used for many different applications. Like using decision tree algorithms, they analyzed the data to distinguish between different behaviors such as grazing, resting, and walking. The study also highlighted the potential of using motion sensors combined with machine learning techniques to automatically monitor animal behaviors in a farm setting. 

Another work by Megahed et al. \cite{megahed2022comparison} addressed the challenge of identifying risk factors for Brucella infection in dairy cattle, comparing two statistical methods: logistic regression and the Classification and Regression Tree (CART) analysis. By conducting a cross-sectional study with 400 animals across several Egyptian governorates, they focused on various potential risk factors like herd size, history of abortion, and practices of disinfection. The findings showed that CART was superior in predicting the risk factors with high accuracy, suggesting it as a more effective tool for managing disease prevention in livestock populations. Furthermore, some work has also been done for the chickens. Like Averos et al. \cite{averos2024potential} explored the use of decision trees to streamline the assessment process for broiler chickens. Traditionally, such assessments are detailed and time-consuming which involves multiple indicators to measure the welfare of the flocks. By applying decision trees, the authors managed to reduce the number of indicators needed to effectively assess welfare while maintaining reliability. Some of the key indicators like cumulative mortality, immobility, and back wounds were identified as critical factors, while others also happened to be of some impact. The simplified model like decision tree allowed for faster and equally reliable welfare assessments.

\noindent \textbf{Clustering Methods.} In the field of livestock management, several studies have employed clustering algorithms to enhance monitoring and data analysis. One study by Ismail et al. \cite{ismail2019efficient} focuses on enhancing livestock monitoring through clustering techniques like density-based spatial clustering (DBSCAN) \cite{deng2020dbscan} and K-means \cite{ahmed2020k}. The study introduces an efficient system combining Faster R-CNN for object detection and clustering algorithms to manage and analyze livestock data. The results show that DBSCAN outperformed K-means in efficiently detecting herds and outliers, optimizing the monitoring process for dairy production by reducing manpower and time costs significantly. This approach provides a scalable solution for large-scale farming operations. Another study by Li et al. \cite{li2019use} explores the deployment of unmanned aerial vehicles (UAVs) equipped with GPS and streaming K-means clustering to optimize livestock monitoring. The study demonstrate that UAVs can effectively track and monitor livestock which is adapted to the animals' mobility by updating their positions based on GPS data. The streaming K-means clustering method offers an improvement in minimizing UAV-animal distances compared to traditional methods, which do not account for animal movement.

Furthermore, Hou et al. \cite{hou2020study} employs K-means and PCA clustering algorithms to analyze untagged multi-dimensional activity data from dairy cows. With a wearable activity data acquisition system, they provides insights into individual behaviors, aiding in the optimization of dairy farm operations by tailoring management practices to meet specific animal needs. Lastly, Skjerve \cite{skjerve2024using} utilizes DBSCAN and fuzzy c-means clustering to manage and analyze slaughter records from a livestock herd. This approach not only cleans data efficiently but also segments it based on attributes like age, carcass weight, and daily gain. The results offer detailed insights into production strategies, enhancing data-driven decisions in livestock management and reaveal how different farming strategies impact production outcomes.

\subsubsection{CNNs}
CNNs have been used for different applications in livestock rangin from breed classification, disease recognition, and animals detection. Following are some example models that have been used in the livestock domain.

\noindent \textbf{VGG16.} VGG16 is also used for the detection of common cattle diseases like foot and mouth disease (FMD), infectious bovine keratoconjunctivitis (IBK), and nodular dermatitis (LSD) with high accuracy \cite{10724717}. The authors used the fine-tuning strategy by using the ImageNet pre-trained VGG16 model on a dataset of 2,000 images. This model resulted in a total test accuracy of 88.14\% in disease diagnosis. Another study by Khan et al. \cite{articlekhan} utilizes the VGG-16 model to classify different types of animals from images including sheep and cows. The study involved training the model on a dataset of 12,984 images, representing six different animal species. The use of VGG-16 helps in accurately processing the images and classifying them by learning from various features like shapes and textures.

Another work by Sun et al. \cite{sun2021recognition} is based on recognizing cattle and sheep using VGG variants. They used data augmentations like random cropping of images to a uniform size of 224 $\times$ 224 pixels (input size required for VGG models), random counterclockwise rotations between 0-20 degrees, and horizontal flipping with 50\% probability rate. This approach allows the model to be trained on a diversified set of images to introduce better generalization to unseen images. The dataset consists of 260 images in total which were then increased to 3900 using data augmentation techniques discussed earlier. They performed experiments with VGG11, VGG13, VGG16, and VGG19 where VGG16 shows better performance with comparably better training time. 

\noindent \textbf{ResNet50.} ResNet-50 model has been used for the classification of the chicken gender by introducing several technical contributions to improve accuracy and efficiency \cite{wu2023improved}. The model integrates the Squeeze-and-Excitation (SE) attention mechanism within the ResNet-50's residual units to focus on the most informative feature channels, which significantly enhances the specificity of gender identification. Additionally, the Swish activation function is employed to address the vanishing gradient problem common with ReLU, aiding smoother and more effective training. The Ranger optimizer, combining RAdam \cite{liu2019variance} and LookAhead techniques, further refines the training process by stabilizing and speeding up convergence. The enhanced model was then trained on a dataset containing 960 images of chickens in various scenarios, leading to a high test accuracy of 98.42\%.

Another work by Gong et al. \cite{gong2022facial} developed a specialized facial recognition model for cattle based on SK-ResNet. SK-ResNet modifies the traditional ResNet-50 architecture by incorporating selective kernel units (SK units) that adjust the receptive field dynamically, enhancing the model's ability to accurately recognize cattle from various angles and under different environmental conditions. The model also features improved skip connections that incorporate maximum pooling to reduce information loss across layers. This technique is important as it is responsible for maintaining feature integrity in deeper networks. The use of the ELU activation function over the traditional ReLU helps overcome issues related to activation saturation, which promotes more robust learning. This model is trained on a dataset of 5677 cattle images which achieved an impressive accuracy of 98.42\%. This robust performance is also confirmed across diverse livestock datasets, including pigs and sheep.

To address the challenge of Lumpy Skin Disease (LSD) in cows, Chinta et al. \cite{chinta2023lumpy} develop a diagnostic model using the ResNet50 \cite{he2016deep}. It leverages image processing techniques such as Gaussian filtering and histogram equalization to enhance the quality of input images. The methodology involves pre-processing images to reduce noise and enhance contrast, followed by the ResNet-50 model to classify images into categories of lumpy or non-lumpy. The model was trained on a dataset of cow images to classify them based on the presence of lumpy skin symptoms, achieving a high classification accuracy of 93\%. Furthermore, several variants of ResNet has also been used for different applications. Wu et al. \cite{wu2024multi} used modified ResNet101 with an Atrous Spatial Pyramid Pooling (ASPP) module for detecting and extracting motion features from dairy cows. This combination is specifically tailored to improve the detection of multi-scale keypoints which is essential for analyzing cow behavior and health, such as lameness. The ResNet101-ASPP model addresses challenges posed by large variations in cow sizes and occlusions typical in real farming scenarios. They employed a dataset consisting of 2,385 images showing various motions of cows, to perform extensive ablation experiments comparing ResNet101-ASPP against other models like ResNet50 and EfficientNet-b0. Notably, the approach facilitated the simultaneous extraction of seven distinct types of motion features, including gait symmetry and back curvature, from the trajectories of identified keypoints. 

Bouchekara et al. \cite{bouchekara2022sift} introduce an image stitching technique called the SIFT-CNN pipeline based on ResNet-50 which is designed to enhance livestock monitoring via drone-based aerial imagery. They focused on addressing the challenges of sparse and feature-sparse images in extensive grazing fields. The method comprises a four-stage process. Initially, it preprocesses images to remove extraneous background, optimizing for feature extraction. This is followed by classifying images to distinguish cattle from empty spaces, which streamlines the stitching process. The core of the pipeline lies in its enhanced SIFT method that adapts to scale and rotation variations, thus improving the matching accuracy of features across different images. The final stage involves a projection transformation technique to stitch the images accurately, maintaining a realistic perspective of the terrain. This pipeline marks a significant step forward in automated surveillance, reducing the need for manual monitoring over large areas.

\noindent \textbf{Mask-RCNN.} A study by Qiao et al. \cite{qiao2019cattle} shows the development and application of a Mask R-CNN model \cite{he2017mask} for high-accuracy cattle segmentation and contour extraction within complex (outdoor) feedlot environments. They also incorporate image enhancement techniques like a two-dimensional gamma function-based image adaptive correction algorithm to reduce the impact of variable lighting conditions, which is often a very common challenge in outdoor settings. The training was done on a challenging dataset comprising over 1,000 high-resolution (1920 $\times$ 1080 at 30 fps) images, which include a variety of cattle movements and postures. Additionally, the framework also facilitates the extraction of detailed cattle contours which can be used to evaluate the health parameters like length, width, and back area of cattles.

\noindent \textbf{EfficientNet.} Himel et al. \cite{himel2024utilizing} use EfficientNet for identifying sheep breeds from low-resolution images which is a challenge faced by farmers with limited resources and expertise in breed differentiation. They employed EfficientNetB5, which achieved a high accuracy of 97.62\% on a test set comprising 10\% of their dataset with 1,680 facial images representing four distinct sheep breeds named Marino, poll Dorset, Suffolk, and white Suffolk. They also evaluated different variants of EfficientNet, like B0, B1, B2, V2, and others. 

Furthermore, Zhao et al. \cite{zhao2023automatic} develop a method for automatically scoring the body condition of dairy cows using a combination of EfficientNet \cite{tan2019efficientnet} and convex hull features derived from point clouds. The method involves preprocessing depth images to extract a 3D structure feature map from the cow’s dorsal area which is from the hook bone to the needle bone. This data is then processed through a two-level model based on the EfficientNet network, optimized by the whale optimization algorithm (WOA) \cite{mirjalili2016whale}, achieving a recognition accuracy of over 91\% for errors less than 0.25 in the body condition score. The framework utilizes 5,119 depth images from 77 cows, showcasing a cost-effective and high-accuracy approach to assess the nutritional and health status of dairy cows, which is vital for improving farm management and animal welfare. 

Moreover, Yin et al. \cite{yin2020using} focus on the recognition of dairy cow motion behaviors in complex environments by using the EfficientNet-LSTM model to distinguish behaviors such as lying, standing, walking, drinking, and feeding. The model uses EfficientNet for spatial feature extraction combined with a bidirectional LSTM that incorporates an attention mechanism, enhancing the temporal analysis of video frame sequences. For training purposes, the authors used over 2,270,250 frames from 1,009 videos, demonstrating a behavior recognition accuracy of 97.87\%, which outperforms traditional models like ResNet50-LSTM. This approach not only offers a more accurate and rapid behavior analysis but also reduces the reliance on manual monitoring and the stress associated with wearable sensors for cows.

\noindent \textbf{YOLO.} YOLO-BYTE \cite{zheng2023yolo} is an advanced multi-object tracking model designed to enhance the monitoring of dairy cows in complex environments. The model builds upon the YOLOv7 \cite{wang2023yolov7} backbone network by incorporating a Self-Attention and Convolution mixed module (ACmix) to better handle the uneven spatial distribution and target scale variation seen in cow tracking. An innovative lightweight Spatial Pyramid Pooling Cross Stage Partial Connections (SPPCSPC-L) module is also employed to reduce the model's complexity. These enhancements lead to an improved ByteTrack algorithm, which precisely matches tracking boxes to the cows by predicting the width and height information directly. The model demonstrated substantial improvements in tracking accuracy metrics such as Precision (97.3\%), Recall, and Average Precision, as well as in High Order Tracking Accuracy (HOTA), Multi-Object Tracking Accuracy (MOTA), and Identification F1 (IDF1) scores. The model's performance surpasses the original YOLOv7 implementation which shows significant reductions in model parameters and still achieving a real-time analysis speed of 47 frames per second.

Another study by Kurniadi et al. \cite{kurniadi2023innovation} explores the application of the YOLOv5 object detection model to monitor cattle using drones. The YOLOv5 model is trained on a dataset of 3,131 cow images and 836 non-cow images, which has been optimized to detect cattle effectively from aerial images captured by drones. The optimal training configuration achieved a precision of 94.3\%, a recall of 92.5\%, and a mean Average Precision (mAP) of 83.1\%. The model's performance varies with altitude with an accuracy of 75\% at five meters, which decreases as the drone's altitude increases.  Furthermore, the E-YOLO \cite{wang2024yolo} model is an adaptation of the YOLOv8 Nano (YOLOv8n) model which is then tailored for detecting estrus in cows that is essential for timely breeding in dairy farming. The adapted model incorporates the Normalized Wasserstein Distance (NWD) loss instead of the typical Intersection over Union (IoU) to improve sensitivity to position deviations in target cows. Additionally, a Context Information Augmentation Module (CIAM) and a Triplet Attention Module (TAM) have been integrated to focus on individual estrus cows by utilizing cross-dimensional interactions. It has been tested on a dataset with 1,716 instances of cow mounting behavior where it achieved a precision of 93.90\% for estrus detection, outperforming other advanced models such as SSD and Faster R-CNN \cite{liu2016ssd,ren2016faster}. The implementation of E-YOLO indeed represents a significant advancement in real-time and accurate monitoring of estrus cows.

\subsubsection{ViTs}

\noindent \textbf{UGTransformer.} UGTransformer \cite{wang2024ugtransformer} is a network developed to improve sheep extraction from remote sensing images having higher spatial resolution ranging from 29,899 $\times$ 26,371 to 29,392 $\times$ 25,900. This model is compared to traditional methods which often rely on hardware like radio frequency equipment and face challenges like loss or damage that makes them less reliable for large-scale monitoring. It incorporates a merge block in its encoder to fuse two scales of features which enhance the multiscale feature fusion capability. This design effectively integrates global context features and spatial detailed features through a global connectivity module containing two sliding sub-modules for horizontal and vertical correlation. The experimental results demonstrate that the UGTransformer outperforms contemporary models like UNet \cite{ronneberger2015u}, Deeplabv3+ \cite{chen2018encoder}, DCSwin \cite{wang2022novel}, BANet \cite{wang2021transformer}, and UNetFormer \cite{wang2022unetformer}, showing at least a 1.8\% increase in mIoU. This methodology enhances capabilities for small-object extraction in remote sensing applications.

\noindent \textbf{RTDETR‑Refa.} RTDETR \cite{zhao2024detrs} stands for real-time detection transformer which uses an efficient hybrid encoder in contrast to the original DETR \cite{carion2020end} model. It is more computationally light and improves performance by decoupling intra-scale interactions and cross-scale fusion. RTDETR-Refa \cite{li2025rtdetr} is developed upon RTDETR and designed for the real-time classification and identification of cattle breeds which uses a modified ResNet18 backbone network and incorporates a Faster-Block module with a 3×3 convolution to replace the standard 1×1, enhancing feature extraction speed and reducing computational load. An Efficient Multiscale Attention (EMA) module is added to boost feature transformation and classification accuracy. The training and validation set contains 1328 and 190 images, respectively. The model results in an improvement in cattle breed classification, achieving an average accuracy of 91.6\% on the classification training set, which is higher than the original model and other classical models by 0.8\% to 5.2\%. The RTDETR-Refa model exemplifies the potential of accurately identifying cattle breeds in real-time applications.

\noindent \textbf{EMSC-DETR.} Efficient Multi-Scale Chicken DETR (EMSC-DETR) \cite{li2024efficient} is an improved version of the RT-DETR \cite{zhao2024detrs} which is adapted for chicken detection in complex free-range farming environments. The model integrates a space-to-depth transformer module (SDTM) and a contextual transformer (CoT) to handle the challenges of multi-scale target detection and overcoming occlusion issues which are caused by chicken aggregation. The dataset contains 4500 images selected from total of nine different scenes. The EMSC-DETR shows exceptional performance with an AP50 of 98.6\% on the testing set, significantly outperforming the original RT-DETR. This model exhibits robust generalization across unknown scenes, making it highly applicable to commercial farming operations.

\noindent \textbf{MHAFF.} Multi-Head Attention Feature Fusion (MHAFF) \cite{dulal2025mhaff} model is a fusion-based approach combining the strengths of CNNs and ViTs for cattle identification using muzzle images. As the traditional methods often fail to capture long-range dependencies in complex muzzle patterns, therefore it shows a limitation that can be addressed by using the transformers to address their global information processing capabilities. MHAFF improves upon simple feature fusion methods like addition and concatenation by utilizing a dynamic, context-aware fusion approach that preserves discriminative information while minimizing dimensionality increase. Here, the CNN module is used to extract the local features while the ViT module attend to the global features. The model significantly outperformed traditional models like Resnet50, VGG16, Inception, ViT, and Swin by achieving near-perfect accuracy on two cattle datasets (99.88\% and 99.52\%).

\noindent \textbf{StarFormer.} The StarFormer \cite{tangirala2021livestock} is an end-to-end domain-adaptive transformer-based framework designed for the comprehensive monitoring of livestock behaviors. It integrates \textbf{S}egmentation, \textbf{T}racking, \textbf{A}ction recognition, and \textbf{R}e-identification (STAR) tasks into a single model, leveraging the capabilities of transformers to handle complex scenarios which are typical in livestock monitoring, such as varying animal sizes and frequent occlusions. The paper also introduces a new dataset named PIGTRACE for benchmarking the performance of livestock monitoring systems, featuring detailed annotations for various pig behaviors in real indoor farming environments. STARFORMER demonstrates superior performance over other models like Mask-RCNN and DETR \cite{he2017mask,carion2020end} by effectively learning instance-level embeddings and performing optimization across multiple STAR tasks simultaneously.

\noindent \textbf{MobileViTFace.} MobileViTFace \cite{li2023combining} is a sheep face recognition model that integrates convolutional structures with ViT architecture to overcome the challenges of training robust models with limited datasets. By incorporating the inductive biases of CNNs and the detailed, global processing strengths of ViT, MobileViTFace achieves significant performance improvements on a sheep face dataset with a high recognition accuracy of 97.13\%. This model addresses ViT's high computational demand and extensive data needs by utilizing downsampled feature maps from MobileNetV2 as input, effectively balancing performance with computational efficiency. The results confirm MobileViTFace's suitability for real-time applications on edge devices as well which demonstrate as a viable solution for deploying deep learning-based facial recognition in livestock management.

\subsubsection{Foundation Models for Livestock}

\noindent \textbf{ChickenSAM.} Yang et al. \cite{yang2023sam} explore the application of the Segment Anything Model (SAM) in poultry science, particularly focusing on cage-free hens. SAM is known for its robust zero-shot segmentation capabilities and it was evaluated against SegFormer \cite{xie2021segformer} and SETR models \cite{liu2022setr} for its effectiveness in chicken segmentation tasks using part-based and whole-body approaches, as well as infrared thermal imaging. Additionally, SAM was employed in tracking tasks to study the movement patterns of broiler birds. The findings indicate that SAM not only excels in segmentation tasks but also provides valuable behavioral data, which can significantly contribute to advancements in poultry management and welfare assessment.

\noindent \textbf{AnimalFormer.} AnimalFormer \cite{qazi2024animalformermultimodalvisionframework} is a groundbreaking multimodal vision framework aimed at enhancing precision livestock farming. The framework leverages three models including GroundingDINO \cite{liu2024grounding}, HQSAM \cite{ke2023segment}, and ViTPose \cite{xu2022vitpose}. These models operate with integrity to analyze animal behavior comprehensively from video data without requiring invasive tagging. GroundingDINO is responsible for generating precise bounding boxes around livestock, HQSAM segments individual animals within these boxes, and ViTPose estimates key body points to facilitate detailed posture and movement analysis. The dataset is curated for 5 different activities of sheep including grazing, running, sitting, standing, and walking. It contains 417 videos with 59 minutes long summing up to a total of 149,327 frames. The system was tested on a dataset featuring various sheep activities such as grazing, running, sitting, standing, and walking, proving its capability to capture a wide range of behaviors and interactions essential for thorough animal health and behavior analysis.

\noindent \textbf{WildCLIP.} WildCLIP \cite{gabeff2024wildclip} leverages domain-adapted VLMs to enhance the annotation and retrieval of camera trap data for ecological and ethological research. By fine-tuning the CLIP \cite{radford2021learning} model on wildlife images, this approach addresses the challenge of annotating vast amounts of camera trap data which allows the retrieval of images based on environmental and ecological attributes. The adaptation involves incorporating an adapter module that enables the extension of the model's vocabulary which improves its ability to handle open-set queries with few-shot learning. The system is evaluated using the Snapshot Serengeti dataset which was collected over eleven seasons and consists of more than 7 million images. They also introduced 10 different templates of the captions and found that training on diverse set of templates does not out-perform the single template training approach.

\noindent \textbf{ChickenFaceMAE.} Ma et al. \cite{ma2022advanced} presents a chicken face detection framework that utilizes Generative Adversarial Networks (GANs) and Masked Autoencoders (MAEs) to enhance the accuracy of detecting chicken faces across different growth stages. This detection is important for various applications like behavior recognition, health monitoring, and day-age detection. They also leverage the YOLOv4 backbone which is why the proposed model achieves notable improvements in real-time inference by reaching a precision of 0.91, an mAP of 0.84, and a processing speed of 37 FPS. These metrics represent significant advancements over traditional two-stage RCNN and EfficientDet models. The framework's design also involves generating synthetic images using GANs and MAEs to augment the training dataset which addresses the scarcity and variability of chicken face images. The enhanced feature extraction is achieved by incorporating a CSPDarknet53 backbone that adapts the detection to different object sizes that is considered an essential feature given the diverse growth stages of chickens.

\noindent \textbf{EfficientSAM.} Noe et al. \cite{noe2023efficient} introduces an efficient segmentation model for extracting mask regions of black cattle in livestock monitoring scenarios. The method employs models like SAM, Grounding Dino, and YOLOv8, alongside DeepOCSort for tracking. This tend to be a comprehensive approach to cattle monitoring. This model excels in creating accurate segmentation masks that are crucial for tracking cattle across different frames. This approach enhances the ability to monitor and analyze cattle behavior and health non-invasively. The integration of these techniques ensures robust detection and tracking, which is validated through rigorous testing.


\section{Conclusion and Future Directions}
\label{sec:conclusion}
The rapid shift from conventional machine learning solutions to domain-tuned foundations and VLMs has re-energized digital agriculture. Yet the journey toward fully automated and sustainable agriculture is far from complete. Below, we outline several concrete research avenues that can shape the next decade of AI for crops, fisheries, and livestock.

\subsection{Unifying Multi-modal Foundation Models}
The next leap in digital agriculture will come from models that natively understand every data stream produced by modern farming practices like drone or satellite imagery, soil-moisture logs, weather forecasts, acoustic cues, feed and veterinary records, and free-form farmer notes using reasoning or Agentic-AI approach. Understanding these data modalities in a single backbone will create rich, transferable embeddings that can recognize subtle cross-modal patterns such as early crop stress or animal welfare issues. Similar Work has already been done on image-text modalities like AgriCLIP \cite{nawaz2024agriclip}, AgroGPT \cite{awais2024agrogpt}, and MarineInst \cite{zheng2024marineinst}. However, such work can be extended by incorporating more diverse modalities to obtain universal encoders that learn contrastively across images, text, and time-series data.

\subsection{Continual and Federated Adaptation}
Agricultural environments change constantly with seasons, breeds, and climate trends, so static models degrade quickly. An emerging solution is to embed continual-learning mechanisms that update parameters on edge without catastrophic forgetting. At the same time, federated learning lets many devices improve a shared model without sending raw data off the region. For this purpose, lightweight adapter layers or prompt-tuning can deliver rapid few-shot adaptation when an agronomist needs to add an unseen pest or disease type into the model. By using such techniques, the models will be able to evolve continuously.

\subsection{Edge-Efficient AI for Real-time Decision Support}
Cloud-scale VLMs are typically too large and power-hungry for small devices like mobiles, drones, battery-operated cameras, or underwater robots, especially those without reliable internet connections. To solve this, researchers are developing smaller and efficient models such as MobileNetV4 \cite{qin2024mobilenetv4}, Mobile-VideoGPT \cite{shaker2025mobile}, and MobileVLM-v2 \cite{chu2024mobilevlm}. Additionally, optimizing computing power, energy use, and data transmission will determine if the device can process data locally in the agricultural field or must send it elsewhere. Therefore, such optimized and efficient solutions are required to enable handy and user-friendly experience.

\subsection{Agentic AI for Autonomous Farming}
Agentic AI brings an additional layer of autonomy and closed-loop reasoning to multi-modal foundation models, transforming them from passive feature extractors into proactive problem solvers. In this paradigm, a single, unified backbone not only embeds and aligns heterogeneous inputs (e.g., imagery, time-series, text) but also directs a sequence of interpretive and decision-making steps: it formulates hypotheses (Is this pattern indicative of moisture stress?), generates plans (Schedule a targeted drone survey for Field B at dawn), invokes specialized sub-modules (Run the soil-moisture predictor on the last 48 h of sensor data), and iteratively refines its conclusions based on feedback or new observations. By integrating chain-of-thought style reasoning directly into the network architecture rather than as a post-hoc prompting strategy, Agentic AI can enable richer, context-aware embeddings that capture not only cross-modal correlations but also the causal and temporal logic underlying on-farm phenomena. When such a system is deployed in digital agriculture, it can autonomously monitor evolving crop conditions, detect anomalies in livestock behavior, and even trigger corrective actions (e.g., adjusting irrigation schedules or flagging veterinary interventions) with minimal human supervision. Ultimately, embedding Agentic AI and explicit reasoning into a universal, contrastively trained encoder promises to deliver truly intelligent farm assistant models that do not merely perceive the world, but also understand, plan, and act within it.

\subsection{Future Outlook}
The next decade will see agriculture transition from isolated AI prototypes to fully integrated, multi-domain intelligent systems spanning crops, fisheries, and livestock. Similar to how the mechanization wave of Agriculture 2.0 and the sensor revolution of Agriculture 3.0 reshaped farming, the coming phase will be defined by multimodal AI, edge-based real-time decision systems, and domain-adaptable foundation models. These systems will merge visual, environmental, and textual data streams to produce actionable insights across the entire food production chain.

Today's global pressure drivers, like climate variability, water scarcity, labor shortages, and rising food demand, are accelerating adoption. The strategic priorities emerging and catalyzing investments worldwide are: \textit{(i)} Edge deployment for low-latency analytics, \textit{(ii)} Cross-domain interoperability between crops, fisheries, and livestock, and \textit{(iii)} tight integration with autonomous systems such as drones, mobile robots, and IoT sensor networks.

As VLMs mature, farms will increasingly combine heterogeneous input like satellite crop images, underwater fish-behavior videos, thermal livestock scans, and soil or water chemistry readings into unified decision-support layers. This will enable classification, detection, segmentation, tracking, and anomaly detection in a single workflow. For example, a single deployed model could diagnose wheat rust from a leaf scan, detect feed anomalies in a cattle barn, and identify abnormal swimming patterns in aquaculture ponds, all on the same edge device.

The integration of edge AI with high-bandwidth networks is critical. Comparable to how 5G pilots in agriculture enabled live-streaming of high-definition data for remote greenhouse and aquaculture management, next-generation connectivity will push latency into the millisecond range, unlocking truly real-time responses. These systems will be self-calibrating through reinforcement or continual learning, adapting to new crop varieties, fish species, and disease strains without full retraining. Several current initiatives point toward this transformation which are as follows:

\begin{enumerate}
    \item \textbf{Crops (Greenhouse Autonomy):} The Netherlands’ Autonomous Greenhouse Challenge demonstrates AI-controlled greenhouses using canopy segmentation, fruit detection, and yield forecasting, delivering higher yields and energy savings compared to manual control \cite{WURstren45}. Concurrently, UAE operations like Pure Harvest greenhouses combine vision analytics with IoT fertigation to maximize water-use efficiency in arid climates \cite{Seedsofc40}.

    \item \textbf{Water-Scarce Agriculture (Middle East):} In the Middle East and North Africa (MENA), where over 80\% of land is desert, regenerative agriculture is being shaped primarily by the need to restore soil health and improve water-use efficiency rather than solely to mitigate climate change. Recent initiatives focus on scalable practices such as biochar application, gaining traction in the UAE, and no-till farming, increasingly adopted in Morocco \cite{NewAGInt93}. These approaches are supported by emerging carbon-credit incentives, which are fostering business models and policy frameworks that promote sustainable soil management in one of the most water-scarce regions of the world. Similarly, Jordan's reuse of treated wastewater integrates crop health anomaly detection to monitor irrigation safety and adjust schedules accordingly \cite{key, wikipediaWaterSupply}. 

    \item \textbf{Fisheries (Aquaculture Health Monitoring in Korea):} South Korea’s National Institute of Fisheries Science (NIFS) has developed and demonstrated Miribom, an AI‑powered tool for diagnosing parasitic diseases in farmed flounder by analyzing images captured via smartphone on-site \cite{Parasiti53}. It enables early-stage identification of conditions like scutica disease and fish emaciation, which are crucial, given that parasitic infections account for over 50\% of flounder farm mortalities. This advance demonstrates how object‑detection models applied to underwater (or smartphone-captured) imagery streamline disease identification and improve aquaculture health management.

    \item \textbf{Livestock (Dairy Welfare Monitoring in Europe):} In dairy farming, advanced computer vision systems are now capable of real-time lameness detection, individual cow tracking, and behavior analysis by extracting multiple locomotion traits such as back posture, head bobbing, stride length, and tracking distance from video using pose estimation models like T‑LEAP \cite{russello2024video}. These systems significantly improve classification accuracy (up to ~80\%) by combining traits versus single-feature analysis. They illustrate how detection, tracking, and fine-grained classification tasks can support automated welfare monitoring.

    \item \textbf{Policy-Driven Monitoring (US Case, California Groundwater):} In California’s Central Valley, machine learning models including XGBoost ensembles are being used to forecast groundwater levels three months in advance, using publicly available meteorological and in‑situ well data, to support Sustainable Groundwater Management Act (SGMA) objectives \cite{may2023forecasting}. These models enable outlier detection and predictive planning in water-stressed regions, forming the basis of automated feedback loops for regulatory compliance and sustainable agriculture water management.

\end{enumerate}

\bibliographystyle{IEEEtran}
\bibliography{cas-refs}











\newpage

\section{Biography Section}
 

\begin{IEEEbiography}[{\includegraphics[width=1in,height=1.4in,clip,keepaspectratio]{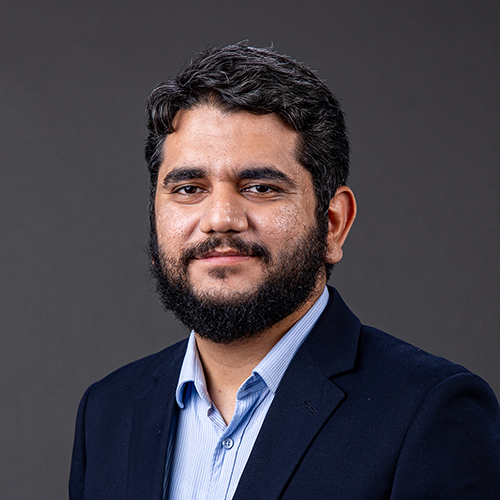}}]{Umair Nawaz}
is a PhD student at MBZUAI, Abu Dhabi, UAE. He has a Master's in Computer Vision from MBZUAI and a B.Sc in Electrical Engineering from the University of Engineering and Technology, Lahore, Pakistan. He has been actively working on applications of deep learning to real‑world problems in agriculture and healthcare. Specifically, his research focuses on vision–language models, Agentic AI, adversarial robustness, and edge‑device deployment. Umair has held research and engineering roles at Qatar University, AIQ Abu Dhabi, Verideal Technology, and LUMS.

\end{IEEEbiography}

\begin{IEEEbiography}[{\includegraphics[width=1in,height=1.4in,clip,keepaspectratio]{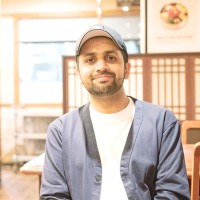}}]{Muhammad Zaigham Zaheer}
received his PhD degree from the University of Science and Technology in 2022. He is currently associated with Mohamed bin Zayed University of Artificial Intelligence as a Research Scientist. Previously, he worked at the Electronics and Telecommunications Research Institute (ETRI) in South Korea, as a postdoctoral researcher.  His current research interests include Vision Language Models and their applications, self-supervised leaning and unsupervised learning.

\end{IEEEbiography}

\begin{IEEEbiography}[{\includegraphics[width=1in,height=1.4in,clip,keepaspectratio]{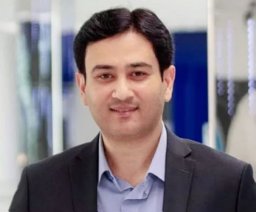}}]{Fahad Shahbaz Khan}
received the MSc degree in intelligent systems design from the Chalmers University of Technology, Sweden, and the PhD degree in computer vision from Computer Vision Center Barcelona and Autonomous University of Barcelona, Spain. He is currently a professor of computer vision with MBZUAI, United Arab Emirates. He also holds a faculty position at Computer Vision Laboratory, Linköping University, Sweden. He has achieved top ranks on various international challenges (Visual Object Tracking VOT: 1st 2014 and 2018, 2nd 2015, 1st 2016; VOT-TIR: 1st 2015 and 2016; OpenCV Tracking: 1st 2015; 1st PASCAL VOC Segmentation and Action Recognition tasks 2010). He received the best paper award in the computer vision track at IEEE ICPR 2016. He has published more than 100 peer-reviewed conference papers, journal articles, and book contributions. His research interests include a wide range of topics within computer vision and machine learning. He serves as a regular senior program committee member for leading conferences such as CVPR, ICCV, and NeurIPS.

\end{IEEEbiography}

\begin{IEEEbiography}[{\includegraphics[width=1in,height=1.4in,clip,keepaspectratio]{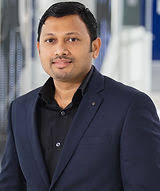}}]{Hisham Cholakkal}
received the MTech degree from IIT Guwahati, India, and the PhD degree from Nanyang Technological University, Singapore. He is an assistant professor with the MBZ University of Artificial Intelligence (MBZUAI), UAE. Previously, from 2018 to 2020, he worked as a research scientist with the Inception Institute of Artificial Intelligence, UAE; from 2016 to 2018, he worked as a senior technical lead with Mercedes-Benz R\&D India. He has also worked as a researcher with BEL-Central Research Lab, India, and Advanced Digital Sciences Center, Singapore. He has organized workshops at top venues such as ICCV 2023, NeurIPS 2022, and ACCV 2022 and regularly serves as a program committee member for top conferences, including CVPR, ICCV, NeurIPS, ICLR, and ECCV.

\end{IEEEbiography}

\begin{IEEEbiography}[{\includegraphics[width=1in,height=1.4in,clip,keepaspectratio]{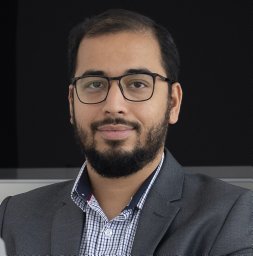}}]{Salman Khan}
received the PhD degree from the University of Western Australia, in 2016. He is an associate professor with the MBZ University of Artificial Intelligence. He has been an adjunct faculty with the Australian National University since 2016. He has been awarded the Outstanding Reviewer award at IEEE CVPR multiple times, won the Best Paper award at 9th ICPRAM 2020, and won 2nd prize in the NTIRE Image Enhancement Competition alongside CVPR 2019. He served as a (senior) program committee member for several premier conferences, including CVPR, ICCV, ICML, ECCV, and NeurIPS. His thesis received an honorable mention on the Dean’s List Award. He has published more than 100 papers in high-impact scientific journals and conferences. His research interests include computer vision and machine learning.

\end{IEEEbiography}

\begin{IEEEbiography}[{\includegraphics[width=1in,height=1.4in,clip,keepaspectratio]{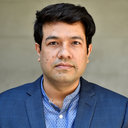}}]{Rao Muhammad Anwer}
is an assistant professor of computer vision with MBZUAI. Before joining MBZUAI, Anwer was with the Inception Institute of Artificial Intelligence (IIAI) in Abu Dhabi, United Arab Emirates, working as a research scientist. Before joining IIAI, he was a researcher with Aalto University, Finland. His research interests include visual object recognition, pedestrian detection and action recognition, efficient and robust deep learning models for comprehensive scene understanding, and human visual relationship detection.

\end{IEEEbiography}

\vfill

\end{document}